\documentclass{ps}
\usepackage[numbers,sort&compress]{natbib}
\usepackage[utf8]{inputenc} 
\usepackage[T1]{fontenc}    
\usepackage[hidelinks,colorlinks=true,linkcolor=blue,citecolor=blue]{hyperref}       
\usepackage{url}            
\usepackage{booktabs}       
\usepackage{amsfonts}       
\usepackage{amsmath}        
\usepackage{amssymb}        
\usepackage{amsthm}         
\usepackage{nicefrac}       
\usepackage{bbm}            
\usepackage{microtype}      
\usepackage{xcolor}         
\usepackage{verbatim}       
\usepackage{caption}        
\usepackage{subcaption}     
\usepackage[pdftex]{graphicx}
\usepackage[ruled,noend]{algorithm2e}
\usepackage[noabbrev]{cleveref}

\crefname{equation}{}{}
\crefname{algorithm}{Algorithm}{Algorithms}

\theoremstyle{plain} 
\newtheorem{theorem}{Theorem}
\newtheorem{lemma}{Lemma}
\newtheorem{proposition}{Proposition}
\newtheorem{corollary}{Corollary}

\crefname{theorem}{Theorem}{Theorems}
\crefname{lemma}{Lemma}{Lemmas}
\crefname{proposition}{Proposition}{Propositions}
\crefname{corollary}{Corollary}{Corollaries}

\theoremstyle{definition} 
\newtheorem{assumption}{Assumption}
\newenvironment{assumptionp}[1]{
\renewcommand\theassumption{\arabic{assumption}#1}\assumption}{\endassumption}

\crefname{definition}{Definition}{Definitions}
\crefname{assumption}{Assumption}{Assumptions}
\crefname{assumptionp}{Assumption}{Assumptions}
\crefname{example}{Example}{Examples}

\theoremstyle{remark} 

\crefname{remark}{Remark}{Remarks}

\DeclareMathOperator*{\argmin}{arg\,min}

\newcommand{\indicator}[1]{\mathbbm{1}_{\scriptscriptstyle\left\{#1\right\}}}

\newcommand{\E}{\mathbb{E}} 
\newcommand{\R}{\mathbb{R}} 
\newcommand{\N}{\mathbb{N}} 
\newcommand{\F}{\mathcal{F}} 

\begin{document}
\title{Non-asymptotic analysis of stochastic approximation algorithms for streaming data}\thanks{This work was supported by R\'egion \^Ile-de-France project number 19006497}
\author{Antoine Godichon-Baggioni}\address{Sorbonne Universit\'e, Laboratoire de Probabilit\'es, Statistique et Mod\'elisation (LPSM), F-75005 Paris, France; \\ \email{\{antoine.godichon\_baggioni, nicklas.werge, olivier.wintenberger\}@sorbonne-universite.fr}}
\author{Nicklas Werge}\sameaddress{1}
\author{Olivier Wintenberger}\sameaddress{1}\secondaddress{Wolfgang Pauli Institut, c/o Fakult\"at f\"ur Mathematik, Universit\"at Wien, 1090 Vienna, Austria }
\date{The dates will be set by the publisher}
\begin{abstract}
We introduce a streaming framework for analyzing stochastic approximation/optimization problems.
This streaming framework is analogous to solving optimization problems using time-varying mini-batches that arrive sequentially.
We provide non-asymptotic convergence rates of various gradient-based algorithms; this includes the famous Stochastic Gradient (SG) descent (a.k.a. Robbins-Monro algorithm), mini-batch SG and time-varying mini-batch SG algorithms, as well as their iterated averages (a.k.a. Polyak-Ruppert averaging).
We show i) how to accelerate convergence by choosing the learning rate according to the time-varying mini-batches, ii) that Polyak-Ruppert averaging achieves optimal convergence in terms of attaining the Cramer-Rao lower bound, and iii) how time-varying mini-batches together with Polyak-Ruppert averaging can provide variance reduction and accelerate convergence simultaneously, which is advantageous for many learning problems, such as online, sequential, and large-scale learning.
We further demonstrate these favorable effects for various time-varying mini-batches.
\end{abstract}
\maketitle

\section{Introduction}

Machine learning-based intelligent systems are becoming more and more widespread in modern society \citep{lecun2015deep,hastie2009elements}.
A crucial component of machine learning is optimization, which, in this context, involves estimating parameters for the intelligent systems to make decisions about future data.
A growing challenge is that these future data will arrive in an endless stream, for example through sensors from real-time measurement of weather, traffic and e-commerce, to name a few; we call these \emph{streaming data}.
Such streaming data arrives sequentially in time-varying mini-batches.
This places wide demands on computational efficiency and the robustness of the underlying optimization algorithms, which must be updated sequentially as more data becomes available.

Stochastic approximation/optimization algorithms have proven effective in handling large amounts of data and perform well across many fields ranging from smooth and strongly convex problems to complex non-convex ones; \citet{bottou2018optimization} reviews such algorithms for large-scale problems in machine learning.
Among these, the most well-known is probably the Stochastic Gradient (SG) descent introduced in \citet{robbins1951stochastic}, which forms the basis of many optimization algorithms used in machine learning \citep{lan2020first,shalev2012online,kushner2003stochastic,nesterov2018lectures}.
In a nutshell, these SG-based algorithms minimize the objective (a.k.a. loss or risk) of a model by iteratively updating the model parameters using stochastic approximations of its gradient.
Traditionally, these gradients are processed individually or in (fixed) mini-batches taken from a (fixed) dataset.
However, in our streaming framework, these gradients must be computed as a sequential stream of time-varying mini-batches.

\textbf{Contributions.} The objective of this paper is to solve stochastic approximation/optimization problems in a streaming framework.
Our main theoretical contribution is the non-asymptotic analysis of SG-based algorithms in this streaming framework, extending the work of \citet{bach2011non}.
This means that we investigate everything from the classical SG descent to time-varying mini-batch SG-based algorithms, as well as their Polyak-Ruppert extensions.
Our results show how to accelerate convergence by choosing the learning rate according to the time-varying mini-batches.
In addition, we show that Polyak-Ruppert averaging \citep{polyak1992acceleration,ruppert1988efficient} achieves optimal convergence in terms of achieving the Cramer-Rao lower bound in this streaming framework.
In particular, we show how time-varying mini-batches together with Polyak-Ruppert averaging can provide variance reduction and accelerate convergence simultaneously, without jeopardizing the computational complexity.
These theoretical findings are demonstrated for various streaming settings of time-varying mini-batches.

\textbf{Organization.} \Cref{sec:sub:problem_formulation} presents the streaming framework, in which we will analyze the stochastic algorithms.
In \cref{sec:ssg_methods}, we introduce our stochastic streaming-gradient algorithms, their projected versions and Polyak-Ruppert extensions.
The main results, namely the non-asymptotic convergence analysis, are presented in \cref{sec:nonasymptotic}.
These results are illustrated in \cref{sec:experiments} for various time-varying mini-batches.
In \cref{sec:conclusion}, we provide some concluding remarks with related future perspectives.

\section{Problem formulation} \label{sec:sub:problem_formulation}

Our objective is to solve stochastic approximation/optimization problems in a streaming framework, where data arrives sequentially in time-varying mini-batches; we consider problems on the form 
\begin{align} \label{eq:so_problem}
\min_{\theta\in\R^{d}}\{F(\theta):=\E[f(\theta)]\}.
\end{align}
We will refer to $F:\R^{d}\rightarrow\R$ as the objective function, but in the literature, $F$ is also known as the expected loss (and risk); See for instance  \citet{bottou2018optimization}.
Let $\theta^{*}$ denote the global minimum of $F$, and assume that $\theta^{*}\in\Theta$, where $\Theta$ is a closed convex set in $\R^{d}$.
Typical convergence results measure how quickly some estimate $\theta_{t}$ approaches $\theta^{*}$ (or the function value $L(\theta_{t})$ approaches $F(\theta^{*})$).
In this paper we are interested in bounding the quantity $\E[\lVert\theta_{t}-\theta^{*}\rVert^{2}]$.
As in \citep{bach2011non,gower2019sgd}, we make the analysis more convenient through convexity and smoothness assumptions on $F$ in \cref{eq:so_problem}.\footnote{Milder degrees of convexity have been investigated; \citet{godichon2019lp} studied SG algorithms under local strongly convexity, \citet{karimi2016linear} studied SG algorithms under the Polyak-Łojasiewicz condition \citep{polyak1963gradient,lojasiewicz1963topological}, and \citet{gadat2023optimal} studied the Ruppert-Polyak averaging estimate under some Kurdyka-Łojasiewicz-type condition \citep{kurdyka1998gradients,lojasiewicz1963topological}.}
The following assumptions are frequently referenced.
\begin{assumptionp}{}[$\mu$-quasi-strong convex \citep{karimi2016linear,necoara2019linear}] \label{assump:L_strong_convexity}
The objective function $F:\R^{d}\rightarrow\R$ is differentiable with $\nabla_{\theta}F(\theta^{*})=0$ and there exists a constant $\mu>0$ such that $\forall\theta\in\Theta$,
\begin{align} \label{eq:assump:L_strong_convexity}
F(\theta^{*})\geq F(\theta)+\langle\nabla_{\theta} F(\theta),\theta^{*}-\theta\rangle + \frac{\mu}{2}\lVert\theta^{*}-\theta\rVert^{2}.
\end{align}
\end{assumptionp}
\begin{assumptionp}{}[$C_{\nabla}$-Lipschitz smoothness] \label{assump:L_lipschitz}
The function $\nabla_{\theta} F$ is $C_{\nabla}$-Lipschitz continuous around $\theta^{*}$, i.e., there exists $C_{\nabla}>0$ such that $\forall\theta\in\Theta$,
\begin{align} \label{eq:assump:L_lipschitz}
\lVert\nabla_{\theta}F(\theta)-\nabla_{\theta}F(\theta^{*})\rVert \leq C_{\nabla}\lVert\theta-\theta^{*}\rVert.
\end{align}
\end{assumptionp}

\textbf{Streaming framework and notation.} Let each $(f_{t}(\theta))$ constitute a sequence of independent differentiable random functions (possibly non-convex) and let their gradients be unbiased estimates of $\nabla_{\theta}F(\theta)$, see e.g. \citet{nesterov2018lectures} for definitions and properties of such functions.
The shorthand notation of 
\begin{align*}
(f_t(\theta)) \text{ represents the sequence of time-varying mini-batches parameterized by } \theta.
\end{align*}
We say that each $f_{t}$ consist of $n_{t}\in\N$ data points, which we denote by the set $\{f_{t,1},\dots,f_{t,n_{t}}\}$.
For example, for a class of models $\{h_{\theta}\}_{\theta\in\Theta}$ parameterized by $\theta$, a loss function $l$, and a regularizer $\Omega$, then $f_{t,i}(\theta)$ can be seen as the composition:
\begin{align} \label{eq:so_example}
f_{t,i}(\theta)=l(y_{t,i},h_{\theta}(x_{t,i}))+\Omega(\theta).
\end{align}
where $\{(x_{t,i},y_{t,i})\}_{i=1}^{n_{t}}$ is a time-varying mini-batch of i.i.d. input-output data points with generic element $(x,y)\in\mathcal{X}\times\mathcal{Y}$.
The associated objective function from \cref{eq:so_problem} thus corresponds to having $F(\theta)=\E[f(\theta)]$ with $f(\theta)=l(y,h_{\theta}(x))+\Omega(\theta)$.

Our streaming framework includes many machine learning problems, from classification, and regression to ranking; this includes stochastic approximation (Robbins-Monro setting \citep{robbins1951stochastic}), learning from i.i.d. data with linear, logistic, softmax, quantile and general ridge regression, and $p$-means and geometric median under regularity conditions \citep{hastie2009elements,teo2007scalable,steinwart2011estimating,cardot2013efficient,nesterov2018lectures,lan2020first}.
More specifically, \cref{eq:so_example} could be the $l_{2}$-regularized least squares regression model, $f(\theta)=(\langle x, \theta\rangle-y)+\frac{\lambda}{2}\lVert\theta\rVert^{2}$ with $\mathcal{X}=\R^{d}$ and $\mathcal{Y}=\R$, or the $l_{2}$-regularized logistic regression for binary classification, $f(\theta)=\log(1+\exp(-y\langle x, \theta\rangle))+\frac{\lambda}{2}\lVert\theta\rVert^{2}$ with $\mathcal{X}=\R^{d}$ and $\mathcal{Y}=\{-1,1\}$; here, we used $h_{\theta}(x)=\langle x, \theta\rangle$.

\section{Stochastic streaming gradient algorithms} \label{sec:ssg_methods}

SG-based algorithms, which dates back to the seminal work of \citet{robbins1951stochastic}, have become the predominant optimization algorithm for solving these stochastic approximation/optimization problems.
To solve problem \cref{eq:so_problem} in our streaming framework, we introduce the Stochastic Streaming Gradient (SSG) algorithm, defined as the recursion
\begin{flalign} \label{eq:streaming_grad}
\textbf{(SSG)} && \theta_{t} =& \theta_{t-1}-\frac{\gamma_{t}}{n_{t}} \sum_{i=1}^{n_{t}}\nabla_{\theta} f_{t,i}(\theta_{t-1}), \quad \theta_{0}\in\R^{d}, &&
\end{flalign}
where $(\gamma_{t})$ is the learning rate satisfying $\sum_{i=1}^{t} \gamma_{i} = \infty$ and  $\sum_{i=1}^{t} \gamma_{i}^{2} < \infty$ for $t \rightarrow \infty$.
This SSG algorithm sequentially processes the time-varying mini-batches.
Note that if for all $ t\ge 1$, $n_{t}=1$, then the SSG algorithm is an online version of the well-known SG descent.

In many machine learning models, there may be restrictions on the parameter space of $\theta$.
We embrace this by defining a projected version of SSG, given as
\begin{flalign} \label{eq:ssg:proj}
\textbf{(PSSG)} && \theta_{t} =& \mathcal{P}_{\Theta}\left( \theta_{t-1} - \frac{\gamma_{t}}{n_{t}} \sum_{i=1}^{n_{t}} \nabla_{\theta} f_{t,i}\left( \theta_{t-1} \right)\right), \quad \theta_{0}\in\Theta, &&
\end{flalign}
where $\mathcal{P}_{\Theta}$ denotes the Euclidean projection onto the closed convex set $\Theta$ in $\R^{d}$, i.e., $\mathcal{P}_{\Theta}(\theta)=\argmin_{\theta'\in\Theta}\lVert\theta-\theta'\rVert_{2}$.
It is worth noting that SG-based algorithms are not gradient descent in the sense that the objective function values often increase, but only decrease on average; examples of this are illustrated in \cref{sec:experiments}.
Therefore, it makes intuitive sense to use sets of stochastic gradient $\{\nabla_{\theta}f_{t,i}\}_{i=1}^{n_{t}}$ in each iteration, as it naturally reduces the variance and makes it easier to adjust the learning rate $(\gamma_{t})$, which (on average) improves the convergence.

Next, let's consider a streaming variant of the celebrated Polyak-Ruppert averaging procedure \citep{polyak1992acceleration,ruppert1988efficient}:
\begin{flalign} \label{eq:avg:streaming_grad_est}
\textbf{(ASSG)/(APSSG)} && \bar{\theta}_{t} =& \frac{1}{N_t} \sum_{i=0}^{t-1} n_{i+1} \theta_{i}, &&
\end{flalign}
where $N_{t}=\sum_{i=1}^{t} n_{i}$ denotes the accumulated number of data points processed at each $t\in\N$.
This averaging procedure sequentially aggregates the estimates of \cref{eq:streaming_grad,eq:ssg:proj}, which stabilizes and accelerates convergence \citep{polyak1992acceleration,nemirovski2009robust}.
In particular, this average allows us to obtain the optimal Cramer-Rao lower bound.
Note that \cref{eq:avg:streaming_grad_est} does not actually change the estimates produced by the SSG or PSSG algorithms, but instead simply keeps track of a running average over the estimates.
Practically, as we handle data sequentially, we will make use of the recursive formula, $\bar{\theta}_{t} = (N_{t-1}/N_{t}) \bar{\theta}_{t-1} + (n_{t}/N_{t})\theta_{t-1}$.
A detailed overview of our stochastic streaming gradient algorithms (defined in \cref{eq:streaming_grad,eq:ssg:proj,eq:avg:streaming_grad_est}) is presented in \cref{algo:ssg}.

\begin{algorithm}[htp]
\caption{Stochastic streaming gradient algorithms (SSG/PSSG/ASSG/APSSG)} \label{algo:ssg}
\SetAlgoLined
\DontPrintSemicolon
\SetKwComment{Comment}{/* }{ */}
\KwIn{$\theta_{0}\in\Theta\subseteq\R^{d}$, $project\in\{\KwSty{True},\KwSty{False}\}$, $average\in\{\KwSty{True},\KwSty{False}\}$}
\KwOut{$\theta_{t}$, $\bar{\theta}_{t}$ (resulting estimates)}
Initialization: $\bar{\theta}_{0}\in\R^{d}$ \;
\For{each $t\geq1$, a time-varying mini-batch of $n_{t}$ data arrives,}{
$\theta_{t} \gets \theta_{t-1} - \frac{\gamma_{t}}{n_{t}} \sum_{i=1}^{n_{t}} \nabla_{\theta} f_{t,i}\left( \theta_{t-1} \right)$ \Comment*[r]{update}
  \If{project}{
    $\theta_{t} \gets \mathcal{P}_{\Theta}(\theta_{t})$ \Comment*[r]{project}
  }
  \If{average}{
    $\bar{\theta}_{t} \gets(N_{t-1}/N_{t}) \bar{\theta}_{t-1} + (n_{t}/N_{t})\theta_{t-1}$ \Comment*[r]{average}
  }
}
\end{algorithm}

\section{Non-asymptotic convergence analysis} \label{sec:nonasymptotic}

Throughout this paper, we consider learning rates $(\gamma_{t})$ of the form
\begin{align*}
\gamma_{t}:=C_{\gamma}n_{t}^{\beta}t^{-\alpha},
\end{align*}
with $C_{\gamma}>0$, $ \beta\in[0,1]$, and $\alpha$ chosen according to the time-varying mini-batches $n_{t}$.
This learning rate allows us to add more weight to larger mini-batches $(n_{t})$ through the $\beta$ parameter.
Note that \citet{bach2011non} considered learning rates of the same form, but with $\beta=0$ (and $n_{t}=1$).
For simplicity, we let the time-varying mini-batches $(n_{t})$ be given as
\begin{align*}
n_{t}:=\lceil C_{\rho} t^{\rho} \rceil,
\end{align*}
with $C_{\rho}\in\N$ and $\rho\in(-1,1)$ such that $n_{t}\geq1$ for all $t\in\N$. This setting includes classical (online) SG descent algorithms (i.e., $\{C_{\rho}=1,\rho=0\}$) and (online) mini-batch procedures of both constant and time-varying size (i.e. $\{C_{\rho}\in\N,\rho=0\}$ and $\{C_{\rho}\in\N,\rho\in(-1,1)\}$), as well as the Polyak-Ruppert average of (online) time-varying mini-batches.
We will refer to $C_{\rho}$ as the \emph{mini-batch size} and $\rho$ as the \emph{mini-batch rate}.

Our goal is to non-asymptotic bound the quantities $\delta_{t}:=\E[\lVert\theta_{t}-\theta^{*}\rVert^{2}]$ and $\bar{\delta}_{t}:=\E[\lVert\bar{\theta}_{t}-\theta^{*}\rVert^{2}]$, such that they solely depend on the parameters of the problem.
To our knowledge, this is the first work that studies the non-asymptotic convergence behavior of SG-based algorithms and their Polyak-Ruppert averaging in a streaming framework.
To do this, we assume for each $t\in\N$ the following about the (stochastic) gradients of $\{f_{t,i}\}_{i=1}^{n_{t}}$.
\begin{assumptionp}{}[unbiased gradients]\label{assump:measurable}
For each $\theta\in\Theta$, the random variable $\nabla_{\theta}f_{t,i}(\theta)$ is square-integrable and $\forall\theta\in\Theta$, $\nabla_{\theta}F(\theta)=\E[\nabla_{\theta}f_{t,i}(\theta)]$.
\end{assumptionp}
\begin{assumptionp}{-p}[$C_{f}$-expected smoothness]\label{assump:ssg:f_lipschitz}
For a positive integer $p$, there exists $C_{f}>0$ such that $\forall\theta\in\Theta$, $\E [\rVert\nabla_{\theta}f_{t,i}(\theta)-\nabla_{\theta}f_{t,i}(\theta^{*})\rVert^{p}]\leq C_{f}^{p}\E[\lVert\theta-\theta^{*}\rVert^{p}]$.
\end{assumptionp}
\begin{assumptionp}{-p}[$\sigma$-gradient noise]\label{assump:ssg:sigma}
For a positive integer $p$, there exists $\sigma>0$ such that $\E[\rVert\nabla_{\theta}f_{t,i}(\theta^{*})\rVert^{p}]\leq\sigma^{p}$.
\end{assumptionp}

\textbf{Discussion of \cref{assump:measurable,assump:ssg:f_lipschitz,assump:ssg:sigma}.} These assumptions are standard for analyzing stochastic approximation/optimization problems with SG algorithms, e.g., see \citet{benveniste2012adaptive,kushner2003stochastic}.
\Cref{assump:measurable} concerns the access to unbiased stochastic approximations of the gradient $\nabla_{\theta}F$, which are common when SG algorithms are used to solve problem \cref{eq:so_problem}.\footnote{The principles for biased gradients are rather different, e.g., see \citep{d2008smooth,schmidt2011convergence}.}
Another common assumption for SG algorithms is that they are uniformly bounded.
But such an assumption is often too restrictive, as it can only hold for some loss functions \citep{bottou2018optimization,godichon2021convergence}.
Instead, we make the weaker expected smoothness assumption of the gradients of $\{f_{t,i}\}_{i=1}^{n_{t}}$ in \cref{assump:ssg:f_lipschitz} \citep{gower2019sgd,bach2011non}.
The last key assumption concerns the finiteness of the gradient noise $\{f_{t,i}\}_{i=1}^{n_{t}}$ at $\theta^{*}$ (\cref{assump:ssg:sigma}).
It is worth noting that \cref{assump:ssg:f_lipschitz,assump:ssg:sigma} can be verified explicitly, e.g., see \citet{gower2019sgd}. 
For SSG and PSSG, \cref{assump:ssg:f_lipschitz,assump:ssg:sigma} only needs to hold for $p=2$, where for ASSG and APSSG, we need $p=4$ to bound the fourth order moment.

\subsection{Stochastic streaming gradients} \label{sec:nonasymptotic:ssg}

In this section, we analysis the SSG and PSSG algorithms from \cref{eq:streaming_grad,eq:ssg:proj}.
To do this, we first derive an explicit upper bound on the $t$-th estimate of \cref{eq:streaming_grad,eq:ssg:proj} for any learning rate $(\gamma_{t})$ and time-varying mini-batch $(n_{t})$ using classical techniques from stochastic approximations \citep{benveniste2012adaptive,kushner2003stochastic}.
\begin{theorem}[SSG/PSSG] \label{thm:nonblock:streaming_grad_upper_bound}
Let $\delta_{t}=\E[\lVert\theta_{t}-\theta^{*}\rVert^{2}]$ for $\delta_{0}\geq0$, where $(\theta_{t})$ either follows the recursion in \cref{eq:streaming_grad} or \cref{eq:ssg:proj}.
Suppose \cref{assump:L_strong_convexity,assump:L_lipschitz,assump:measurable,assump:ssg:f_lipschitz,assump:ssg:sigma} hold with $p=2$.
Then, for any learning rate $(\gamma_{t})$ and time-varying mini-batch $(n_{t})$, we have
\begin{align} \label{eq:thm:nonblock:streaming_grad_upper_bound}
\delta_{t} 
\leq& \exp \left( - \mu \sum_{i=t/2}^{t} \gamma_{i} \right) \pi_{t}^{\delta}
+ \frac{2 \sigma^{2}}{\mu} \max_{t/2 \leq i \leq t} \frac{\gamma_{i}}{n_{i}},
\end{align}
with $\pi_{t}^{\delta} = \exp(4C_{f}^{2}\sum_{i=1}^{t}\gamma_{i}^{2}/n_{i})\exp(2C_{\nabla}^{2}\sum_{i=1}^{t}\mathbbm{1}_{\{n_{i}>1\}}\gamma_{i}^{2})(\delta_{0}+2\sigma^{2}/C_{f}^{2})$.
\end{theorem}
\textbf{Sketch of proof.} Under \cref{assump:L_strong_convexity,assump:L_lipschitz,assump:measurable,assump:ssg:f_lipschitz,assump:ssg:sigma} with $p=2$, we show that $(\delta_{t})$ (derived using \cref{eq:streaming_grad}) satisfies the recursive relation
\begin{align}\label{eq:lem:nonblock:streaming_grad_recursive_bound_grad}
\delta_{t} \leq [1-2\mu\gamma_{t}+(2C_{f}^{2}+(n_{t}-1)C_{\nabla}^{2})n_{t}^{-1}\gamma_{t}^{2}]\delta_{t-1} + 2\sigma^{2}n_{t}^{-1}\gamma_{t}^{2},
\end{align}
for any $(\gamma_{t})$ and $(n_{t})$ fulfilling the conditions imposed on the learning rate \citep{robbins1951stochastic}.
This recursive relation is then explicitly upper bounded in a non-asymptotic manner using \cref{prop:appendix:delta_recursive_upper_bound_nt_simpler} in \cref{sec:appendix}.
Bounding the projected estimate in \cref{eq:ssg:proj} follows directly from the fact that $\E[\lVert\mathcal{P}_{\Theta}(\theta)-\theta^{*}\rVert^{2}]\leq\E[\lVert\theta-\theta^{*}\rVert^{2}]$, $\forall\theta\in\Theta$ \citep{zinkevich2003online}.
Alternatively, the projected estimate can also be shown without \cref{assump:ssg:f_lipschitz,assump:ssg:sigma}, but instead with a bounded gradient assumption, e.g., see \citet{bach2011non}.

\textbf{Related work.} When $n_{t}=1$ in \cref{eq:thm:nonblock:streaming_grad_upper_bound}, we obtain (an online version of) the usual SG descent studied in \citet{bach2011non}. 
As mentioned in \citet{bach2011non}, \cref{thm:nonblock:streaming_grad_upper_bound} forms an upper bound on the function values, $\E[F(\theta_{t})-F(\theta^{*})]\leq C_{f}\delta_{t}/2$; this follows from the Cauchy-Schwarz inequality and \cref{assump:ssg:f_lipschitz}.

\textbf{Decay of the initial conditions.} 
The learning rate $(\gamma_{t})$ should satisfy the conditions $\sum_{i=1}^{t}\gamma_{i}=\infty$ and $\sum_{i=1}^{t}\gamma_{i}^{2}<\infty$ as $t\rightarrow\infty$ of \citet{robbins1951stochastic}.
These conditions directly imply that $\pi_{t}^{\delta}<\infty$.
Thus, our attention is on reducing the \emph{noise term} $\max_{t/2 \leq i \leq t}\gamma_{i}/n_{i}$ without damaging the natural decay of the \emph{sub-exponential term} $\exp(-\mu\sum_{i=t/2}^{t}\gamma_{i})$.
In particular, the non-asymptotic bound in \cref{eq:thm:nonblock:streaming_grad_upper_bound} holds for any learning rate fulfilling these conditions.
In addition, the scaling of $n_{t}$ in the noise term shows an obvious possibility of variance reduction.

Before considering time-varying mini-batches, we consider the constant case where $n_{t}$ follows the constant $C_{\rho}\in\N$, i.e., an online (projected) mini-batch SG variant.
\begin{corollary}[SSG/PSSG with constant mini-batches] \label{cor:nonblock:streaming_grad_upper_bound_constant}
Let $\delta_{t}=\E[\lVert\theta_{t}-\theta^{*}\rVert^{2}]$ for $\delta_{0}\geq0$, where $(\theta_{t})$ either follows the recursion in \cref{eq:streaming_grad} or \cref{eq:ssg:proj}.
Suppose \cref{assump:L_strong_convexity,assump:L_lipschitz,assump:measurable,assump:ssg:f_lipschitz,assump:ssg:sigma} hold with $p=2$.
Then, if $\gamma_{t}=C_{\gamma}n_{t}^{\beta}t^{-\alpha}$ with $n_{t}=C_{\rho}$, for $\alpha\in(1/2,1)$, we have
\begin{align}  \label{eq:sgd:bound:constant}
\delta_{t} 
\leq& \exp \left( - \frac{\mu C_{\gamma} N_{t}^{1-\alpha}}{2^{1-\alpha} C_{\rho}^{1-\alpha-\beta}} \right) 
 \pi_{\infty}^{c}
+ \frac{2^{1+\alpha} \sigma^{2} C_{\gamma}}{\mu C_{\rho}^{1-\alpha-\beta} N_{t}^{\alpha}},
\end{align}
where $\pi_{\infty}^{c} = \exp(4\alpha C_{\gamma}^{2}(2C_{f}^{2}+C_{\rho}\mathbbm{1}_{\{C_{\rho}>1\}}C_{\nabla}^{2})/(2\alpha-1)C_{\rho}^{1-2\beta})(\delta_{0}+2\sigma^{2}/C_{f}^{2})$ is a finite constant.
\end{corollary}
\textbf{Decay of the initial conditions.} The bound in \cref{cor:nonblock:streaming_grad_upper_bound_constant} depends on the initial condition $\delta_{0} =   \lVert \theta_{0} - \theta^{*} \rVert^{2} $ and the variance $\sigma^{2}$ in the noise term.
The initial condition $\delta_{0}$ vanish sub-exponentially fast for $\alpha \in (1/2,1)$; the condition of having $\alpha\in(1/2,1)$ is a natural restriction from \citet{robbins1951stochastic}. 
Thus, the asymptotic term is $2^{1+\alpha} \sigma^{2} C_{\gamma}/\mu C_{\rho}^{1-\alpha-\beta} N_{t}^{\alpha}$, i.e., $\delta_{t}=\mathcal{O}( N_{t}^{-\alpha})$.
Moreover, the bound in \cref{eq:sgd:bound:constant} is optimal (up to some constants) for quadratic functions $(f_{t,i})$, since the deterministic recursion in \cref{eq:lem:nonblock:streaming_grad_recursive_bound_grad} would be with equality.
It is worth noting that if $C_{\gamma}C_{f}$ or $C_{\gamma}C_{\nabla}$ is chosen too large, they may produce a large $\pi_{\infty}^{c}$ constant.
In addition, $\pi_{\infty}^{c}$ is positively affected by $C_{\rho}$ when $\beta<1/2$.
Obviously, the hyper-parameter $\beta$ only comes into play if the mini-batch size $C_{\rho}$ is larger than one, i.e., $C_{\rho}>1$.
Nonetheless, the effect of $\pi_{\infty}^{c}$ will decrease exponentially fast due to the sub-exponentially decaying factor in front.

\textbf{Variance reduction from larger mini-batches.} 
Not surprisingly, larger mini-batches $C_{\rho}$ cause a variance reducing effect, e.g., see the illustrations in \cref{sec:experiments}.
Nevertheless, \cref{eq:sgd:bound:constant} explicitly shows the variance reducing effect in each term, which can help us better understand how to optimally tune the learning rate.
In particular, the asymptotic term is divided by $C_{\rho}^{1-\alpha-\beta}$, implying we should take $\alpha+\beta\leq1$ when $C_{\rho}$ is large.
However, this must be done with moderation as larger mini-batches $C_{\rho}$ simultaneously damage the sub-exponential term.
Another important point from this is that mini-batches do not provide a better convergence rate, but simply scale, i.e., the slope of the rate of convergence is unchanged, but the intercept is lowered (e.g., see \cref{fig:lr:c}).

Having fixed size mini-batches is not the most realistic streaming framework, these mini-batches are much more likely to vary in size over time.
For the convenience of notation, let $\tilde{\rho} = \rho \mathbbm{1}_{\{\rho\geq0\}}$.
\begin{corollary}[SSG/PSSG with time-varying mini-batches] \label{cor:nonblock:learning_rate_tau_1}
Let $\delta_{t}=\E[\lVert\theta_{t}-\theta^{*}\rVert^{2}]$ for $\delta_{0}\geq0$, where $(\theta_{t})$ either follows the recursion in \cref{eq:streaming_grad} or \cref{eq:ssg:proj}.
Suppose \cref{assump:L_strong_convexity,assump:L_lipschitz,assump:measurable,assump:ssg:f_lipschitz,assump:ssg:sigma} hold with $p=2$.
Then, if $\gamma_{t}=C_{\gamma}n_{t}^{\beta}t^{-\alpha}$ with $n_{t}=\lceil C_{\rho}t^{\rho} \rceil$, for $\alpha-\beta\tilde{\rho}\in(1/2,1)$, we have
\begin{align}
\delta_{t} 
\leq& 
\exp \left( - \frac{\mu C_{\gamma} N_{t}^{1-\phi}}{2^{(2+\rho)(1-\phi)} C_{\rho}^{1-\beta-\phi}} \right) \pi_{\infty}^{v} + \frac{2^{1+(2+\rho)\phi} \sigma^{2} C_{\gamma}}{\mu C_{\rho}^{(1-\beta)\mathbbm{1}_{\{\rho\geq0\}} - \phi} N_{t}^{\phi}},
\end{align}
where $\phi=((1-\beta)\tilde{\rho}+\alpha)/(1+\tilde{\rho})$ and $\pi_{\infty}^{v}=\exp(4(\alpha-\beta\tilde{\rho})C_{\gamma}^{2}C_{\rho}^{2\beta}(2C_{f}^{2}+C_{\nabla}^{2})/(2(\alpha-\beta\tilde{\rho})-1))(\delta_{0}+2\sigma^{2}/C_{f}^{2})$ is a finite constant.
\end{corollary}
\textbf{Accelerated decay with increasing mini-batches.} As mentioned for \cref{cor:nonblock:streaming_grad_upper_bound_constant}, $\alpha-\beta\tilde{\rho}\in(1/2,1)$ is a natural condition from \citet{robbins1951stochastic}; this relaxes the condition of having $\alpha\in(1/2,1)$ for $\rho\geq0$. 
In particular, this shows that we can accelerate convergence by taking increasing mini-batches, e.g., taking $\alpha=2/3$ and $\beta=0$ yields $\delta_{t}=\mathcal{O}(N_{t}^{-(2/3+\rho)/(1+\rho)})$ when $\rho>0$.
Conversely, when $\rho<0$, we obtain the same decay as in \cref{cor:nonblock:streaming_grad_upper_bound_constant}, namely, $\delta_{t}=\mathcal{O}( N_{t}^{-\alpha })$.
These effects are illustrated in \cref{fig:lr:v1,fig:lr:v8,fig:lr:v64,fig:lr:v128} for $C_{\rho}\in\{1,8,64,128\}$.

\textbf{Variance reduction from larger mini-batches.} Similarly to \cref{cor:nonblock:streaming_grad_upper_bound_constant}, the sub-exponential and asymptotic term is scaled by $C_{\rho}^{1-\beta-\phi}$ for $\rho\geq0$, implying we should take $\alpha+\beta\leq1$ to obtain variance reduction. 
However, as discussed above, this variance reduction is only beneficial in the beginning and does not contribute to a better convergence rate (relative to the slope).
Thus, large mini-batch sizes $C_{\rho}$ and negative mini-batch rates $\rho$ will give (an initial) variance reduction but the same convergence rate as in \cref{cor:nonblock:streaming_grad_upper_bound_constant}.

\subsection{Polyak-Ruppert averaging} \label{sec:nonasymptotic:assg}

In what follows, we consider the Polyak-Ruppert averaging estimate $(\bar{\theta}_{n})$ given in \cref{eq:avg:streaming_grad_est}, where $(\theta_{t})$ follows the recursion in \cref{eq:streaming_grad} or \cref{eq:ssg:proj}.
Besides having \cref{assump:ssg:f_lipschitz,assump:ssg:sigma} to hold for $p=4$, additional assumptions are needed for bounding the Polyak-Ruppert averaging estimate.
First, we make an additional smoothness assumption on the objective function $F$.
\begin{assumptionp}{}[$C_{\nabla}'$-Lipschitz continuous Hessian operator] \label{avg:assump:L_lipschitz}
The function $F$ is twice differentiable with $C_{\nabla}'$-Lipschitz continuous Hessian operator $\nabla_{\theta}^{2}F$, meaning, there exists $C_{\nabla}'\geq0$ such that $\forall\theta\in\Theta$,
\begin{align} \label{eq:avg:assump:L_lipschitz}
\lVert\nabla_{\theta}^{2}F(\theta)-\nabla_{\theta}^{2}F(\theta^{*})\rVert \leq C_{\nabla}'\lVert\theta-\theta^{*}\rVert.
\end{align}
\end{assumptionp}
Next, in continuation of \cref{assump:ssg:sigma}, we make the following assumption about covariance of $(\nabla_{\theta}f_{t,i}(\theta^{*}))$, which we interpret as the sequence of score vectors with respect to the parameter vector $\theta^{*}$.
\begin{assumptionp}{}[Covariance of the scores]\label{assump:assg:L_lipschitz}
There exists a non-negative self-adjoint operator $\Sigma$ such that $\E[\nabla_{\theta}f_{t,i}(\theta^{*})\nabla_{\theta}f_{t,i}(\theta^{*})^\top]\preceq\Sigma$.
\end{assumptionp}
Note that the operator $\Sigma$ always exists when $\sigma$ is finite for order $p=4$ in \cref{assump:ssg:sigma}.

\subsubsection{Polyak-Ruppert averaging of stochastic streaming gradients (ASSG)} \label{sec:polyakruppert:unbounded}

As in \cref{sec:nonasymptotic:ssg}, we conduct a general study for any learning rate $(\gamma_{t})$ and time-varying mini-batch $(n_{t})$ when applying the Polyak-Ruppert averaging estimate $(\bar{\theta}_{n})$ from \cref{eq:avg:streaming_grad_est}, where $(\theta_{t})$ follows the recursion in \cref{eq:streaming_grad}, i.e., the ASSG.

\begin{theorem}[ASSG] \label{thm:avg:nonblock:streaming_grad_upper_bound}
Let $\bar{\delta}_{t}=\E[\lVert \bar{\theta}_{t}-\theta^{*}\rVert^{2}]$ with $(\bar{\theta}_{t})$ given by \cref{eq:avg:streaming_grad_est}, where $(\theta_{t})$ follows the recursion in \cref{eq:streaming_grad}.
Suppose \cref{assump:L_strong_convexity,assump:L_lipschitz,assump:measurable,assump:ssg:f_lipschitz,assump:ssg:sigma,avg:assump:L_lipschitz,assump:assg:L_lipschitz} hold with $p=4$.
Then, for any learning rate $(\gamma_t)$ and time-varying mini-batch $(n_{t})$, we can upper bound $\bar{\delta}_{t}^{1/2}$ by
\begin{align} \label{eq:avg:thm:nonblock:streaming_grad_upper_bound}
\frac{\Lambda^{1/2}}{N_{t}^{1/2}}
+\frac{1}{\mu N_{t}} \sum_{i=1}^{t-1} \left| \frac{n_{i+1}}{\gamma_{i+1}} - \frac{n_{i}}{\gamma_{i}} \right| \delta_{i}^{1/2}
+ \frac{n_{t}}{\mu \gamma_{t}N_{t}} \delta_{t}^{1/2} 
+ \frac{n_{1}}{\mu N_{t}} \left( \frac{1}{\gamma_{1}} + C_{f} \right) \delta_{0}^{1/2} 
+ \frac{C_{f}}{\mu N_{t}} \left( \sum_{i=1}^{t-1} n_{i+1} \delta_{i} \right)^{1/2}
+ \frac{C_{\nabla}'}{\mu N_{t}} \sum_{i=0}^{t-1} n_{i+1} \Delta_{i}^{1/2},
\end{align}
where $\Lambda=\operatorname{Tr}(\nabla_{\theta}^{2}F(\theta^{*})^{-1}\Sigma\nabla_{\theta}^{2}F(\theta^{*})^{-1})$ and $\Delta_{t}=\E[\lVert\theta_{t}-\theta^{*}\rVert^{4}]$ for some $\Delta_{0}\geq0$.
\end{theorem}
As noticed in \citet{polyak1992acceleration}, the leading term $\Lambda/N_{t}$ achieves the Cramer-Rao lower bound \citep{murata1999statistical,gadat2023optimal}.
Note that the leading term $\Lambda/N_{t}$ is invariant of the learning rate $(\gamma_{t})$ and the time-varying mini-batches $(n_{t})$.
Moreover, the bound is $\mathcal{O}(N_{t}^{-1})$ without inverting the Hessian.
Next, the processes $(\delta_{t})$ and $(\Delta_{t})$ can be bounded by the recursive relations in \cref{eq:thm:nonblock:streaming_grad_upper_bound,eq:fourth_order_moment_upper_bound_general}.
There are no sub-exponential decaying terms for the initial conditions in \cref{thm:avg:nonblock:streaming_grad_upper_bound}, which is a common problem for averaging. 
However, as mentioned previously, we are more interested in advancing the decay of the asymptotic terms.
To ease notation, we make use of the functions $\psi_{x}^{y}(t) : \R \rightarrow \R$, given as
\begin{align*}
\psi_{x}^{y} (t) =
\begin{cases}
   t^{(1-x)/(1+y)}/(1-x) &\text{if } x < 1, \\
   (1+y)\log(t) &\text{if } x = 1 ,\\
   x/(x-1) &\text{if } x > 1,
  \end{cases}
\end{align*} 
with $y\in\R_{+}$, such that $\sum_{i=1}^{t}i^{-x}\leq\psi_{x}^{0}(t)$ for any $x\in\R_{+}$. 
Note that $\psi_{x}^{y}(t)/t=\mathcal{O}(t^{-(x+y)/(1+y)})$ if $x<1$, $\psi_{x}^{y}(t)/t=\mathcal{O}(\log(t)t^{-1})$ if $x=1$, and $\psi_{x}^{y}(t)/t=\mathcal{O}(t^{-1})$ if $x>1$. 
Hence, for any $x,y\in\R_{+}$, $\psi_{x}^{y}(t)/t=\tilde{\mathcal{O}}(t^{-(x+y)/(1+y)})$, where the $\tilde{\mathcal{O}}(\cdot)$ notation hides logarithmic factors.
\begin{corollary}[ASSG with constant mini-batches] \label{cor:avg:streaming_grad_upper_bound_constant}
Let $\bar{\delta}_{t}=\E[\lVert \bar{\theta}_{t}-\theta^{*}\rVert^{2}]$ with $(\bar{\theta}_{t})$ given by \cref{eq:avg:streaming_grad_est}, where $(\theta_{t})$ follows the recursion in \cref{eq:streaming_grad}.
Suppose \cref{assump:L_strong_convexity,assump:L_lipschitz,assump:measurable,assump:ssg:f_lipschitz,assump:ssg:sigma,avg:assump:L_lipschitz,assump:assg:L_lipschitz} hold with $p=4$.
Then, if $\gamma_{t} = C_{\gamma}n_{t}^{\beta} t^{-\alpha}$ with $n_{t}=C_{\rho}$, for $\alpha\in(1/2,1)$, we have
\begin{align*}
\bar{\delta}_{t}^{1/2} 
\leq& \frac{\Lambda^{1/2}}{N_{t}^{1/2}}  
+ \frac{6 \sigma C_{\rho}^{(1-\alpha-\beta)/2} }{\mu^{3/2}C_{\gamma}^{1/2}N_{t}^{1-\alpha /2}}
+ \frac{2^{\alpha} 6 C_{\nabla}' \sigma^{2} C_{\gamma}}{\mu^{2}C_{\rho}^{1-\alpha-\beta}N_{t}^{\alpha}}
+ \frac{2 C_{f} \sigma C_{\gamma}^{1/2}}{\mu^{3/2}C_{\rho}^{(1-\alpha-\beta)/2}N_{t}^{(1+\alpha)/2}}
+ \frac{C_{\rho} \Gamma_{c}}{\mu N_{t}} 
\\ &+ \frac{C_{\rho}^{2-\alpha-\beta} \sqrt{\pi_{\infty}^{c}} A_{\infty}^{c}}{\mu C_{\gamma}N_{t}^{2-\alpha}}
+ \frac{(6+7 \mathbbm{1}_{\{C_{\rho} > 1\}}) 2^{3\alpha/2} C_{\nabla}'  \sigma^{2}C_{\gamma}^{3/2}C_{\rho}^{3\beta/2}\psi_{3\alpha/2}^{0}(N_{t}/C_{\rho})}{\mu^{3/2}N_{t}},
\end{align*}
with $\Gamma_{c}$ given by $( 1/C_{\gamma} C_{\rho}^{\beta} + C_{f}) \delta_{0}^{1/2} + C_{f} \sqrt{\pi_{\infty}^{c}A_{\infty}^{c}/C_{\rho}} +\sqrt{\pi_{\infty}^{c}} A_{\infty}^{c}/C_{\gamma} C_{\rho}^{\beta} + C_{\nabla}' \sqrt{\Pi_{\infty}^{c}} A_{\infty}^{c}$, consisting of the finite constants $\pi_{\infty}^{c}$, $\Pi_{\infty}^{c}$ and $A_{\infty}^{c}$, that only depends on $\mu$, $\delta_{0}$, $\Delta_{0}$, $C_{f}$, $\sigma$, $C_{\nabla}$, $C_{\nabla}'$, $C_{\gamma}$, $C_{\rho}$, $\beta$ and $\alpha$.
\end{corollary}
\textbf{Accelerated decay the initial conditions.} By averaging, we have increased the rate of convergence from $\mathcal{O} (N_{t}^{-\alpha})$ to the optimal rate $\mathcal{O} (N_{t}^{-1})$ (when we compare to SSG with constant mini-batches in \cref{cor:nonblock:streaming_grad_upper_bound_constant}).
The two subsequent terms are the main remaining terms decaying at the rate $\mathcal{O} (N_{t}^{\alpha-2})$ and $\mathcal{O} (N_{t}^{-2\alpha})$, which suggest taking $\alpha=2/3$.
The remaining terms are negligible.
Next, it is worth noting that having $\alpha+\beta=1$ in \cref{cor:avg:streaming_grad_upper_bound_constant}, we would give no impact in the main remaining terms from the mini-batch size $C_{\rho}$.
At last, as we do not rely on sub-exponentially decaying terms, we need to be more careful when picking our hyper-parameters, e.g., taking $C_{\gamma} C_{f}$ too large may cause $\Gamma_{c}$ to be significant.
Nevertheless, the term consisting of $\Gamma_{c}$ decay at a rate of at least $\mathcal{O} (N_{t}^{-2})$.
\begin{corollary}[ASSG with time-varying mini-batches] \label{cor:avg:streaming_grad_upper_bound_increasing}
Let $\bar{\delta}_{t}=\E[\lVert \bar{\theta}_{t}-\theta^{*}\rVert^{2}]$ with $(\bar{\theta}_{t})$ given by \cref{eq:avg:streaming_grad_est}, where $(\theta_{t})$ follows the recursion in \cref{eq:streaming_grad}.
Suppose \cref{assump:L_strong_convexity,assump:L_lipschitz,assump:measurable,assump:ssg:f_lipschitz,assump:ssg:sigma,avg:assump:L_lipschitz,assump:assg:L_lipschitz} hold with $p=4$.
Then, if $\gamma_{t}=C_{\gamma} n_{t}^{\beta} t^{-\alpha}$ with $n_{t} = \lceil C_{\rho} t^{\rho} \rceil$, for $\alpha -\beta \tilde{\rho} \in (1/2,1)$, we have
\begin{align*}
\bar{\delta}_{t}^{1/2}
\leq& \frac{\Lambda^{1/2}}{N_{t}^{1/2}}
+ \frac{2^{3+\phi(1+\tilde{\rho})} \sigma C_{\rho}^{(1-\phi-\beta)/2\mathbbm{1}_{\{\rho\geq0}\}}}{\mu^{3/2}C_{\gamma}^{1/2}N_{t}^{1-\phi/2}}
+\frac{2^{(1+\phi)(1+\tilde{\rho})-2} C_{\nabla}' \sigma^{2} C_{\gamma}}{\mu^{2} C_{\rho}^{1-\phi-\beta} N_{t}^{\phi}}
+ \frac{2^{\phi(1+\tilde{\rho})/2} C_{f}\sigma C_{\gamma}^{1/2}}{\mu^{3/2} C_{\rho}^{(1-\phi-\beta)/2\mathbbm{1}_{\{\rho\geq0\}}} N_{t}^{(1+\phi)/2}}
+\frac{C_{\rho} \Gamma_{v}}{\mu N_{t}}
\\ &+ \frac{C_{\rho}^{2-\phi-\beta} \sqrt{\pi_{\infty}^{v}} A_{\infty}^{v}}{\mu C_{\gamma} N_{t}^{2-\phi}}
+ \frac{2^{3(1+\phi)(1+\tilde{\rho})/2} C_{\nabla}' \sigma^{2} C_{\gamma}^{3/2} C_{\rho}^{1+3\beta/2}\psi_{3(\alpha-\beta\tilde{\rho})/2}^{\tilde{\rho}}(N_{t}/C_{\rho})}{\mu^{3/2} C_{\rho}^{\mathbbm{1}_{\{\rho\geq0}\}}N_{t}},
\end{align*}
with $\Gamma_{v}$ given by $( 1/C_{\gamma} C_{\rho}^{\beta} + C_{f}) \delta_{0}^{1/2} + 2^{\tilde{\rho}} C_{f} \sqrt{\pi_{\infty}^{v}A_{\infty}^{v}/C_{\rho}} + 2 \sqrt{\pi_{\infty}^{v}} A_{\infty}^{v}/C_{\gamma} C_{\rho}^{\beta} + 2^{\tilde{\rho}} C_{\nabla}' \sqrt{\Pi_{\infty}^{v}} A_{\infty}^{v}$, consisting of the finite constants $\pi_{\infty}^{v}$, $\Pi_{\infty}^{v}$ and $A_{\infty}^{v}$, that only depends on $\mu$, $\delta_{0}$, $\Delta_{0}$, $C_{f}$, $\sigma$, $C_{\nabla}$, $C_{\nabla}'$, $C_{\gamma}$, $C_{\rho}$, $\beta$ and $\alpha$.
\end{corollary}
\textbf{Robustness towards mini-batch rate $\rho$:} Following the arguments above, the two main remainder terms suggest that $\phi = 2/3 \Leftrightarrow \alpha-\beta\tilde{\rho} = (2-\tilde{\rho})/3$, e.g., by setting $\beta=0$, we should pick $\alpha = (2-\tilde{\rho})/3$.
Likewise, if $\rho=0$, we yield the same conclusion as in \cref{cor:avg:streaming_grad_upper_bound_constant}, namely $\alpha=2/3$.
However, these hyper-parameter choices are not resilient against any time-varying streaming rate $\rho$.
Nonetheless, we can \emph{robustly} achieve $\phi=2/3$ for any $\rho \in (-1,1)$ by setting $\alpha=2/3$ and $\beta=1/3$.
In other words, we can achieve the same convergence for any time-varying mini-batch rate by having $\alpha=2/3$ and $\beta=1/3$; this is illustrated in \cref{fig:lr:v8:phi,fig:gm:v8:phi}.

\subsubsection{Polyak-Ruppert averaging of projected stochastic streaming gradients (APSSG)} \label{sec:assg:bounded}

In this section, we analyze the projected Polyak-Ruppert averaging estimate (a.k.a. APSSG), where $(\theta_{t})$ follows the recursion in \cref{eq:ssg:proj}.
To avoid calculating the six-order moment, we make the unnecessary assumption that $\lVert\nabla_{\theta}f_{t,i}(\theta)\rVert$ is uniformly bounded on $\Theta$; the derivation of the six-order moment can be found in \citet{godichon2016estimating}.
\begin{assumptionp}{}[$G_{\Theta}$-bounded stochastic gradients]\label{assump:assg:bounded}
Let $D_{\theta} = \inf_{\theta\in\partial\Theta} \lVert\theta-\theta^{*}\rVert>0$ with $\partial\Theta$ denoting the frontier of $\Theta$.
Assume there exists $G_{\Theta}>0$ such that $\forall t \geq 1$, $\sup_{\theta\in\Theta} \lVert \nabla_{\theta} f_{t,i}(\theta)\rVert^{2} \leq G_{\Theta}^{2}$ a.s., with $i=1,\dots,n_{t}$.
\end{assumptionp}
\begin{corollary}[APSSG with constant mini-batches] \label{cor:assg:bounded:constant}
Let $\bar{\delta}_{t}=\E[\lVert \bar{\theta}_{t}-\theta^{*}\rVert^{2}]$ with $(\bar{\theta}_{t})$ given by \cref{eq:avg:streaming_grad_est}, where $(\theta_{t})$ follows the recursion in \cref{eq:ssg:proj}.
Suppose \cref{assump:L_strong_convexity,assump:L_lipschitz,assump:measurable,assump:ssg:f_lipschitz,assump:ssg:sigma,avg:assump:L_lipschitz,assump:assg:L_lipschitz,assump:assg:bounded} hold with $p=4$.
Then, if $\gamma_{t}=C_{\gamma} n_{t}^{\beta} t^{-\alpha}$ with $n_{t}=C_{\rho}$, for $\alpha\in(1/2,1)$, we have
\begin{align*}
\bar{\delta}_{t}^{1/2} 
\leq& \frac{\Lambda^{1/2}}{N_{t}^{1/2}}  
+ \frac{6 \sigma C_{\rho}^{(1-\alpha-\beta)/2} }{\mu^{3/2}C_{\gamma}^{1/2}N_{t}^{1-\alpha /2}}
+ \frac{2^{\alpha} 6 C_{\nabla}' \sigma^{2} C_{\gamma}}{\mu^{2}C_{\rho}^{1-\alpha-\beta}N_{t}^{\alpha}}
+ \frac{2 C_{f} \sigma C_{\gamma}^{1/2}}{\mu^{3/2}C_{\rho}^{(1-\alpha-\beta)/2}N_{t}^{(1+\alpha)/2}}
+ \frac{C_{\rho} \Gamma_{c}}{\mu N_{t}} 
\\ &+ \frac{C_{\rho}^{2-\alpha-\beta} \sqrt{\pi_{\infty}^{c}} A_{\infty}^{c}}{\mu C_{\gamma}N_{t}^{2-\alpha}}
+ \frac{(6+7 \mathbbm{1}_{\{C_{\rho} > 1\}}) 2^{3\alpha/2} C_{\nabla}'' \sigma^{2}C_{\gamma}^{3/2}C_{\rho}^{3\beta/2}\psi_{3\alpha/2}^{0}(N_{t}/C_{\rho})}{\mu^{3/2}N_{t}},
\end{align*}
with $C_{\nabla}''=C_{\nabla}'+2^{2}G_{\Theta}/D_{\theta}^{2}$ and $\Gamma_{c}$ given by $(1/C_{\gamma} C_{\rho}^{\beta} + C_{f}) \delta_{0}^{1/2} + C_{f} \sqrt{\pi_{\infty}^{c}A_{\infty}^{c}/C_{\rho}} +\sqrt{\pi_{\infty}^{c}} A_{\infty}^{c}/C_{\gamma} C_{\rho}^{\beta} + C_{\nabla}' \sqrt{\Pi_{\infty}^{c}} A_{\infty}^{c}$, consisting of the finite constants $\pi_{\infty}^{c}$, $\Pi_{\infty}^{c}$ and $A_{\infty}^{c}$, that only depends on $\mu$, $\delta_{0}$, $\Delta_{0}$, $C_{f}$, $\sigma$, $C_{\nabla}$, $C_{\nabla}'$, $C_{\gamma}$, $C_{\rho}$, $\beta$ and $\alpha$.
\end{corollary}
\begin{corollary}[APSSG with time-varying mini-batches] \label{cor:assg:bounded:varying}
Let $\bar{\delta}_{t}=\E[\lVert \bar{\theta}_{t}-\theta^{*}\rVert^{2}]$ with $(\bar{\theta}_{t})$ given by \cref{eq:avg:streaming_grad_est}, where $(\theta_{t})$ follows the recursion in \cref{eq:ssg:proj}.
Suppose \cref{assump:L_strong_convexity,assump:L_lipschitz,assump:measurable,assump:ssg:f_lipschitz,assump:ssg:sigma,avg:assump:L_lipschitz,assump:assg:L_lipschitz,assump:assg:bounded} hold with $p=4$.
Then, if $\gamma_{t}=C_{\gamma}n_{t}^{\beta}t^{-\alpha}$ with $n_{t}=\lceil C_{\rho}t^{\rho}\rceil$, for $\alpha-\beta\tilde{\rho}\in(1/2,1)$, we have
\begin{align*}
\bar{\delta}_{t}^{1/2}
\leq& \frac{\Lambda^{1/2}}{N_{t}^{1/2}}
+ \frac{2^{3+\phi(1+\tilde{\rho})} \sigma C_{\rho}^{(1-\phi-\beta)/2\mathbbm{1}_{\{\rho\geq0\}}}}{\mu^{3/2}C_{\gamma}^{1/2}N_{t}^{1-\phi/2}}
+\frac{2^{(1+\phi)(1+\tilde{\rho})-2} C_{\nabla}' \sigma^{2} C_{\gamma}}{\mu^{2} C_{\rho}^{1-\phi-\beta} N_{t}^{\phi}}
+ \frac{2^{\phi(1+\tilde{\rho})/2} C_{f}\sigma C_{\gamma}^{1/2}}{\mu^{3/2} C_{\rho}^{(1-\phi-\beta)/2\mathbbm{1}_{\{\rho\geq0\}}} N_{t}^{(1+\phi)/2}}
+\frac{C_{\rho} \Gamma_{v}}{\mu N_{t}}
\\ &+ \frac{C_{\rho}^{2-\phi-\beta} \sqrt{\pi_{\infty}^{v}} A_{\infty}^{v}}{\mu C_{\gamma} N_{t}^{2-\phi}}
+ \frac{2^{3(1+\phi)(1+\tilde{\rho})/2} C_{\nabla}'' \sigma^{2} C_{\gamma}^{3/2} C_{\rho}^{1+3\beta/2}\psi_{3(\alpha-\beta\tilde{\rho})/2}^{\tilde{\rho}}(N_{t}/C_{\rho})}{\mu^{3/2} C_{\rho}^{\mathbbm{1}_{\{\rho\geq0\}}}N_{t}},
\end{align*}
with $C_{\nabla}''=C_{\nabla}'+2^{2}G_{\Theta}/D_{\theta}^{2}$ and $\Gamma_{v}$ given by $( 1/C_{\gamma} C_{\rho}^{\beta} + C_{f}) \delta_{0}^{1/2} + 2^{\tilde{\rho}} C_{f} \sqrt{\pi_{\infty}^{v}A_{\infty}^{v}/C_{\rho}} + 2 \sqrt{\pi_{\infty}^{v}} A_{\infty}^{v}/C_{\gamma} C_{\rho}^{\beta} + 2^{\tilde{\rho}} C_{\nabla}' \sqrt{\Pi_{\infty}^{v}} A_{\infty}^{v}$, consisting of the finite constants $\pi_{\infty}^{v}$, $\Pi_{\infty}^{v}$ and $A_{\infty}^{v}$, that only depends on $\mu$, $\delta_{0}$, $\Delta_{0}$, $C_{f}$, $\sigma$, $C_{\nabla}$, $C_{\nabla}'$, $C_{\gamma}$, $C_{\rho}$, $\beta$ and $\alpha$.
\end{corollary}

\section{Experiments} \label{sec:experiments}

In this section, we demonstrate the theoretical results presented in \cref{sec:nonasymptotic} for various time-varying mini-batches.
The performance is measured over one-hundred replications of the quadratic mean error, i.e., $(\E[\lVert\theta_{N_{t}}-\theta^{*}\rVert^{2}])_{t\geq0}$ and $(\E[\lVert\bar{\theta}_{N_{t}}-\theta^{*}\rVert^{2}])_{t\geq0}$.
Note that averaging over several replications gives a reduction in variability, which mainly benefits the SSG and PSSG.
All metrics are shown in log-scale and normalized such that the first iteration is one, namely, $\E[\lVert\theta_{0}-\theta^{*}\rVert^{2}]=\E[\lVert\bar{\theta}_{0}-\theta^{*}\rVert^{2}]=1$.

\subsection{Linear regression} \label{sec:experiments:lr}

We continue the generic notation from \cref{sec:sub:problem_formulation}, where the linear regression is defined by $y=x^{T}\theta+\epsilon$, where $y\in\R$ is the measure, $x\in\R^{d}$ is a random feature vector, $\theta\in\R^{d}$ is the parameters vector, and $\epsilon$ is a random variable with zero mean, and $x$ and $ \epsilon$ are independent and identically distributed.
Thus, $\theta^{*}$ is the minimizer of $F(\theta)=\E[(y-x^{T}\theta)^{2}]$.
In this example, we fix $d=10$, set $\theta=(-4,-3,2,1,0,1,2,3,4,5)^{T} \in \R^{10}$, and let $x$ and $\epsilon$ be standard Gaussian.
It is well-known that $C_{\gamma}$ can substantially impact convergence; when $C_{\gamma}$ is too large, instability can occur, leading to an explosion during the first iterations. If $C_{\gamma}$ is too small, the convergence can become very slow and destroy the desired learning rate.
To focus on the various time-varying mini-batches, we set $C_{\gamma}=1/2$ and $\alpha=2/3$.

\begin{figure}[t!]
\caption{Linear regression with learning rate $\gamma_{t}=C_{\gamma}n_{t}^{\beta}t^{-\alpha}$ and time-varying mini-batch $n_{t}=\lceil C_{\rho}t^{\rho} \rceil$. See \texorpdfstring{\cref{sec:experiments:lr}}{4.1} for details.}
\begin{subfigure}{.3\linewidth}
  \centering
  \caption{Constant mini-batches with $\{\rho=0,\beta=0\}$}
  \includegraphics[width=1.0\linewidth]{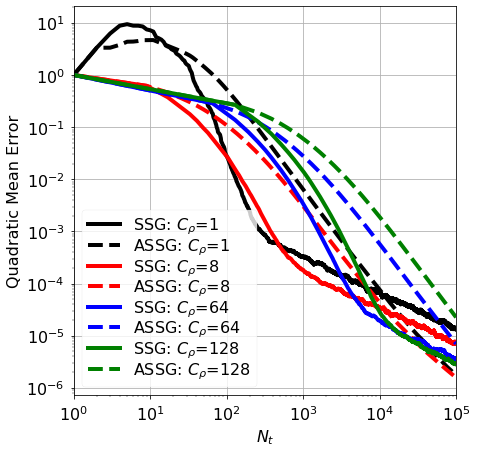}
  \label{fig:lr:c}
\end{subfigure}
\begin{subfigure}{.3\linewidth}
  \centering
  \caption{Time-varying mini-batches with $\{C_{\rho}=1,\beta=0\}$}
  \includegraphics[width=1.0\linewidth]{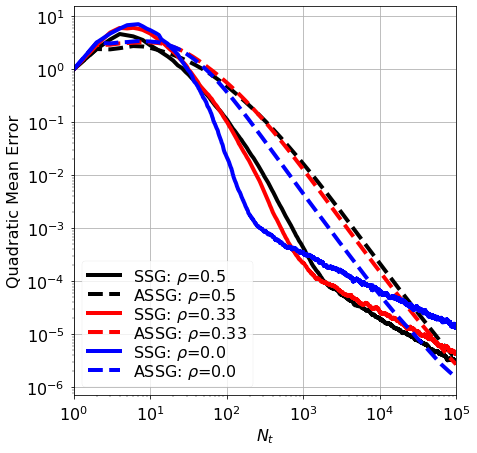}
  \label{fig:lr:v1}
\end{subfigure}
\begin{subfigure}{.3\linewidth}
  \centering
  \caption{Time-varying mini-batches with $\{C_{\rho}=8,\beta=0\}$}
  \includegraphics[width=1.0\linewidth]{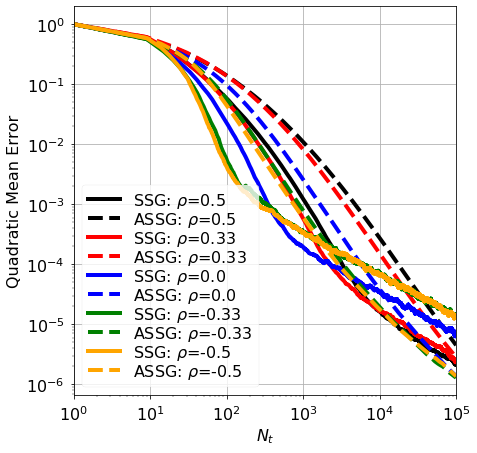}
  \label{fig:lr:v8}
\end{subfigure}
\newline
\begin{subfigure}{.3\linewidth}
  \centering
  \caption{Time-varying mini-batches with $\{C_{\rho}=64,\beta=0\}$}
  \includegraphics[width=1.0\linewidth]{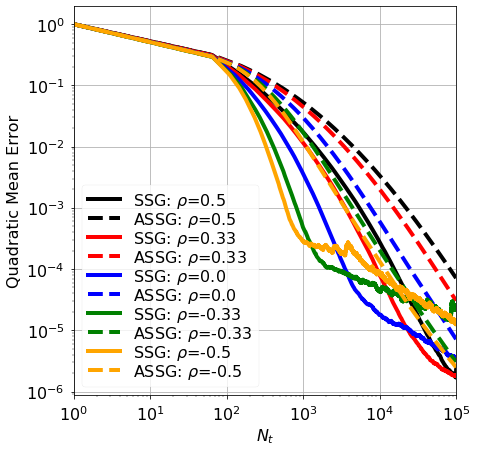}
  \label{fig:lr:v64}
\end{subfigure}
\begin{subfigure}{.3\linewidth}
  \centering
  \caption{Time-varying mini-batches with $\{C_{\rho}=128,\beta=0\}$}
  \includegraphics[width=1.0\linewidth]{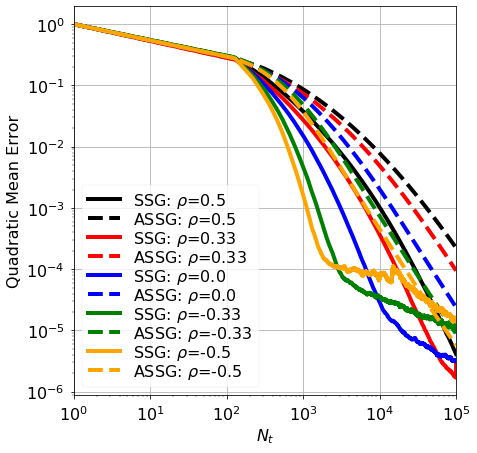}
  \label{fig:lr:v128}
\end{subfigure}
\begin{subfigure}{.3\linewidth}
  \centering
  \caption{Time-varying mini-batches with $\{C_{\rho}=8,\beta=1/3\}$}
  \includegraphics[width=1.0\linewidth]{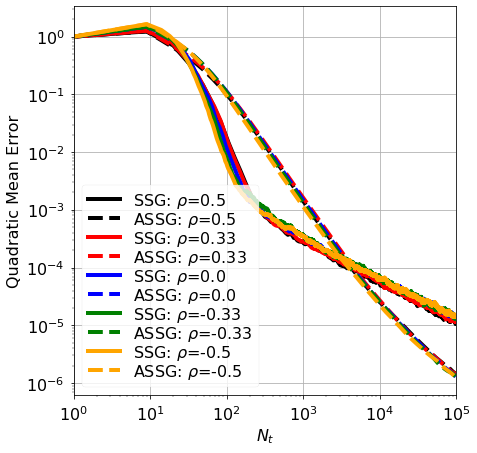}
  \label{fig:lr:v8:phi}
\end{subfigure}
\label{fig:lr}
\end{figure}

\textbf{Discussion.} In \cref{fig:lr:c}, we consider constant mini-batches to illustrate the results in \cref{cor:nonblock:streaming_grad_upper_bound_constant,cor:avg:streaming_grad_upper_bound_constant}.
This figure show a solid decay rate proportional to $\alpha=2/3$ for any mini-batch size $C_{\rho} \in \{1, 8, 64, 128\}$ with $\beta=0$, as shown in \cref{cor:nonblock:streaming_grad_upper_bound_constant}.
In particular, the mini-batches does not provide better convergence rates, but simply
scales the error, i.e. the slope of the rate of convergence is unchanged, but the intercept is lowered.
As explained after \cref{cor:avg:streaming_grad_upper_bound_constant}, we see an acceleration in decay by averaging.
Both algorithms show a noticeable reduction in variance when $C_{\rho}$ increases which are particularly beneficial in the beginning.
Next, in \cref{fig:lr:v1,fig:lr:v8,fig:lr:v64,fig:lr:v128}, we vary the mini-batch rate $\rho$ for (fixed) mini-batch sizes $C_{\rho}=1$, $8$, $64$, and $128$, respectively, with $\beta=0$.
These figures shows an increase in decay of the SSG when the mini-batch rate $\rho$ increase.
Despite this, we still achieve better convergence for the ASSG algorithm, which seems more immune to the different choices of mini-batch rate $\rho$, e.g., see the discussion after \cref{cor:avg:streaming_grad_upper_bound_increasing}.
We know this from \cref{cor:nonblock:learning_rate_tau_1}, as $\phi = (\tilde{\rho}+\alpha)/(1+\tilde{\rho}) \geq \alpha$ for $\beta=0$.
In addition, we see that $C_{\rho}$ has a positive effect on the noise (i.e., variance reduction), but if $C_{\rho}$ becomes too large, it may slow down convergence (as seen in \cref{fig:lr:v128}).
Alternatively, we could think around the problem in another way: how can we choose $\alpha$ and $\beta$ such that we have obtain decay of $\phi=2/3$ for any $\rho$. 
In other words, for any arrival schedule that may occur, how should we choose our hyper-parameters such that we achieve decay of $\phi=2/3$.
As discussed after \cref{cor:avg:streaming_grad_upper_bound_increasing}, one example of this could be achieved by setting $\alpha=2/3$ and $\beta=1/3$ such that $\phi=2/3$ for any $\rho$.
\Cref{fig:lr:v8:phi} shows an example of this where we (indeed) achieve the same decay rate for any mini-batch rate $\rho$.

\subsection{Geometric median} \label{sec:experiments:gm}

Robust estimators such as the geometric median may be preferred over the mean when the data is noisy; the geometric median is a generalization of the real median introduced by \citet{haldane1948note}.
In addition, SG-based algorithms are preferred in our streaming framework, as they can process large samples of high-dimensional data efficiently \citep{cardot2013efficient,godichon2016estimating,cardot2017online}.
The geometric median of $x\in\R^{d}$ is found by minimizing the objective $F(\theta)=\E[\lVert x-\theta\rVert-\lVert x\rVert]$ using gradients of the form $\nabla_{\theta}f(\theta)=-(x-\theta)/\lVert x-\theta\rVert$.
Properties of this geometric median, such as existence, uniqueness and robustness, can be found in, e.g., \citet{kemperman1987median,gervini2008robust}. 
Note that this objective function only possesses locally strong convexity properties \citep{cardot2013efficient}.
But by projecting the gradients, one could adapt the proof of \citet{gadat2023optimal} to a streaming setting.
Otherwise, if $x$ is bounded, one can adapt \citet{cardot2012fast} to the streaming setting showing that the estimates are bounded, and there is no use to project it in this case.
Similarly to \cref{sec:experiments:lr}, we fix $d=10$ and let $x$ be standard Gaussian centered at $\theta=(-4,-3,2,1,0,1,2,3,4,5)^{T} \in \R^{10}$.
Moreover, following the reasoning of \citet{cardot2013efficient}, we set $C_{\gamma}=\sqrt{d}=\sqrt{10}$, and let $\alpha=2/3$.

\begin{figure}[t!]
\caption{Geometric median with learning rate $\gamma_{t}=C_{\gamma}n_{t}^{\beta}t^{-\alpha}$ and time-varying mini-batch $n_{t}=\lceil C_{\rho}t^{\rho} \rceil$. See \texorpdfstring{\cref{sec:experiments:gm}}{4.2} for details.}
\begin{subfigure}{.3\linewidth}
  \centering
  \caption{Constant mini-batches with $\{\rho=0,\beta=0\}$}
  \includegraphics[width=1.0\linewidth]{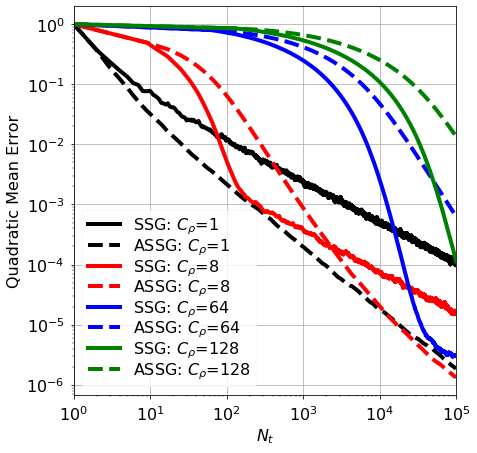}
  \label{fig:gm:c}
\end{subfigure}
\begin{subfigure}{.3\linewidth}
  \centering
  \caption{Time-varying mini-batches with $\{C_{\rho}=1,\beta=0\}$}
  \includegraphics[width=1.0\linewidth]{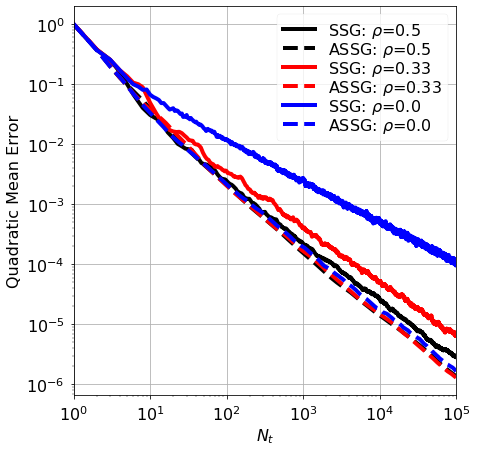}
  \label{fig:gm:v1}
\end{subfigure}
\begin{subfigure}{.3\linewidth}
  \centering
  \caption{Time-varying mini-batches with $\{C_{\rho}=8,\beta=0\}$}
  \includegraphics[width=1.0\linewidth]{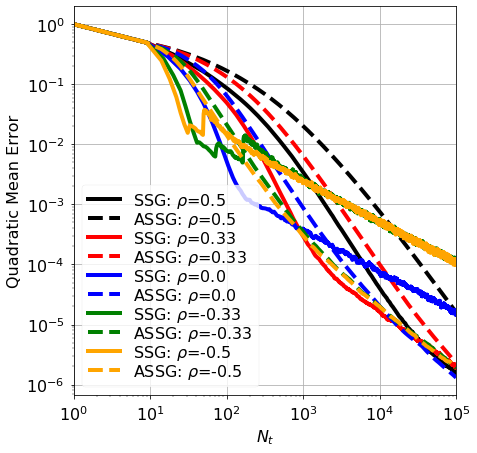}
  \label{fig:gm:v8}
\end{subfigure}
\newline
\begin{subfigure}{.3\linewidth}
  \centering
  \caption{Time-varying mini-batches $\{C_{\rho}=64,\beta=0\}$}
  \includegraphics[width=1.0\linewidth]{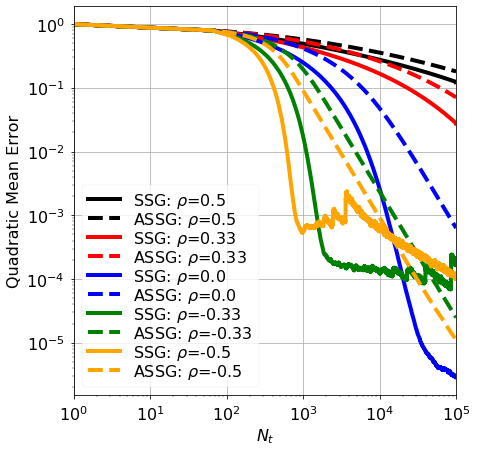}
  \label{fig:gm:v64}
\end{subfigure}
\begin{subfigure}{.3\linewidth}
  \centering
  \caption{Time-varying mini-batches with $\{C_{\rho}=128,\beta=0\}$}
  \includegraphics[width=1.0\linewidth]{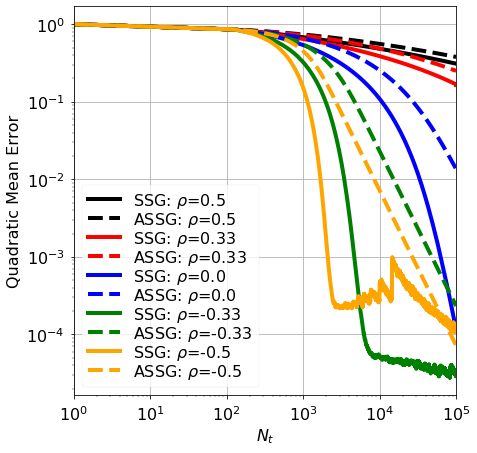}
  \label{fig:gm:v128}
\end{subfigure}
\begin{subfigure}{.3\linewidth}
  \centering
  \caption{Time-varying mini-batches with $\{C_{\rho}=8,\beta=1/3\}$}
  \includegraphics[width=1.0\linewidth]{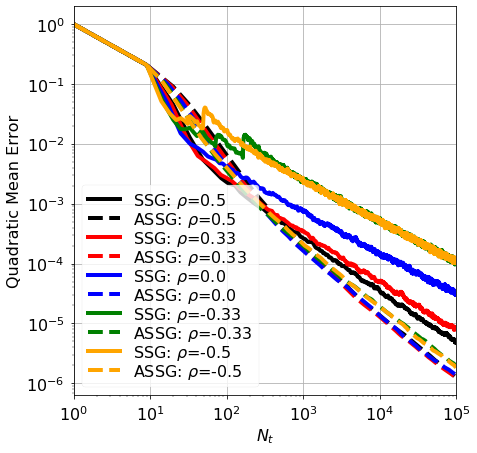}
  \label{fig:gm:v8:phi}
\end{subfigure}
\label{fig:gm}
\end{figure}

\textbf{Discussion.} \Cref{fig:gm:c} shows the variance reduction effect for different constant mini-batches $C_{\rho}$ with $\beta=0$.
However, the robustness of the geometric median leaves only a small positive impact for further variance reduction.
Thus, too large (constant) mini-batch sizes $C_{\rho}$ hinders the convergence as we make too few iterations.
These findings can be extended to \cref{fig:gm:v1,fig:gm:v8,fig:gm:v64,fig:gm:v128}, where we vary the mini-batch rate $\rho$ for mini-batch sizes $C_{\rho}=1$, $8$, $64$, and $128$, respectively, with $\beta=0$.
The lack of convergence improvements comes from $\beta=0$, which means we do not exploit the potential of using more observations to accelerate convergence.
As shown in \cref{fig:gm:v8:phi}, we can achieve this acceleration by simply taking $\beta=1/3$.
In addition, $\beta=1/3$ provides improved convergence robust to any mini-batch rate $\rho$.
Choosing a proper $\beta>0$ is particularly important when $C_{\rho}$ is large, as robustness is an integral part of the geometric median.

\section{Conclusions} \label{sec:conclusion}

We introduced a streaming framework for analyzing stochastic approximation/optimization problems.
This streaming framework was analogous to solving optimization problems using time-varying mini-batches that arrive sequentially.
We provided non-asymptotic convergence rates for different gradient-based algorithms; this included the famous Stochastic Gradient (SG) descent (a.k.a. Robbins-Monro algorithm), mini-batch SG, and time-varying mini-batch SG algorithms, as well as their iterated averages (a.k.a. Polyak-Ruppert averaging).
We showed how time-varying mini-batches together with Polyak-Ruppert averaging can provide variance reduction and accelerate convergence simultaneously.
We further demonstrated the beneficial effect of adapting learning to the time-varying mini-batches under different streaming settings.

\begin{figure}[t!]
\caption{WASSG with learning rate $\gamma_{t}=C_{\gamma}n_{t}^{\beta}t^{-\alpha}$ and time-varying mini-batch $n_{t}=\lceil C_{\rho}t^{\rho} \rceil$. See \texorpdfstring{\cref{sec:conclusion}}{5} for details.}
\begin{subfigure}{.3\linewidth}
  \centering
  \caption{Linear regression: Time-varying mini-batches with $\{C_{\rho}=8,\beta=1/3\}$}
  \includegraphics[width=1.0\linewidth]{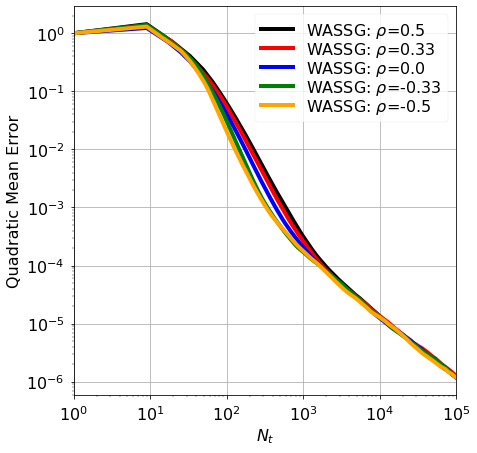}  
  \label{fig:lr:con}
\end{subfigure}
\begin{subfigure}{.3\linewidth}
  \centering
  \caption{Geometric median: Time-varying mini-batches with $\{C_{\rho}=8,\beta=1/3\}$}
  \includegraphics[width=1.0\linewidth]{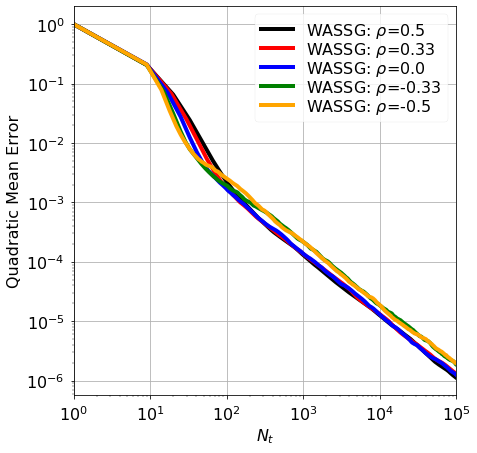}  
  \label{fig:gm:con}
\end{subfigure}
\label{fig:con}
\end{figure}

\textbf{Future perspectives.} There are several ways to expand our work: first, we can extend our analysis to include time-varying mini-batches of any size.
Second, many machine learning problems encounter correlated variables and high-dimensional data, thus an extension to non-strongly convex objectives would be advantageous \citep{bach2013non}, e.g., in \citet{werge2022adavol}, they use SG-based algorithms to make adaptive volatility predictions through optimization of the GARCH model.
Third, \cref{assump:measurable} requires unbiased (and independent) gradient estimates, thus, an obvious extension could incorporate a more realistic dependency assumption, thereby increasing the applicability.
Moreover, studying dependence may give insight into how to process dependent information \emph{optimally}.
Next, a natural extension would be to modify our Polyak-Ruppert averaging estimate from \cref{eq:avg:streaming_grad_est} to a weighted averaged version \citep{mokkadem2011generalization,boyer2022asymptotic}:
\begin{flalign} \label{eq:wasgd}
\textbf{(WASSG)} && \bar{\theta}_{t,\lambda} =& \frac{1}{\sum_{i=1}^{t} n_{i} \log(1+i)^{\lambda}} \sum_{i=1}^{t} n_{i} \log(1+i)^{\lambda} \theta_{i-1}, &&
\end{flalign}
for $\lambda>0$ with $(\theta_{t})$ following \cref{eq:streaming_grad} or \cref{eq:ssg:proj}.
One can limit the effect of bad initializations by placing more weight on the newest estimates.
Following the demonstrations in \cref{sec:experiments}, an example of this WASSG estimate $(\bar{\theta}_{t,\lambda})$ can be found in \cref{fig:con} with use of $\lambda=2$.
Here we see that although the WASSG estimate in \cref{eq:wasgd} may not achieve a better final error (compared to the ASSG and APSSG estimates in \cref{fig:lr:v8:phi,fig:gm:v8:phi}), it still achieves a better decay along the way, often referred to as \emph{parameter tracking}.

\begin{acknowledgement}
The authors thank the anonymous reviewer for the  valuable and helpful comments.
\end{acknowledgement}

\bibliographystyle{abbrvnat}
\bibliography{bibliography}

\appendix

\section{Proofs} \label{sec:proofs}

In this appendix, we provide detailed proofs of the results.
Purely technical results used in the proofs can be found in \cref{sec:appendix}.
Let $(\F_{t})_{t \geq 1}$ be an increasing family of $\sigma$-fields, namely $\F_{t}=\sigma(f_{1},\dots,f_{t})$ with $f_{t}:=\{f_{t,1},\dots,f_{t,n_{t}}\}$.
Furthermore, we expand the notation with $\mathcal{F}_{t-1,i} = \sigma ( f_{1,1} , \dots , f_{t-1, n_{t-1}}, f_{t,1} , \dots , f_{t,i} )$ such that $\mathcal{F}_{t-1,0} = \mathcal{F}_{t-1}$.
Meaning, $\forall 0 \leq i < j$, we have $\mathcal{F}_{t-1} \subseteq \mathcal{F}_{t-1,i} \subset \mathcal{F}_{t-1,j}$.
Thus, by the independence of the differentiable random functions $\{f_{t,i}\}$, \cref{assump:measurable} yields that $\forall t\geq1$, $\E[\nabla_{\theta}f_{t,i}(\theta_{t-1})\vert\F_{t-1,i-1}]=\nabla_{\theta}F(\theta_{t-1})$ with $i=1,\dots,n_{t}$.

\subsection{Proofs for \texorpdfstring{\cref{sec:nonasymptotic}}{2}}

The section is structured such that we start by analyzing the recursive relations and bounding them for every choice of learning rate $(\gamma_{t})$ and time-varying mini-batch $(n_{t})$. Next, we look at specific choices of $(\gamma_{t})$ and $(n_{t})$.

\begin{proof}[Proof of \cref{thm:nonblock:streaming_grad_upper_bound}]
Taking the quadratic norm on both sides of \cref{eq:streaming_grad}, expanding it, and take the conditional expectation, yields
\begin{align} \label{eq:lem:nonblock:sg_norm_fsigma}
\E[\lVert\theta_{t}-\theta^{*}\rVert^{2}\vert\F_{t-1}] = \lVert\theta_{t-1}-\theta^{*}\rVert^{2} + \frac{\gamma_{t}^{2}}{n_{t}^{2}} \E\left[\left\lVert\sum_{i=1}^{n_{t}} \nabla_{\theta} f_{t,i} \left( \theta_{t-1}\right)\right\rVert^{2} \middle\vert\F_{t-1}\right] - \frac{2\gamma_{t}}{n_{t}} \sum_{i=1}^{n_{t}}\E[\langle\nabla_{\theta}f_{t,i}(\theta_{t-1}),\theta_{t-1}-\theta^{*}\rangle\vert\F_{t-1}]. 
\end{align}
To bound the second term (on the right-hand side) of \cref{eq:lem:nonblock:sg_norm_fsigma}, we first expand it as follows,
\begin{align}
\sum_{i=1}^{n_{t}}\E[\lVert\nabla_{\theta}f_{t,i}(\theta_{t-1})\rVert^{2}\vert \F_{t-1}] + \sum_{i\neq j}^{n_{t}}\E[\langle\nabla_{\theta}f_{t,i}(\theta_{t-1}), \nabla_{\theta}f_{t,j}(\theta_{t-1})\rangle\vert\F_{t-1}]. \label{eq:ssg:proof20}
\end{align}
For first term of \cref{eq:ssg:proof20}, we utilize the Lipschitz continuity of $\nabla_{\theta} f_{t,i}$, together with \cref{assump:measurable,assump:ssg:f_lipschitz,assump:ssg:sigma}, to obtain
\begin{align}
\E[\lVert\nabla_{\theta}f_{t,i}(\theta_{t-1})\rVert^{2}\vert\F_{t-1}] 
& \leq 2\E[\lVert\nabla_{\theta}f_{t,i}(\theta_{t-1})-\nabla_{\theta}f_{t,i}(\theta^{*})\rVert^{2}\vert\F_{t-1}] + 2\E[\lVert\nabla_{\theta}f_{t,i}(\theta^{*})\rVert^{2}\vert\F_{t-1}] \nonumber \\
& \leq 2C_{f}^{2}\lVert\theta_{t-1}-\theta^{*}\rVert^{2} + 2\sigma^{2}, \label{eq:ssg:a23}
\end{align}
using $\lVert x+y \rVert^{2} \leq 2(\lVert x \rVert^{2} + \lVert y \rVert^{2})$.
Next, for the second term in \cref{eq:ssg:proof20}: as $\mathcal{F}_{t-1} \subseteq \mathcal{F}_{t-1,i} \subset \mathcal{F}_{t-1,j}$ for all $0 \leq i < j$, we have
\begin{align*}
\E[\langle\nabla_{\theta}f_{t,i}(\theta_{t-1}),\nabla_{\theta}f_{t,j}(\theta_{t-1})\rangle\vert\F_{t-1}]
= \E[\E[\langle\nabla_{\theta}f_{t,i}(\theta_{t-1}),\nabla_{\theta}F(\theta_{t-1})\rangle\vert\F_{t-1,i-1}]\vert\F_{t-1}],
\end{align*}
since $\theta_{t-1}$ and $f_{t,i}$ are $\mathcal{F}_{t-1,j-1}$-measurable for all $0 \leq i < j$, and similarly, as $\theta_{t-1}$ is $\F_{t-1}$-measurable and $\F_{t-1,i-1}$-measurable for all $i \geq 0$, we also have
\begin{align*}
\E[\E[\langle\nabla_{\theta}f_{t,i}(\theta_{t-1}),\nabla_{\theta}F( \theta_{t-1})\rangle\vert\F_{t-1,i-1}]\vert\F_{t-1}]
= \E[\langle\E[\nabla_{\theta}f_{t,i}(\theta_{t-1})\vert\F_{t-1,i-1}]  , \nabla_{\theta}F(\theta_{t-1})\rangle\vert\F_{t-1}]
=\lVert\nabla_{\theta}F(\theta_{t-1})\rVert^{2},
\end{align*}
where $\lVert\nabla_{\theta}F(\theta_{t-1})\rVert^{2} \leq C_{\nabla}^{2}\lVert\theta_{t-1}-\theta^{*}\rVert^{2}$ as $\nabla_{\theta}F$ is $C_{\nabla}$-Lipschitz continuous and $\nabla_{\theta}F(\theta^{*})=0$.
Thus, we obtained a bound for the second term (on the right-hand side) of \cref{eq:lem:nonblock:sg_norm_fsigma} using the bounds of the two terms in \cref{eq:ssg:proof20}:
\begin{align}
\sum_{i=1}^{n_t}(2C_{f}^{2}\lVert\theta_{t-1} - \theta^{*}\rVert^{2} + 2\sigma^2) + \sum_{i \neq j}^{n_{t}} C_{\nabla}^{2} \lVert\theta_{t-1} - \theta^{*}\rVert^{2}
= (2C_{f}^{2}n_{t}+C_{\nabla}^{2}(n_t-1)n_{t})\lVert\theta_{t-1}-\theta^{*}\rVert^{2} + 2\sigma^{2}n_{t}. \label{eq:thm:upper_bound_fulf_block}
\end{align}
For the third term (on the right-hand side) of \cref{eq:lem:nonblock:sg_norm_fsigma} we use that $F$ is $\mu$-quasi-strong convex and $\theta_{t-1}$ is $\F_{t-1}$-measurable,
\begin{align} \label{eq:thm:third_term_uppper_bound}
\E[\langle\nabla_{\theta}f_{t,i}(\theta_{t-1}),\theta_{t-1}-\theta^{*}\rangle\vert\F_{t-1}] = \langle\E[\nabla_{\theta}f_{t,i}(\theta_{t-1})\vert\F_{t-1}],\theta_{t-1}-\theta^{*}\rangle 
= \langle \nabla_{\theta} F(\theta_{t-1}), \theta_{t-1} - \theta^{*}\rangle
\geq \mu\lVert\theta_{t-1}-\theta^{*}\rVert^{2}, 
\end{align}
by \cref{assump:measurable}.
Combining the inequalities from \cref{eq:thm:third_term_uppper_bound,eq:thm:upper_bound_fulf_block} into \cref{eq:lem:nonblock:sg_norm_fsigma} and taking the expectation on both sides of the inequality, yields the recursive relation \eqref{eq:lem:nonblock:streaming_grad_recursive_bound_grad}:
\begin{align*} 
\delta_{t} \leq [1-2\mu\gamma_{t}+(2C_{f}^{2}+(n_{t}-1)C_{\nabla}^{2})n_{t}^{-1}\gamma_{t}^{2}]\delta_{t-1} + 2\sigma^{2}n_{t}^{-1}\gamma_{t}^{2},
\end{align*}
with $\delta_{t}=\E[\lVert\theta_{t}-\theta^{*}\rVert^{2}]$ with some $\delta_{0}\geq0$.
At last, by \cref{prop:appendix:delta_recursive_upper_bound_nt_simpler}, we obtain the desired inequality in \cref{eq:thm:nonblock:streaming_grad_upper_bound}, namely
\begin{align*}
\delta_{t} 
\leq& \exp \left( - \mu \sum_{i=t/2}^{t} \gamma_{i} \right)
\exp \left( 4 C_{f}^{2} \sum_{i=1}^{t} \frac{\gamma_{i}^{2}}{n_{i}} \right) 
\exp \left( 2 C_{\nabla}^{2} \sum_{i=1}^{t} \indicator{n_{i} > 1} \gamma_{i}^{2} \right)
\left( \delta_{0} 
+ \frac{2 \sigma^{2}}{C_{f}^{2}} \right) 
+ \frac{2 \sigma^{2}}{\mu} \max_{t/2 \leq i \leq t}\frac{\gamma_{i}}{n_{i}}.
\end{align*}
using that $(n_{t}-1)n_{t}^{-1}\leq\indicator{n_{t}>1}$, $n_{t}\geq1$, and that $\max_{1 \leq i \leq t} 2\sigma^{2}/(2C_{f}^{2}+(n_{i}-1)C_{\nabla}^{2}) \leq \max_{1 \leq i \leq t} 2\sigma^{2}/2C_{f}^{2}=\sigma^{2}/C_{f}^{2}$.
\end{proof}
\begin{proof}[Proof of \cref{cor:nonblock:streaming_grad_upper_bound_constant}]
By \cref{thm:nonblock:streaming_grad_upper_bound}, we have the upper bound giving as
\begin{align} \label{eq:cor:nonblock:streaming_grad_upper_bound_constant_n}
\delta_{t} \leq& \exp\left(-\mu \sum_{i=t/2}^{t}\gamma_{i}\right)\pi_{t}^{c} + \frac{2\sigma^{2}}{\mu C_{\rho}}\max_{t/2 \leq i \leq t}\gamma_{i}.
\end{align}
as $n_t=C_{\rho}$, with $\pi_{t}^{c}=\exp((4 C_{f}^{2}/C_{\rho})\sum_{i=1}^{t}\gamma_{i}^{2}) \exp(2C_{\nabla}^{2}\indicator{C_{\rho}>1}\sum_{i=1}^{t}\gamma_{i}^{2}) (\delta_{0}+\sigma^{2}/C_{f}^{2})$.
The sum term $\sum_{i=1}^{t}\gamma_{i}^{2}=C_{\gamma}^{2}C_{\rho}^{2\beta}\sum_{i=1}^{t}i^{-2\alpha}$ in $\pi_{t}^{c}$ can be bounded with the help of integral tests for convergence, $\sum_{i=1}^{t}i^{-2\alpha}=1+\sum_{i=2}^{t}i^{-2\alpha}\leq1+\int_{1}^{t}x^{-2\alpha} \, dx \leq 1+1/(2\alpha-1)=2\alpha/(2\alpha-1)$, as $\alpha\in(1/2,1)$.
Likewise, plugging $\gamma_{t} = C_{\gamma} C_{\rho}^{\beta} t^{- \alpha}$ into the first term of
\cref{eq:cor:nonblock:streaming_grad_upper_bound_constant_n}, gives
\begin{align*}
\exp \left( - \mu \sum_{i=t/2}^{t} \gamma_{i} \right) 
= \exp \left( - \mu C_{\gamma} C_{\rho}^{\beta} \sum_{i=t/2}^{t} i^{-\alpha} \right)
\leq \exp \left( - \mu C_{\gamma} C_{\rho}^{\beta} \int_{t/2}^{t} x^{-\alpha} \, dx \right)
\leq \exp \left( - \frac{\mu C_{\gamma}C_{\rho}^{\beta} t^{1-\alpha}}{2^{1-\alpha}}\right),
\end{align*}
using the integral test for convergence.
Next, as $(\gamma_{t})_{t\geq1}$ is decreasing, then $\max_{t/2 \leq i \leq t} \gamma_{t} = \gamma_{t/2}$.
Combining all these findings into \cref{eq:cor:nonblock:streaming_grad_upper_bound_constant_n}, gives us
\begin{align} \label{eq:cor:nonblock:streaming_grad_upper_bound_constant_n_tterm}
\delta_{t} \leq& \exp \left( - \frac{\mu C_{\gamma} C_{\rho}^{\beta} t^{1-\alpha}}{2^{1-\alpha}} \right) \pi_{\infty}^{c} + \frac{2^{1+\alpha} \sigma^{2} C_{\gamma} }{\mu C_{\rho}^{1-\beta} t^{\alpha}},
\end{align}
with $\pi_{\infty}^{c} = \exp(4\alpha C_{\gamma}^{2}(2C_{f}^{2}+C_{\rho}\indicator{C_{\rho}>1}C_{\nabla}^{2})/(2\alpha-1)C_{\rho}^{1-2\beta})(\delta_{0}+2\sigma^{2}/C_{f}^{2})$.
At last, converting \cref{eq:cor:nonblock:streaming_grad_upper_bound_constant_n_tterm} into terms of $N_{t}$ using $N_{t}\ge C_{\rho}t$, yields the desired.
\end{proof}

\begin{proof}[Proof of \cref{cor:nonblock:learning_rate_tau_1}]
For convenience, we divided the proof into two cases to comprehend that $n_{t}\geq1$ for all $t$.
First, we bound each term of \cref{eq:thm:nonblock:streaming_grad_upper_bound} (from \cref{thm:nonblock:streaming_grad_upper_bound}) after inserting $\gamma_{t}=C_{\gamma} n_{t}^{\beta} t^{- \alpha}$ and $n_{t}=\lceil C_{\rho}t^{\rho}\rceil$ into the inequality.
Here, we use $\gamma_{t} \geq C_{\gamma} C_{\rho}^{\beta} t^{\beta \rho - \alpha}$ if $\rho \geq 0$, $\gamma_{t} \geq C_{\gamma}t^{-\alpha}$ if $\rho < 0$, and $x\leq\lceil x \rceil \leq x+1$ for $x\in\R_{+}$.
Thus, for $\rho\geq0$, the first term of \cref{eq:thm:nonblock:streaming_grad_upper_bound} can be bounded, as follows: 
\begin{align*}
\exp \left( - \mu \sum_{i=t/2}^{t} \gamma_{i} \right) 
\leq \exp \left( - \mu C_{\gamma} C_{\rho}^{\beta} \sum_{i=t/2}^{t} i^{\beta \rho -\alpha} \right)
\leq \exp \left( - \frac{\mu C_{\gamma} C_{\rho}^{\beta} t^{1+\beta \rho-\alpha}}{2^{1 + \beta \rho - \alpha}} \right),
\end{align*}
using that $\alpha - \beta \rho \in ( 1/2, 1 )$ and the integral test for convergence. In a same way, for $\rho<0$, one has
\begin{align*}
\exp \left( - \mu \sum_{i=t/2}^{t} \gamma_{i} \right) 
\leq \exp \left( - \mu C_{\gamma} \sum_{i=t/2}^{t} i^{-\alpha} \right) \leq \exp \left( - \frac{\mu C_{\gamma} t^{1-\alpha}}{2^{1-\alpha}}\right).
\end{align*}
Likewise, with the help of integral tests for convergence, we have for $\rho\geq0$, that $\sum_{i=1}^{t}\gamma_{i}^{2}/n_{i} \leq \sum_{i=1}^{t}\gamma_{i}^{2} \leq 2(\alpha-\beta\rho)C_{\gamma}^{2}C_{\rho}^{2\beta}/(2(\alpha-\beta\rho)-1)$, as $n_{t}\geq1$ and $\alpha - \rho \beta > 1/2$. 
For $\rho<0$, one has $\sum_{i=1}^{t}\gamma_{i}^{2}/n_{i} \leq \sum_{i=1}^{t}\gamma_{i}^{2} \leq 2\alpha C_{\gamma}^{2}C_{\rho}^{2\beta}/(2\alpha-1)$ since $C_{\rho}\geq n_{t}\geq1$.
Next, as $(1-\beta)\rho+\alpha > 0$ for $\rho \geq 0$, then we can bound the last term of \cref{eq:thm:nonblock:streaming_grad_upper_bound} by  
\begin{align*}
\frac{2 \sigma^{2}}{\mu} \max_{t/2 \leq i \leq t} \frac{\gamma_{i}}{n_{i}}
\leq \frac{2 \sigma^{2} C_{\gamma}}{\mu C_{\rho}^{1-\beta}} \max_{t/2 \leq i \leq t} \frac{1}{i^{\left(1-\beta \right)\rho+\alpha}}
\leq \frac{2^{1+\left(1-\beta \right)\rho+\alpha} \sigma^{2} C_{\gamma}}{\mu C_{\rho}^{1-\beta} t^{\left(1-\beta \right)\rho+\alpha}}.
\end{align*}
using $n_{t}=\lceil C_{\rho}t^{\rho} \rceil \geq C_{\rho}t^{\rho}$. Likewise, if $\rho < 0$, we have
\begin{align*}
\frac{2 \sigma^{2}}{\mu} \max_{t/2 \leq i \leq t} \frac{\gamma_{i}}{n_{i}}
=  \frac{2 \sigma^{2} C_{\gamma}}{\mu } \max_{t/2 \leq i \leq t} \frac{1}{n_{i}^{1-\beta} i^{\alpha}}
\leq \frac{2^{1+\alpha} \sigma^{2} C_{\gamma}}{\mu  t^{\alpha}},
\end{align*}
since $n_{t} \geq 1$ and $\beta \leq 1$.
Combining all these findings gives
\begin{align} \label{eq:ssg:varying:t}
\delta_{t} \leq& \exp \left( - \frac{\mu C_{\gamma} C_{\rho}^{\beta\indicator{\rho\geq0}} t^{(1-\phi)(1+\tilde{\rho})}}{2^{(1-\phi)(1+\tilde{\rho})}} \right) \pi_{\infty}^{v} + \frac{2^{1+\phi(1+\tilde{\rho})}\sigma^{2}C_{\gamma}}{\mu C_{\rho}^{(1-\beta)\indicator{\rho\geq0}} t^{\phi(1+\tilde{\rho})}},
\end{align}
where $\pi_{\infty}^{v}=\exp(4(\alpha-\beta\tilde{\rho})C_{\gamma}^{2}C_{\rho}^{2\beta}(2C_{f}^{2}+C_{\nabla}^{2})/2(\alpha-\beta\tilde{\rho})-1)$ with $\tilde{\rho}=\rho\indicator{\rho\geq0}$ and $\phi=((1-\beta) \tilde{\rho}+\alpha)/(1+ \tilde{\rho})$.
To write this in terms of $N_{t}$, we use the bounds following bounds: for $\rho \geq 0$, we have that
\begin{align*}
N_{t}=&\sum_{i=1}^{t}n_{i}\leq \sum_{i=1}^{t}(C_{\rho}i^{\rho}+1)=t+C_{\rho}t^{\rho}+C_{\rho}\sum_{i=1}^{t-1}i^{\rho}\leq t+ C_{\rho}t^{\rho}+C_{\rho}\int_{1}^{t}x^{\rho} \, dx
\\ \leq& t+ C_{\rho}t^{\rho}+C_{\rho}(t^{1+\rho}-1)\leq t+C_{\rho}(t^{\rho}-1) + C_{\rho}t^{1+\rho} \leq 2 C_{\rho} t^{1+\rho},
\end{align*}
thus, $t \geq (N_{t}/2 C_{\rho})^{1/(1+\rho)}$.
Similarly, for $\rho <0$, we have that $N_{t} \leq C_{\rho} t$, i.e, $t \geq N_{t}/C_{\rho}$.
\end{proof}

\subsection{Proofs for \texorpdfstring{\cref{sec:nonasymptotic:assg}}{3}}

\begin{lemma}[ASSG/APSSG] \label{lem:avg:nonblock:lrate4}
Let $\Delta_{t}=\E[\lVert\theta_{t}-\theta^{*}\rVert^{4}]$ for $\Delta_{0}\geq0$, where $(\theta_{t})$ either follows the recursion in \cref{eq:streaming_grad} or \cref{eq:ssg:proj}.
Suppose \cref{assump:L_strong_convexity,assump:L_lipschitz,assump:measurable,assump:ssg:f_lipschitz,assump:ssg:sigma,avg:assump:L_lipschitz,assump:assg:L_lipschitz} hold with $p=4$.
Then, for any learning rate $(\gamma_{t})$ and time-varying mini-batch $(n_{t})$, we have
\begin{align} \label{eq:fourth_order_moment_upper_bound_general}
\Delta_{t} 
\leq& \exp \left( - \mu \sum_{i=t/2}^{t} \gamma_{i} \right) \Pi_{t}^{\Delta}
+ \frac{32 \sigma^{4}}{\mu^{2}} \max_{t/2 \leq i \leq t} \frac{\gamma_{i}^{2}}{n_{i}^{2}} 
+ \frac{48 \sigma^{4}}{\mu} \max_{t/2 \leq i \leq t} \frac{\gamma_{i}^{3}}{n_{i}^{3}} 
+ \frac{114 \sigma^{4}}{\mu} \max_{t/2 \leq i \leq t} \frac{\gamma_{i}^{3} \indicator{n_{i} > 1}}{n_{i}^{2}},
\end{align}
with $\Pi_{t}^{\Delta}$ given in \cref{eq:avg:4momoent:pi}.
\end{lemma}

\begin{proof}[Proof of \cref{lem:avg:nonblock:lrate4}]
We will now derive the recursive step sequence for the fourth-order moment using the same arguments as in proof for \cref{thm:nonblock:streaming_grad_upper_bound}.
Thus, one can show that
\begin{align*}
\E[\lVert\theta_{t}-\theta^{*}\rVert^{4}\vert\F_{t-1}]
\leq& \lVert\theta_{t-1}-\theta^{*}\rVert^{4} 
+ \frac{\gamma_{t}^{4}}{n_{t}^{4}} \E \left[ \left\lVert\sum_{i=1}^{n_{t}} \nabla_{\theta} f_{t,i} \left( \theta_{t-1}\right)\right\rVert^{4} \middle\vert \F_{t-1} \right]
+ \frac{4\gamma_{t}^{2}}{n_{t}^{2}}  \E \left[ \left\langle \sum_{i=1}^{n_{t}} \nabla_{\theta} f_{t,i} \left( \theta_{t-1}\right), \theta_{t-1} - \theta^* \right\rangle^{2} \middle\vert \F_{t-1} \right]
\\ +& \frac{2\gamma_{t}^{2}}{n_{t}^{2}} \lVert\theta_{t-1} - \theta^{*}\rVert^{2} \E \left[\left\lVert\sum_{i=1}^{n_{t}} \nabla_{\theta} f_{t,i} \left( \theta_{t-1}\right)\right\rVert^{2} \middle\vert \F_{t-1} \right]
- \frac{4 \gamma_{t}}{n_{t}} \lVert\theta_{t-1} - \theta^{*}\rVert^{2} \sum_{i=1}^{n_{t}} \langle \E [ \nabla_{\theta} f_{t,i} ( \theta_{t-1}) \vert \F_{t-1} ], \theta_{t-1} - \theta^{*} \rangle
\\ +& \frac{4 \gamma_{t}^{3}}{n_{t}^{3}} \E \left[ \left\lVert \sum_{i=1}^{n_{t}} \nabla_{\theta} f_{t,i} \left( \theta_{t-1}\right)\right\rVert^{2} 
 \left\langle \sum_{i=1}^{n_{t}} \nabla_{\theta} f_{t,i} \left( \theta_{t-1}\right), \theta_{t-1} - \theta^* \right\rangle \middle\vert \F_{t-1} \right],
\end{align*}
using $\theta_{t-1}$ is $\F_{t-1}-$measurable.
Note, by \cref{assump:measurable}, we have
\begin{align*}
\langle \E [ \nabla_{\theta} f_{t,i} ( \theta_{t-1}) \vert \F_{t-1} ] , \theta_{t-1} - \theta^{*} \rangle 
= \langle \nabla_{\theta} F (\theta_{t-1}), \theta_{t-1} - \theta^{*} \rangle
\geq \mu \lVert\theta_{t-1} - \theta^{*}\rVert^{2},
\end{align*}
as $F$ is $\mu$-quasi-strong convex.
Combining this with Cauchy-Schwarz inequality (i.e., $\langle x,y \rangle \leq \lVert x \rVert \lVert y \rVert$), we obtain the simplified expression:
\begin{align*}
\E [ \lVert\theta_{t} - \theta^{*}\rVert^{4} \vert \F_{t-1} ]
\leq& \lVert\theta_{t-1} - \theta^{*}\rVert^{4} 
+ \frac{\gamma_{t}^{4}}{n_{t}^{4}} \E \left[ \left\lVert\sum_{i=1}^{n_{t}} \nabla_{\theta} f_{t,i} \left( \theta_{t-1}\right)\right\rVert^{4} \middle\vert \F_{t-1} \right]
+ \frac{6\gamma_{t}^{2}}{n_{t}^{2}} \lVert\theta_{t-1} - \theta^{*}\rVert^{2} \E \left[\left\lVert\sum_{i=1}^{n_{t}} \nabla_{\theta} f_{t,i} \left( \theta_{t-1}\right)\right\rVert^{2} \middle\vert \F_{t-1} \right]
\\ &- 4 \mu \gamma_{t} \lVert\theta_{t-1} - \theta^{*}\rVert^{4}
+ \frac{4 \gamma_{t}^{3}}{n_{t}^{3}} \lVert\theta_{t-1} - \theta^{*}\rVert \E \left[ \left\lVert \sum_{i=1}^{n_{t}} \nabla_{\theta} f_{t,i} \left( \theta_{t-1}\right)\right\rVert^{3} \middle\vert \F_{t-1} \right].
\end{align*}
Next, recall Young's inequality for products, i.e., for any $a_{t}, b_{t}, c_{t} >0$, we have $a_{t} b_{t} \leq a^{2}_{t} c_{t}^{2}/2 + b_{t}^{2}/2 c_{t}^{2}$,
\begin{align*}
\left\lVert \sum_{i=1}^{n_{t}} \nabla_{\theta} f_{t,i} \left( \theta_{t-1}\right)\right\rVert^{3} 
\leq \frac{\gamma_{t}}{2n_{t} \left\lVert\theta_{t-1} - \theta^{*}\right\rVert} \left\lVert \sum_{i=1}^{n_{t}} \nabla_{\theta} f_{t,i} \left( \theta_{t-1}\right)\right\rVert^{4}
+ \frac{2n_{t} \left\lVert\theta_{t-1} - \theta^{*}\right\rVert}{\gamma_{t}} \left\lVert \sum_{i=1}^{n_{t}} \nabla_{\theta} f_{t,i} \left( \theta_{t-1}\right)\right\rVert^{2},
\end{align*}
giving us
\begin{align}
\E [ \lVert\theta_{t} - \theta^{*}\rVert^{4} \vert \F_{t-1} ]
\leq& (1- 4 \mu \gamma_{t} )\lVert\theta_{t-1} - \theta^{*}\rVert^{4} 
+ \frac{3\gamma_{t}^{4}}{n_{t}^{4}} \E \left[ \left\lVert\sum_{i=1}^{n_{t}} \nabla_{\theta} f_{t,i} \left( \theta_{t-1}\right)\right\rVert^{4} \middle\vert \F_{t-1} \right] \nonumber
\\ &+ \frac{8\gamma_{t}^{2}}{n_{t}^{2}} \lVert\theta_{t-1} - \theta^{*}\rVert^{2} \E \left[\left\lVert\sum_{i=1}^{n_{t}} \nabla_{\theta} f_{t,i} \left( \theta_{t-1}\right)\right\rVert^{2} \middle\vert \F_{t-1} \right].  \label{thm:eq:recursive_step_avg_ft1}
\end{align}
To bound the second and fourth-order terms in \cref{thm:eq:recursive_step_avg_ft1}, we would need to study the recursive sequences: firstly, utilizing the Lipschitz continuity of $\nabla_{\theta} f_{t,i}$, together with \cref{assump:ssg:f_lipschitz,assump:ssg:sigma}, and that $\theta_{t-1}$ is $\F_{t-1}$-measurable (\cref{assump:measurable}), we obtain
\begin{align}
 \E [ \lVert\nabla_{\theta} f_{t,i} ( \theta_{t-1} )\rVert^{p} \vert \F_{t-1} ] 
& \leq 2^{p-1} [\E [\lVert\nabla_{\theta} f_{t,i} ( \theta_{t-1} ) - \nabla_{\theta} f_{t,i} ( \theta^{*})\rVert^{p} \vert \F_{t-1}]
+ \E [ \lVert\nabla_{\theta} f_{t,i} ( \theta^{*})\rVert^{p} \vert \F_{t-1} ] ] \nonumber
\\ & \leq 2^{p-1}[ C_{f}^{p} \lVert\theta_{t-1} - \theta^{*}\rVert^{p} + \sigma^{p}] , \label{thm:eq:p_bound_lemma}
\end{align}
for any $p \in [1,4]$ using the bound $\lVert x + y \rVert^{p} \leq 2^{p-1}(\lVert x \rVert^{p} + \lVert y \rVert^{p})$.
Thus, we can bound the second-order term in \cref{thm:eq:recursive_step_avg_ft1} by
\begin{align} 
\E \left[ \left\lVert \sum_{i=1}^{n_{t}} \nabla_{\theta} f_{t,i} \left( \theta_{t-1} \right)\right\rVert^{2} \middle\vert \F_{t-1} \right] 
& \leq [2C_{f}^{2} n_{t} +  C_{\nabla}^{2}(n_t-1)n_{t} ] \lVert\theta_{t-1} - \theta^{*}\rVert^{2} + 2\sigma^2 n_{t} \nonumber
\\ & \leq [2C_{f}^{2} n_{t} +  C_{\nabla}^{2}n_{t}^{2} \indicator{n_{t} > 1} ] \lVert\theta_{t-1} - \theta^{*}\rVert^{2} + 2\sigma^2 n_{t}, \label{eq:thm:second_for_avg}
\end{align}
following the same steps in the proof of \cref{thm:nonblock:streaming_grad_upper_bound}, but with use of \cref{thm:eq:p_bound_lemma}.
Bounding the fourth-order term is a bit heavier computationally, but let us recall that $\lVert \sum_{i} x_{i} \rVert^{2} = \sum_{i} \lVert x_{i} \rVert^{2} + \sum_{i \neq j} \langle x_{i} , x_{j} \rangle = \sum_{i} \lVert x_{i} \rVert^{2} + 2\sum_{i < j} \langle x_{i} , x_{j} \rangle$.
Then, we have that
\begin{align}
\left\lVert \sum_{i=1}^{n_{t}} \nabla_{\theta} f_{t,i} ( \theta_{t-1} ) \right\rVert^{4} 
=& \left(\sum_{i=1}^{n_{t}} \lVert \nabla_{\theta} f_{t,i} ( \theta_{t-1} ) \rVert^{2} 
+ \sum_{i \neq j}^{n_{t}} \langle \nabla_{\theta} f_{t,i} ( \theta_{t-1} ), \nabla_{\theta} f_{t,j} ( \theta_{t-1} ) \rangle \right)^{2} \nonumber
\\ \leq&  2 \left(\sum_{i=1}^{n_{t}} \lVert \nabla_{\theta} f_{t,i} ( \theta_{t-1} ) \rVert^{2} \right)^{2} 
+ 4 \left( \sum_{i < j}^{n_{t}} \langle \nabla_{\theta} f_{t,i} ( \theta_{t-1} ), \nabla_{\theta} f_{t,j} ( \theta_{t-1} ) \rangle \right)^{2}, \label{eq:thm:4thmoment_expand}
\end{align}
as $(x + y)^{2} \leq 2x^{2} + 2y^{2}$.
For the first term of \cref{eq:thm:4thmoment_expand}, we have
\begin{align*}
\E \left[ \left( \sum_{i=1}^{n_{t}} \lVert \nabla_{\theta} f_{t,i} ( \theta_{t-1} ) \rVert^{2} \right)^{2} \middle\vert \F_{t-1} \right] 
=& \sum_{i=1}^{n_{t}} \E [ \lVert \nabla_{\theta} f_{t,i} ( \theta_{t-1} ) \rVert^{4} \vert \F_{t-1} ]
+ \sum_{i \neq j}^{n_{t}} \E [ \lVert \nabla_{\theta} f_{t,i} ( \theta_{t-1} ) \rVert^{2} \lVert \nabla_{\theta} f_{t,j} ( \theta_{t-1} ) \rVert^{2} \vert \F_{t-1} ]
\\ \leq& 8 n_{t} [ C_{f}^{4} \lVert\theta_{t-1} - \theta^{*}\rVert^{4} + \sigma^{4}] + 4n_{t}^{2} \indicator{n_{t} > 1} [ C_{f}^{2} \lVert\theta_{t-1} - \theta^{*}\rVert^{2} + \sigma^{2}]^{2},
\end{align*}
using the bound from \cref{thm:eq:p_bound_lemma}, $n_{t}(n_{t}-1)\leq n_{t}^{2}\indicator{n_{t}>1}$, and that $\mathcal{F}_{t-1} \subseteq \mathcal{F}_{t-1,i} \subset \mathcal{F}_{t-1,j}$ for all $0 \leq i < j$.
To bound the second term of \cref{eq:thm:4thmoment_expand}, we ease notation by denoting $\nabla_{\theta} f_{t,i} ( \theta_{t-1} )$ by $\upsilon_{i}$, giving us
\begin{align*}
\left( \sum_{i < j}^{n_{t}} \langle \upsilon_{i}, \upsilon_{j} \rangle \right)^{2}
 & = \sum_{i<j}^{n_{t}} \langle \upsilon_{i}, \upsilon_{j} \rangle^{2} 
+ \sum_{\substack{i<j,k<l\\(i,j)\neq(k,l)}}^{n_{t}} \langle \upsilon_{i}, \upsilon_{j} \rangle \langle \upsilon_{k}, \upsilon_{l} \rangle
 \\ & = \underbrace{\sum_{i<j}^{n_{t}} \langle \upsilon_{i}, \upsilon_{j} \rangle^{2}}_{A}
+ \underbrace{\sum_{\substack{i<j,k<l\\(i,j)\neq(k,l),j=l}}^{n_{t}} \langle \upsilon_{i}, \upsilon_{j} \rangle \langle \upsilon_{k}, \upsilon_{l}\rangle}_{B}
+ \underbrace{\sum_{\substack{i<j,k<l\\(i,j)\neq(k,l),j\neq l}}^{n_{t}} \langle \upsilon_{i}, \upsilon_{j}\rangle\langle \upsilon_{k}, \upsilon_{l}\rangle}_{C}.
\end{align*}
By Cauchy-Schwarz inequality, we can bound the first term $A$, by
\begin{align*}
\E [ A \vert \F_{t-1} ] 
\leq& \sum_{i < j}^{n_{t}} \E [ \lVert\upsilon_{i}\rVert^{2} \lVert\upsilon_{j}\rVert^{2} \vert \F_{t-1} ]
\leq 2n_{t}(n_{t}-1) [ C_{f}^{2} \lVert\theta_{t-1} - \theta^{*}\rVert^{2} + \sigma^{2}]^{2}
\leq 2n_{t}^{2} \indicator{n_{t} > 1} [ C_{f}^{2} \lVert\theta_{t-1} - \theta^{*}\rVert^{2} + \sigma^{2}]^{2},
\end{align*}
using that $\mathcal{F}_{t-1} \subseteq \mathcal{F}_{t-1,i} \subset \mathcal{F}_{t-1,j}$ for all $0 \leq i < j$.
Next, since $l=j$ implies $i \neq k$, we have
\begin{align*}
\E [ B \vert \F_{t-1} ]
=& \sum_{i < j, k < l, i\neq k, j=l }^{n_{t}} \E [ \langle \upsilon_{i}, \upsilon_{j} \rangle \langle \upsilon_{k}, \upsilon_{l} \rangle \vert \F_{t-1} ]
\\ =& \sum_{i < j,k < l,i\neq k,j=l}^{n_{t}} \E [ \E [ \langle  \E [ \upsilon_{i} \vert \F_{t-1,i-1} ], \upsilon_{j} \rangle \langle \E [ \upsilon_{k} \vert \F_{t-1,k-1} ], \upsilon_{l} \rangle \vert \F_{t-1,l-1} ] \vert \F_{t-1} ] 
\\ =& \sum_{i < j,k < l,i\neq k,j=l}^{n_{t}} \E [ \E [ \langle \nabla_{\theta} F ( \theta_{t-1} ) , \upsilon_{l} \rangle^{2} \vert \F_{t-1,l-1} ] \vert \F_{t-1} ] 
\\ \leq& \sum_{i < j,k < l,i\neq k,j=l}^{n_{t}} \E [  \lVert\nabla_{\theta} F ( \theta_{t-1} )\rVert^{2} \E [ \lVert\upsilon_{l}\rVert^{2} \vert \F_{t-1,l-1} ] \vert \F_{t-1} ]
\\ \leq& \sum_{i < j,k < l,i\neq k,j=l}^{n_{t}} 2 C_{\nabla}^{2} \lVert \theta_{t-1} - \theta^{*}\rVert^{2} [C_{f}^{2} \lVert\theta_{t-1} - \theta^{*}\rVert^{2} +  \sigma^{2} ]
\\ =& n_{t}( n_{t}-1) ( n_{t}-2 ) C_{\nabla}^{2} \lVert \theta_{t-1} - \theta^{*}\rVert^{2} [C_{f}^{2} \lVert\theta_{t-1} - \theta^{*}\rVert^{2} +  \sigma^{2} ]
\\ \leq& n_{t}^{3} \indicator{n_{t} > 1}  C_{\nabla}^{2} \lVert \theta_{t-1} - \theta^{*}\rVert^{2} [C_{f}^{2} \lVert\theta_{t-1} - \theta^{*}\rVert^{2} +  \sigma^{2} ],
\end{align*}
using Cauchy-Schwarz inequality and the bound in \cref{thm:eq:p_bound_lemma}.
In the same way, as $j \neq l$ includes $(i,j)\neq(k,l)$, we can rewrite $C$ as
\begin{align*}
C = \sum_{i < j, k < l, j \neq l }^{n_{t}} \langle \upsilon_{i}, \upsilon_{j} \rangle \langle \upsilon_{k}, \upsilon_{l} \rangle
= \underbrace{\sum_{i < j,k < l, i=k, j \neq l}^{n_{t}} \langle \upsilon_{i}, \upsilon_{j} \rangle \langle \upsilon_{k}, \upsilon_{l} \rangle}_{C_{1}}
+ \underbrace{\sum_{i < j,k < l, i \neq k , j \neq l }^{n_{t}} \langle \upsilon_{i}, \upsilon_{j} \rangle \langle \upsilon_{k}, \upsilon_{l} \rangle}_{C_{2}},
\end{align*}
where $\E [ C_{1} \vert \F_{t-1} ] = \E [ B \vert \F_{t-1} ]$.
Finally, we can rewrite $C_{2}$ as
\begin{align*}
C_{2} & =  \underbrace{\sum_{i < j,k < l, i \neq k , j \neq l, i=l , j\neq k}^{n_{t}} \langle \upsilon_{i} \upsilon_{j} \rangle \langle \upsilon_{k} \upsilon_{l} \rangle}_{C_{2,1}} 
+ \underbrace{\sum_{i < j,k < l, i \neq k , j \neq l, i \neq l , j = k}^{n_{t}} \langle \upsilon_{i} \upsilon_{j} \rangle \langle \upsilon_{k} \upsilon_{l} \rangle}_{C_{2,2}}
+ \underbrace{\sum_{i < j,k < l, i\neq j \neq k \neq l}^{n_{t}} \langle \upsilon_{i} \upsilon_{j} \rangle \langle \upsilon_{k} \upsilon_{l} \rangle}_{C_{2,3}},
\end{align*}
where $\E [ C_{2,1} \vert \F_{t-1} ] = \E [ C_{2,2} \vert \F_{t-1} ] = \E[ B \vert \F_{t-1}]$, and 
\begin{align*}
\E [ C_{2,3} \vert \F_{t-1} ] 
& = \sum_{i < j,k < l, i\neq j \neq k \neq l}^{n_{t}} \E [ \lVert \nabla_{\theta} F ( \theta_{t-1} ) \rVert^{4} \vert \F_{t-1} ] 
\\ & \leq  n_{t} ( n_{t} -1 ) ( n_{t} -2 ) ( n_{t} -3 ) C_{\nabla}^{4} \lVert\theta_{t-1} - \theta^{*}\rVert^{4}
\\ & \leq  n_{t}^{4} \indicator{n_{t} > 1} C_{\nabla}^{4} \lVert\theta_{t-1} - \theta^{*}\rVert^{4}.
\end{align*}
Thus, the fourth-order term of \cref{thm:eq:recursive_step_avg_ft1}, is bounded by
\begin{align}
\E \left[ \left\lVert\sum_{i=1}^{n_{t}} \nabla_{\theta} f_{t,i} \left( \theta_{t-1} \right)\right\rVert^{4} \middle\vert \F_{t-1} \right] 
\leq& 16 n_{t} [ C_{f}^{4} \lVert\theta_{t-1} - \theta^{*}\rVert^{4} + \sigma^{4}]
+ 16 n_{t}^{2} \indicator{n_{t} > 1} [ C_{f}^{2} \lVert\theta_{t-1} - \theta^{*}\rVert^{2} + \sigma^{2}]^{2} \nonumber
\\ &+ 12 n_{t}^{3} \indicator{n_{t} > 1} C_{\nabla}^{2} \lVert \theta_{t-1} - \theta^{*}\rVert^{2} [C_{f}^{2} \lVert\theta_{t-1} - \theta^{*}\rVert^{2} +  \sigma^{2} ]
+ 4n_{t}^{4} \indicator{n_{t} > 1} C_{\nabla}^{4} \lVert\theta_{t-1} - \theta^{*}\rVert^{4}\nonumber
\\ \leq& [ 16 C_{f}^{4} n_{t} + 16 C_{f}^{4} n_{t}^{2} \indicator{n_{t} > 1} + 12 C_{\nabla}^{2} C_{f}^{2} n_{t}^{3} \indicator{n_{t} > 1} + 4 C_{\nabla}^{4} n_{t}^{4} \indicator{n_{t} > 1} ] \lVert\theta_{t-1} - \theta^{*}\rVert^{4} \nonumber
\\ &+ [ 32 C_{f}^{2} \sigma^{2} n_{t}^{2} \indicator{n_{t} > 1} + 12 C_{\nabla}^{2} \sigma^{2} n_{t}^{3} \indicator{n_{t} > 1} ] \lVert\theta_{t-1} - \theta^{*}\rVert^{2}
+ 16 \sigma^{4} n_{t} + 16 \sigma^{4} n_{t}^{2} \indicator{n_{t} > 1}. \label{eq:thm:fourth_for_avg}
\end{align}
Combining the bound from \cref{eq:thm:second_for_avg,eq:thm:fourth_for_avg} into \cref{thm:eq:recursive_step_avg_ft1}, we can bound the fourth-order moment $\E [ \lVert\theta_{t} - \theta^{*}\rVert^{4} \vert \F_{t-1} ]$ by the recursive relation:
\begin{align*}
&[ 1- 4 \mu \gamma_{t} + 8C_{\nabla}^{2} \indicator{n_{t} > 1} \gamma_{t}^{2} + 16C_{f}^{2} n_{t}^{-1} \gamma_{t}^{2} + 48 C_{f}^{4} n_{t}^{-3} \gamma_{t}^{4} + 48 C_{f}^{4} n_{t}^{-2} \indicator{n_{t} > 1} \gamma_{t}^{4} + 36 C_{\nabla}^{2} C_{f}^{2} n_{t}^{-1} \indicator{n_{t} > 1} \gamma_{t}^{4} 
\\ &+ 12 C_{\nabla}^{4} \indicator{n_{t} > 1} \gamma_{t}^{4} ] \lVert\theta_{t-1} - \theta^{*}\rVert^{4}
+ [16 \sigma^2 n_{t}^{-1} \gamma_{t}^{2} + 96 C_{f}^{2} \sigma^{2} n_{t}^{-2} \indicator{n_{t} > 1} \gamma_{t}^{4} + 36 C_{\nabla}^{2} \sigma^{2} n_{t}^{-1} \indicator{n_{t} > 1} \gamma_{t}^{4} ] \lVert\theta_{t-1} - \theta^{*}\rVert^{2}
\\ &+ 48 \sigma^{4} n_{t}^{-3} \gamma_{t}^{4} + 48 \sigma^{4} n_{t}^{-2} \indicator{n_{t} > 1} \gamma_{t}^{4}.
\end{align*}
By Young's inequality for products, one have
\begin{align*}
2C_{\nabla}^{2}C_{f}^{2} & \leq n_{t}C_{\nabla}^{4}+n_{t}^{-1}C_{f}^{4}, \\
16\sigma^{2}n_{t}^{-1}\gamma_{t}^{2}\lVert\theta_{t-1}-\theta^{*}\rVert^{2} & \leq 2\mu\gamma_{t}\lVert\theta_{t}-\theta^{*}\rVert^{4}+32\sigma^{4}\mu^{-1}n_{t}^{-2}\gamma_{t}^{3}, \\
2C_{f}^{2}\sigma^{2}n_{t}^{-2}\indicator{n_{t}>1}\gamma_{t}^{4}\lVert\theta_{t-1}-\theta^{*}\rVert^{2} & \leq C_{f}^{4}n_{t}^{-2}\indicator{n_{t}>1}\gamma_{t}^{4}\lVert\theta_{t}-\theta^{*}\rVert^{4}+\sigma^{4}n_{t}^{-2}\indicator{n_{t}>1}\gamma_{t}^{4}, \\
2C_{\nabla}^{2}\sigma^{2}n_{t}^{-1}\indicator{n_{t}>1}\gamma_{t}^{4}\lVert\theta_{t-1}-\theta^{*}\rVert^{2} & \leq C_{\nabla}^{4}\indicator{n_{t}>1}\gamma_{t}^{4}\lVert\theta_{t}-\theta^{*}\rVert^{4}+\sigma^{4}n_{t}^{-2}\indicator{n_{t}>1}\gamma_{t}^{4},
\end{align*}
which yields the bound on $\E [ \lVert\theta_{t} - \theta^{*}\rVert^{4} \vert \F_{t-1} ]$,
\begin{align}
&[ 1- 2 \mu \gamma_{t} + 8C_{\nabla}^{2} \indicator{n_{t} > 1} \gamma_{t}^{2} + 16C_{f}^{2} n_{t}^{-1} \gamma_{t}^{2} + 48 C_{f}^{4} n_{t}^{-3} \gamma_{t}^{4} + 114 C_{f}^{4} n_{t}^{-2} \indicator{n_{t} > 1} \gamma_{t}^{4} + 48 C_{\nabla}^{4} \indicator{n_{t} > 1} \gamma_{t}^{4} ] \lVert\theta_{t-1} - \theta^{*}\rVert^{4} \nonumber
\\ &+ 32\mu^{-1} \sigma^{4} n_{t}^{-2} \gamma_{t}^{3} + 48 \sigma^{4} n_{t}^{-3} \gamma_{t}^{4} + 114 \sigma^{4} n_{t}^{-2} \indicator{n_{t} > 1} \gamma_{t}^{4}. \label{eq:thm:recursive_form_4m_ft1}
\end{align}
Taking, the expectation on both sides of the inequality in \cref{eq:thm:recursive_form_4m_ft1} yields the recursive relation for the fourth-order moment:
\begin{align}
\Delta_{t}
\leq& [ 1- 2 \mu \gamma_{t} + 8C_{\nabla}^{2} \indicator{n_{t} > 1} \gamma_{t}^{2} + 16C_{f}^{2} n_{t}^{-1} \gamma_{t}^{2} + 48 C_{f}^{4} n_{t}^{-3} \gamma_{t}^{4} + 114 C_{f}^{4} n_{t}^{-2} \indicator{n_{t} > 1} \gamma_{t}^{4} + 48 C_{\nabla}^{4} \indicator{n_{t} > 1} \gamma_{t}^{4} ] \Delta_{t-1} \nonumber
\\ &+ 32\mu^{-1} \sigma^{4} n_{t}^{-2} \gamma_{t}^{3} + 48 \sigma^{4} n_{t}^{-3} \gamma_{t}^{4} + 114 \sigma^{4} n_{t}^{-2} \indicator{n_{t} > 1} \gamma_{t}^{4}. \label{eq:thm:recursive_form_4m_f}
\end{align}
with $\Delta_{t} = \E [ \lVert\theta_{t} - \theta^{*}\rVert^{4} ]$ for some $\Delta_{0} \geq 0$.
By \cref{prop:appendix:delta_recursive_upper_bound_nt_simpler}, we achieve the (upper) bound of $\Delta_{t}$ in \cref{eq:thm:recursive_form_4m_f}, given as
\begin{align*}
\Delta_{t} \leq& \exp \left( - \mu \sum_{i=t/2}^{t} \gamma_{i} \right) \Pi_{t}^{\Delta} + \frac{32 \sigma^{4}}{\mu^{2}} \max_{t/2 \leq i \leq t} \frac{\gamma_{i}^{2}}{n_{i}^{2}} + \frac{48 \sigma^{4}}{\mu} \max_{t/2 \leq i \leq t} \frac{\gamma_{i}^{3}}{n_{i}^{3}} + \frac{114 \sigma^{4}}{\mu} \max_{t/2 \leq i \leq t} \frac{\gamma_{i}^{3} \indicator{n_{i} > 1}}{n_{i}^{2}}.
\end{align*}
where $\Pi_{t}^{\Delta}$ is given by
\begin{align} 
&\exp \left( 32 C_{f}^{2} \sum_{i=1}^{t} \frac{\gamma_{i}^{2}}{n_{i}} \right) \exp \left( 96 C_{f}^{4} \sum_{i=1}^{t} \frac{\gamma_{i}^{4}}{n_{i}^{3}} \right) \exp \left( 228 C_{f}^{4} \sum_{i=1}^{t} \frac{\indicator{n_{i} > 1} \gamma_{i}^{4}}{n_{i}^{2}} \right) \nonumber
\\ &\exp \left( 16 C_{\nabla}^{2} \sum_{i=1}^{t} \indicator{n_{i} > 1} \gamma_{i}^{2} \right) \exp \left( 96 C_{\nabla}^{4} \sum_{i=1}^{t} \indicator{n_{i} > 1} \gamma_{i}^{4} \right) \left( \Delta_{0} + \frac{2\sigma^{4}}{C_{f}^{4}} + \frac{4\sigma^{4}\gamma_1}{ \mu C_{f}^{2}n_1} \right),\label{eq:avg:4momoent:pi}
\end{align}
with use of 
\begin{align*}
\max_{1 \leq i \leq t} \frac{32 \mu^{-1} \sigma^{4} n_{i}^{-2} \gamma_{i} + 48 \sigma^{4} n_{i}^{-3} \gamma_{i}^{2} + 114 \sigma^{4} n_{i}^{-2} \indicator{n_{i} > 1} \gamma_{i}^{2}}{8 C_{\nabla}^{2} \indicator{n_{i} > 1} + 16C_{f}^{2} n_{i}^{-1} + 48 C_{f}^{4} n_{i}^{-3} \gamma_{i}^{2} + 114 C_{f}^{4} n_{i}^{-2} \indicator{n_{i} > 1} \gamma_{i}^{2} + 48 C_{\nabla}^{4} \indicator{n_{i} > 1} \gamma_{i}^{2}}
\leq \frac{\sigma^{4}}{C_{f}^{4}} + \frac{2\sigma^{4}\gamma_1}{ \mu C_{f}^{2}n_1}.
\end{align*}
At last, bounding the projected estimate \cref{eq:ssg:proj} follows from that $\E[\lVert\mathcal{P}_{\Theta}(\theta)-\theta^{*}\rVert^{2}]\leq\E[\lVert\theta-\theta^{*}\rVert^{2}]$, $\forall\theta\in\Theta$.
\end{proof}

\subsubsection{Proofs for \texorpdfstring{\cref{sec:polyakruppert:unbounded}}{3.1}}

\begin{proof}[Proof of \cref{thm:avg:nonblock:streaming_grad_upper_bound}]
Following \citet{polyak1992acceleration}, we rewrite \cref{eq:streaming_grad} to
\begin{align}
\theta_{t} = \theta_{t-1} - \frac{\gamma_{t}}{n_{t}} \sum_{i=1}^{n_{t}} \nabla_{\theta} f_{t,i} ( \theta_{t-1} )
\iff \frac{1}{\gamma_{t}} ( \theta_{t-1} - \theta_{t} ) = \nabla_{\theta} f_{t} ( \theta_{t-1} ), \label{eq:thm:grad_f_theta_relation}
\end{align}
where $\nabla_{\theta}f_{t}(\theta_{t-1})$ denotes $n_{t}^{-1}\sum_{i=1}^{n_{t}}\nabla_{\theta}f_{t,i}(\theta_{t-1})$.
Observe that
\begin{align*}
\nabla_{\theta}^{2} F ( \theta^{*} ) ( \theta_{t-1} - \theta^{*} )
=& \nabla_{\theta} f_{t} ( \theta_{t-1} ) 
- \nabla_{\theta} f_{t} ( \theta^{*} ) 
- \underbrace{[\nabla_{\theta} f_{t} ( \theta_{t-1} ) - \nabla_{\theta} f_{t} ( \theta^{*} ) - \nabla_{\theta} F ( \theta_{t-1} )]}_{\text{martingale term}} 
- \underbrace{[\nabla_{\theta} F ( \theta_{t-1} ) - \nabla_{\theta}^{2} F ( \theta^{*} ) ( \theta_{t-1} - \theta^{*} )]}_{\text{rest term}},
\end{align*}
where $\nabla_{\theta}^{2} F(\theta^{*})$ is invertible with lowest eigenvalue greater than $\mu$, i.e., $\nabla_{\theta}^{2}F(\theta^{*}) \geq \mu$.
Thus, summing the parts and using the Minkowski's inequality, we obtain the inequality:
\begin{align*}
\left( \E \left[ \left\lVert\bar{\theta}_{t} - \theta^{*}\right\rVert^{2} \right] \right)^{\frac{1}{2}}
\leq& \left( \E \left[ \left\lVert \nabla_{\theta}^{2} F \left( \theta^{*} \right)^{-1} \frac{1}{N_{t}} \sum_{i=1}^{t} n_{i} \nabla_{\theta} f_{i} \left(\theta^{*} \right)\right\rVert^{2} \right] \right)^{\frac{1}{2}}
+ \left( \E \left[ \left\lVert\nabla_{\theta}^{2} F \left( \theta^{*} \right)^{-1} \frac{1}{N_{t}} \sum_{i=1}^{t} n_{i} \nabla_{\theta} f_{i} \left( \theta_{i-1} \right) \right\rVert^{2} \right] \right)^{\frac{1}{2}}
\\ &+ \left( \E \left[ \left\lVert\nabla_{\theta}^{2} F \left( \theta^{*} \right)^{-1} \frac{1}{N_{t}} \sum_{i=1}^{t} n_{i} \left[\nabla_{\theta} f_{i} \left( \theta_{i-1} \right) - \nabla_{\theta} f_{i} \left( \theta^{*} \right) - \nabla_{\theta} F \left( \theta_{i-1} \right)\right]\right\rVert^{2} \right] \right)^{\frac{1}{2}}
\\ &+ \left( \E \left[ \left\lVert\nabla_{\theta}^{2} F \left( \theta^{*} \right)^{-1} \frac{1}{N_{t}} \sum_{i=1}^{t} n_{i} \left[ \nabla_{\theta} F \left( \theta_{i-1} \right) - \nabla_{\theta}^{2} F \left( \theta^{*} \right) \left( \theta_{i-1} - \theta^{*} \right)\right] \right\rVert^{2} \right] \right)^{\frac{1}{2}}.
\end{align*}
As $(\nabla_{\theta}f_{t,i}(\theta^{*}))$ is a square-integrable martingale increment sequences on $\R^d$ (\cref{assump:measurable}), we have
\begin{align} 
\E \left[ \left\lVert \nabla_{\theta}^{2} F \left( \theta^{*} \right)^{-1} \frac{1}{N_{t}} \sum_{i=1}^{t} n_{i} \nabla_{\theta} f_{i} \left(\theta^{*} \right)\right\rVert^{2} \right]
\leq \frac{1}{N_{t}^{2}}\sum_{i=1}^{t}\sum_{j=1}^{n_{i}}\E\left[\left\lVert \nabla_{\theta}^{2} F \left( \theta^{*} \right)^{-1} \nabla_{\theta} f_{i,j} \left(\theta^{*} \right) \right\rVert^{2} \right]
\leq \frac{\operatorname{Tr} \left[\nabla_{\theta}^{2} F(\theta^{*})^{-1} \Sigma \nabla_{\theta}^{2} F(\theta^{*})^{-1} \right]}{N_{t}}, \label{eq:thm:minkowski_1}
\end{align}
using \cref{assump:assg:L_lipschitz}.
To ease notation, we denote $\operatorname{Tr} [\nabla_{\theta}^{2} F(\theta^{*})^{-1} \Sigma \nabla_{\theta}^{2} F(\theta^{*})^{-1}]$ by $\Lambda$.
Next, note that for all $t \geq 1$, we have the relation in \cref{eq:thm:grad_f_theta_relation}, giving us
\begin{align*}
\frac{1}{N_{t}} \sum_{i=1}^{t} n_{i} \nabla_{\theta} f_{i} \left( \theta_{i-1} \right) 
= \frac{1}{N_{t}} \sum_{i=1}^{t} \frac{n_{i}}{\gamma_{i}} \left( \theta_{i-1} - \theta_{i} \right)
= \frac{1}{N_{t}} \sum_{i=1}^{t-1} \left( \theta_{i} - \theta^{*} \right) \left(\frac{n_{i+1}}{\gamma_{i+1}} - \frac{n_{i}}{\gamma_{i}} \right)
- \frac{1}{N_{t}} \left( \theta_{t} - \theta^{*} \right) \frac{n_{t}}{\gamma_{t}} 
+ \frac{1}{N_{t}} \left( \theta_{0} - \theta^{*} \right) \frac{n_{1}}{\gamma_{1}},
\end{align*}
leading to 
\begin{align*}
\left\lVert \nabla_{\theta}^{2} F \left( \theta^{*} \right)^{-1} \frac{1}{N_{t}} \sum_{i=1}^{t} n_{i} \nabla_{\theta} f_{i} \left( \theta_{i-1} \right)\right\rVert
\leq \frac{1}{N_{t} \mu} \sum_{i=1}^{t-1} \left\lVert \theta_{i} - \theta^{*}\right\rVert \left\vert \frac{n_{i+1}}{\gamma_{i+1}} - \frac{n_{i}}{\gamma_{i}} \right\vert
+ \frac{1}{N_{t} \mu} \left\lVert\theta_{t} - \theta^{*}\right\rVert \frac{n_{t}}{\gamma_{t}} 
+ \frac{1}{N_{t} \mu} \left\lVert\theta_{0} - \theta^{*}\right\rVert \frac{n_{1}}{\gamma_{1}}.
\end{align*}
Hence, with the notion of $\delta_{t}=\E[\lVert\theta_{t}-\theta^{*}\rVert^{2}]$ this expression can be simplified to
\begin{align}
\left(\E \left[ \left\lVert\nabla_{\theta}^{2} F \left( \theta^{*} \right)^{-1} \frac{1}{N_{t}} \sum_{i=1}^{t} n_{i} \nabla_{\theta} f_{i} \left( \theta_{i-1} \right)\right\rVert^{2} \right] \right)^{\frac{1}{2}}
\leq \frac{1}{N_{t} \mu} \sum_{i=1}^{t-1} \delta_{i}^{\frac{1}{2}} \left\vert \frac{n_{i+1}}{\gamma_{i+1}} - \frac{n_{i}}{\gamma_{i}} \right\vert
+ \frac{n_{t}}{N_{t} \gamma_{t} \mu} \delta_{t}^{\frac{1}{2}} 
+ \frac{n_{1}}{N_{t} \gamma_{1} \mu} \delta_{0}^{\frac{1}{2}}. \label{eq:thm:minkowski_2}
\end{align}
For the martingale term, we have
\begin{align}
\E &\left[ \left\lVert\nabla_{\theta}^{2} F \left( \theta^{*} \right)^{-1} \frac{1}{N_{t}} \sum_{i=1}^{t} n_{i} \left[\nabla_{\theta} f_{i} \left( \theta_{i-1} \right) - \nabla_{\theta} f_{i} \left( \theta^{*} \right) - \nabla_{\theta} F \left( \theta_{i-1} \right)\right]\right\rVert^{2} \right]
\leq \frac{1}{N_{t}^{2} \mu^{2}} \sum_{i=1}^{t} n_{i}^{2} \E \left[ \left\lVert\nabla_{\theta} f_{i} \left( \theta_{i-1} \right) - \nabla_{\theta} f_{i} \left( \theta^{*} \right)\right\rVert^{2} \right] \nonumber
\\ &= \frac{1}{N_{t}^{2} \mu^{2}} \sum_{i=1}^{t} \E \left[ \left\lVert \sum_{j=1}^{n_{i}} \nabla_{\theta} f_{i,j} \left( \theta_{i-1} \right) - \nabla_{\theta} f_{i,j} \left( \theta^{*} \right)\right\rVert^{2} \right] 
\leq \frac{1}{N_{t}^{2} \mu^{2}} \sum_{i=1}^{t} \sum_{j=1}^{n_{i}} \left( \E \left[ \left\lVert \nabla_{\theta} f_{i,j} \left( \theta_{i-1} \right) - \nabla_{\theta} f_{i,j} \left( \theta^{*} \right)\right\rVert^{2} \right] \right)^{\frac{1}{2}} \nonumber \\
&\leq \frac{C_{f}^{2}}{N_{t}^{2} \mu^{2}} \sum_{i=1}^{t} n_{i} \delta_{i-1}, \label{eq:thm:minkowski_3}
\end{align}
by Cauchy-Schwarz inequality and \cref{assump:ssg:f_lipschitz}.
For all $t \geq 1$, the rest term is directly bounded by \cref{eq:avg:assump:L_lipschitz}:
\begin{align}
\left( \E \left[ \left\lVert\nabla_{\theta}^{2} F \left( \theta^{*} \right)^{-1} \frac{1}{N_{t}} \sum_{i=1}^{t} n_{i} \left[ \nabla_{\theta} F \left( \theta_{i-1} \right) - \nabla_{\theta}^{2} F \left( \theta^{*} \right) \left( \theta_{i-1} - \theta^{*} \right)\right] \right\rVert^{2} \right] \right)^{\frac{1}{2}}
\leq& \frac{C_{\nabla}'}{N_{t} \mu} \sum_{i=1}^{t} n_{i} \Delta_{i-1}^{\frac{1}{2}}, \label{eq:thm:minkowski_4}
\end{align}
with the notion $\Delta_{t} = \E [\lVert\theta_{t} - \theta^{*}\rVert^{4} ]$.
Finally, combining the terms from \cref{eq:thm:minkowski_1,eq:thm:minkowski_2,eq:thm:minkowski_3,eq:thm:minkowski_4}, gives us
\begin{align} \label{eq:thm:minkowski_bound}
\bar{\delta}_{t}^{1/2}
\leq& 
\frac{\Lambda^{1/2}}{N_{t}^{1/2}}
+\frac{1}{N_{t}\mu} \sum_{i=1}^{t-1} \delta_{i}^{1/2} \left\vert \frac{n_{i+1}}{\gamma_{i+1}} - \frac{n_{i}}{\gamma_{i}} \right\vert
+ \frac{n_{t}}{N_{t} \gamma_{t} \mu} \delta_{t}^{1/2} 
+ \frac{n_{1}}{N_{t} \gamma_{1} \mu} \delta_{0}^{1/2}
+ \frac{C_{f}}{N_{t} \mu} \left( \sum_{i=1}^{t} n_{i} \delta_{i-1} \right)^{1/2}
+ \frac{C_{\nabla}'}{N_{t} \mu} \sum_{i=1}^{t} n_{i} \Delta_{i-1}^{1/2},
\end{align}
where $\bar{\delta}_{t} = \E[\lVert\bar{\theta}_{t}-\theta^{*}\rVert^{2}]$, which can be simplified into \cref{eq:avg:thm:nonblock:streaming_grad_upper_bound} by shifting the indices and collecting the $\delta_{0}$ terms.
\end{proof}

\begin{proof}[Proof of \cref{cor:avg:streaming_grad_upper_bound_constant}]
As $n_{t}=C_{\rho}$ for all $t\geq1$, we simplify the bound for $\bar{\delta}_{t}^{1/2}$ in \cref{eq:avg:thm:nonblock:streaming_grad_upper_bound} to
\begin{align} \label{eq:avg:thm:nonblock:streaming_grad_upper_bound_constant}
\frac{\Lambda^{1/2}}{N_{t}^{1/2}}
+\frac{C_{\rho}}{N_{t}\mu} \sum_{i=1}^{t-1} \delta_{i}^{1/2} \left\vert \frac{1}{\gamma_{i+1}} - \frac{1}{\gamma_{i}} \right\vert
+ \frac{C_{\rho}}{N_{t} \gamma_{t} \mu} \delta_{t}^{1/2} 
+ \frac{C_{\rho}}{N_{t} \mu} \left( \frac{1}{\gamma_{1}} + C_{f} \right) \delta_{0}^{1/2} 
+ \frac{C_{f} C_{\rho}^{\frac{1}{2}}}{N_{t} \mu} \left( \sum_{i=1}^{t-1} \delta_{i} \right)^{1/2}
+ \frac{C_{\nabla}' C_{\rho}}{N_{t} \mu} \sum_{i=0}^{t-1} \Delta_{i}^{1/2}.
\end{align}
The second-order moment $\delta_{t}$ is bounded by \cref{cor:nonblock:streaming_grad_upper_bound_constant} but with use of \cref{eq:cor:nonblock:streaming_grad_upper_bound_constant_n_tterm} as we work in terms of $t$.
The fourth-order moment $\Delta_{t}$ from \cref{lem:avg:nonblock:lrate4} can be simplified to:
\begin{align*}
\Delta_{t} 
\leq& \exp \left( - \mu \sum_{i=t/2}^{t} \gamma_{i} \right)  \Pi_{\infty}^{c}
+ \frac{1}{\mu} \left( \frac{32 \sigma^{4}}{\mu C_{\rho}^{2}} \max_{t/2 \leq i \leq t} \gamma_{i}^{2} 
+  \frac{48 \sigma^{4}}{C_{\rho}^{3}} \max_{t/2 \leq i \leq t} \gamma_{i}^{3}
+ \frac{114 \sigma^{4} \indicator{C_{\rho} > 1}}{C_{\rho}^{2}} \max_{t/2 \leq i \leq t} \gamma_{i}^{3} \right)
\\ \leq& \exp \left( - \frac{\mu C_{\gamma} C_{\rho}^{\beta} t^{1-\alpha}}{2^{1-\alpha}}\right) \Pi_{\infty}^{c}
+ \frac{1}{\mu} \left( \frac{2^{2\alpha}32 \sigma^{4} C_{\gamma}^{2} C_{\rho}^{2\beta}}{\mu C_{\rho}^{2} t^{2\alpha}}
+  \frac{2^{3\alpha}48 \sigma^{4} C_{\gamma}^{3} C_{\rho}^{3 \beta}}{C_{\rho}^{3} t^{3\alpha}}
+ \frac{2^{3\alpha}114 \sigma^{4} C_{\gamma}^{3} C_{\rho}^{3 \beta} \indicator{C_{\rho} > 1}}{C_{\rho}^{2} t^{3 \alpha}} \right),
\end{align*}
using that $\gamma_{t} = C_{\gamma} C_{\rho}^{\beta} t^{-\alpha}$ is decreasing as $\alpha \in (1/2, 1 )$.
Regarding $\Pi_{t}^{\Delta}$ defined in \cref{eq:avg:4momoent:pi}, we can bound it by
\begin{align*}
\Pi_{\infty}^{c} =& \exp \left( \frac{64 \alpha C_{f}^{2} C_{\gamma}^{2} C_{\rho}^{2\beta}}{\left( 2 \alpha -1 \right) C_{\rho}} \right)
\exp \left(\frac{(192+456C_{\rho}\indicator{C_{\rho}>1}) C_{f}^{4} C_{\gamma}^{4} C_{\rho}^{4 \beta}}{C_{\rho}^{3}} \right)
\exp \left( \frac{32 \alpha C_{\nabla}^{2} C_{\gamma}^{2} C_{\rho}^{2\beta} \indicator{C_{\rho} > 1}}{2 \alpha -1} \right)
\\ &\exp \left( 192 C_{\nabla}^{4} C_{\gamma}^{4} C_{\rho}^{4 \beta} \indicator{C_{\rho} > 1} \right)\left( \Delta_{0} + \frac{2\sigma^{4}}{C_{f}^{4}} + \frac{4\sigma^{4}C_{\gamma}}{ \mu C_{f}^{2}C_{\rho}^{1-\beta}} \right),
\end{align*}
using $\sum_{i=1}^{t} i^{-2\alpha} \leq 2\alpha/(2\alpha -1)$ and $\sum_{i=1}^{t} i^{-4\alpha} \leq 2$.
Note that $\Pi_{\infty}^{c}$ is a finite constant, independent of $t$.
To bound the first term of \cref{eq:avg:thm:nonblock:streaming_grad_upper_bound_constant}, namely $\frac{C_{\rho}}{N_{t}\mu} \sum_{i=1}^{t-1} \delta_{i}^{1/2} \lvert \gamma_{i+1}^{-1} - \gamma_{i}^{-1} \rvert$, we remark that $\lvert \gamma_{t+1}^{-1} - \gamma_{t}^{-1} \rvert \leq C_{\gamma}^{-1}C_{\rho}^{-\beta}\alpha t^{\alpha -1}$, one has (since $\sqrt{a+b} \leq \sqrt{a} + \sqrt{b}$),
\begin{align} \label{eq:avg:constant_sb_bound_1}
\frac{C_{\rho}}{N_{t}\mu} \sum_{i=1}^{t-1} \delta_{i}^{\frac{1}{2}} \left\vert \frac{1}{\gamma_{i+1}} - \frac{1}{\gamma_{i}} \right\vert 
\leq&  \frac{C_{\rho}^{1-\beta}\alpha}{C_{\gamma}\mu N_{t}} 
\sum_{i=1}^{t} i^{\alpha -1} 
\left( 
\exp \left( - \frac{\mu C_{\gamma}C_{\rho}^{\beta} i^{1-\alpha}}{2^{2-\alpha}}\right) \sqrt{\pi_{\infty}^{c}}
+ \frac{2^{\frac{1+\alpha}{2}}\sigma\sqrt{C_{\gamma}}} {\sqrt{\mu}C_{\rho}^{\frac{1-\beta}{2}}i^{\alpha /2}} 
\right).
\end{align}
For simplicity, let us denote 
\begin{align*}
A_{\infty}^{c}
= \sum_{i=0}^{\infty} \exp \left( - \frac{\mu C_{\gamma}C_{\rho}^{\beta} i^{1-\alpha}}{2^{2-\alpha}} \right)
\geq \sum_{i=0}^{\infty} i^{\alpha -1} \exp \left( - \frac{\mu C_{\gamma}C_{\rho}^{\beta} i^{1-\alpha}}{2^{2-\alpha}}\right),
\end{align*}
as $\alpha < 1$.
Thus, the first part of \cref{eq:avg:constant_sb_bound_1} is bounded as follows:
\begin{align*}
\frac{C_{\rho}^{1-\beta}\alpha\sqrt{\pi_{\infty}^{c}}}{C_{\gamma} \mu N_{t}} 
\sum_{i=1}^{t} i^{\alpha -1}
\exp \left( - \frac{\mu C_{\gamma} C_{\rho}^{\beta} i^{1-\alpha}}{2^{2-\alpha}}\right)
\leq \frac{C_{\rho}^{1-\beta}\alpha \sqrt{\pi_{\infty}^{c}} A_{\infty}^{c}}{C_{\gamma} \mu N_{t}}.
\end{align*}
Furthermore, with the help of an integral test for convergence, one has
$\sum_{i=1}^{t}i^{\alpha /2-1} \leq 1 + \int_{1}^{t} s^{\alpha/2-1} \, ds = 1 + (2/\alpha)t^{\alpha /2} - (2/\alpha) \leq (2/\alpha)t^{\alpha /2}$, such that the second part of \cref{eq:avg:constant_sb_bound_1} can be bounded by
\begin{align*}
\frac{2^{\frac{1+\alpha}{2}} \sigma C_{\rho}^{\frac{1-\beta}{2}} \alpha}{C_{\gamma}^{1/2} \mu^{3/2} N_{t}} \sum_{i=1}^{t} i^{\alpha /2-1} 
\leq \frac{2^{\frac{3+\alpha}{2}} \sigma C_{\rho}^{\frac{1-\beta}{2}}  t^{\alpha/2}}{C_{\gamma}^{1/2} \mu^{3/2} N_{t}} 
= \frac{2^{\frac{3+\alpha}{2}} \sigma C_{\rho}^{\frac{1-\alpha-\beta}{2}} }{C_{\gamma}^{1/2} \mu^{3/2} N_{t}^{1-\alpha/2}} .
\end{align*}
By combining this, we get
\begin{align}
\frac{C_{\rho}}{N_{t}\mu} \sum_{i=1}^{t-1} \delta_{i}^{\frac{1}{2}} \left\vert \frac{1}{\gamma_{i+1}} - \frac{1}{\gamma_{i}} \right\vert 
\leq \frac{C_{\rho}^{1-\beta}\alpha \sqrt{\pi_{\infty}^{c}} A_{\infty}^{c}}{C_{\gamma} \mu N_{t}} 
+ \frac{2^{\frac{3+\alpha}{2}} \sigma C_{\rho}^{\frac{1-\alpha-\beta}{2}} }{\sqrt{C_{\gamma}}\mu^{3/2}N_{t}^{1-\alpha /2}} .
\end{align}
Similarly, second term of \cref{eq:avg:thm:nonblock:streaming_grad_upper_bound_constant}, can be bounded by
\begin{align*}
\frac{C_{\rho}}{N_{t} \gamma_{t} \mu} \delta_{t}^{\frac{1}{2}} 
\leq& \frac{C_{\rho}^{1-\alpha-\beta}}{ C_{\gamma} \mu N_{t}^{1-\alpha}} 
\left( \exp \left( - \frac{\mu C_{\gamma}C_{\rho}^{\beta}t^{1-\alpha}}{2^{2-\alpha}} \right) \sqrt{\pi_{\infty}^{c}}
+ \frac{2^{\frac{1+\alpha}{2}}\sigma \sqrt{C_{\gamma}}}{\sqrt{\mu} C_{\rho}^{\frac{1-\beta}{2}}t^{\alpha /2}} \right) 
\leq \frac{C_{\rho}^{2-\alpha-\beta} \sqrt{\pi_{\infty}^{c}} A_{\infty}^{c}}{ C_{\gamma} \mu N_{t}^{2-\alpha}} 
+ \frac{2^{\frac{1+\alpha}{2}} C_{\rho}^{\frac{1-\alpha-\beta}{2}}\sigma}{ \sqrt{C_{\gamma}} \mu^{3/2} N_{t}^{1-\alpha/2}},
\end{align*}
using $\exp(-\mu C_{\gamma}C_{\rho}^{\beta}t^{1-\alpha}/2^{2-\alpha})=A_{t}^{c} \leq t^{-1}\sum_{i=1}^{t}A_{i}^{c}\leq t^{-1}A_{\infty}^{c}$ as $A_{t}^{c}$ is decreasing.
In a same way, one has
\begin{align*}
\frac{C_{f} C_{\rho}^{\frac{1}{2}}}{N_{t} \mu} 
\left( \sum_{i=1}^{t-1} \delta_{i} \right)^{\frac{1}{2}}
\leq& \frac{C_{f} C_{\rho}^{\frac{1}{2}}}{N_{t} \mu} 
\left( A_{\infty}^{c} \pi_{\infty}^{c}
+ \frac{2^{1+\alpha} \sigma^{2} C_{\gamma} t^{1-\alpha}}{\left( 1 - \alpha \right) \mu C_{\rho}^{1-\beta} }  \right)^{1/2}  
\leq \frac{C_{f} C_{\rho}^{\frac{1}{2}} \sqrt{\pi_{\infty}^{c}} \sqrt{A_{\infty}^{c}} }{N_{t} \mu} + \frac{2^{\frac{1+\alpha}{2}} C_{f} \sigma \sqrt{C_{\gamma}}}{C_{\rho}^{\frac{1-\alpha-\beta}{2}} \mu^{3/2} N_{t}^{\frac{1+\alpha}{2}}}.
\end{align*}
Bound the last term of \cref{eq:avg:thm:nonblock:streaming_grad_upper_bound_constant}, is done as follows,
\begin{align*}
\frac{C_{\nabla}' C_{\rho}}{N_{t} \mu} \sum_{i=0}^{t-1} \Delta_{i}^{\frac{1}{2}}
\leq& \frac{C_{\nabla}' C_{\rho}}{N_{t} \mu} \sum_{i=0}^{t-1} 
\exp \left( - \frac{\mu C_{\gamma} C_{\rho}^{\beta} i^{1-\alpha}}{2^{2-\alpha}} \right) \sqrt{\Pi_{\infty}^{c}}
+ \frac{2^{\alpha} 6 C_{\nabla}' \sigma^{2} C_{\gamma} C_{\rho}^{\beta}}{N_{t} \mu^{2}} \sum_{i=1}^{t-1} i^{-\alpha}
\\ &+  \frac{(6+7\indicator{C_{\rho} > 1})2^{3\alpha/2} C_{\nabla}' \sigma^{2} C_{\gamma}^{3/2} C_{\rho}^{3\beta/2}}{N_{t} \mu^{3/2}} \sum_{i=1}^{t-1} i^{-3\alpha/2}
\\ \leq& 
\frac{C_{\nabla}' C_{\rho}\sqrt{\Pi_{\infty}^{c}} A_{\infty}^{c}}{N_{t} \mu} 
+ \frac{2^{\alpha} 6 C_{\nabla}' \sigma^{2} C_{\gamma}}{C_{\rho}^{1-\alpha-\beta} \mu^{2} N_{t}^{\alpha}}
+ \frac{(6+7\indicator{C_{\rho} > 1}) 2^{3\alpha/2} C_{\nabla}'  \sigma^{2}C_{\gamma}^{3/2}C_{\rho}^{3\beta /2} \psi_{3\alpha/2}^{0}(N_{t}/C_{\rho})}{\mu^{3/2}N_{t}}.
\end{align*}
Thus, by collecting the terms above, we obtain:
\begin{align*}
\bar{\delta}_{t}^{1/2} 
\leq& \frac{\Lambda^{1/2}}{N_{t}^{1/2}}  
+ \frac{6 \sigma C_{\rho}^{\frac{1-\alpha-\beta}{2}} }{\sqrt{C_{\gamma}}\mu^{3/2}N_{t}^{1-\alpha /2}}
+ \frac{2^{\alpha} 6 C_{\nabla}' \sigma^{2} C_{\gamma}}{C_{\rho}^{1-\alpha-\beta} \mu^{2} N_{t}^{\alpha}}
+ \frac{C_{\rho}^{2-\alpha-\beta} \sqrt{\pi_{\infty}^{c}} A_{\infty}^{c}}{ C_{\gamma} \mu N_{t}^{2-\alpha}}
\\ &+ \frac{2^{\frac{1+\alpha}{2}} C_{f} \sigma \sqrt{C_{\gamma}}}{C_{\rho}^{\frac{1-\alpha-\beta}{2}} \mu^{3/2} N_{t}^{\frac{1+\alpha}{2}}} + \frac{C_{\rho} \Gamma_{c}}{\mu N_{t}} 
+ \frac{(6+7 \indicator{C_{\rho} > 1}) 2^{3\alpha/2} C_{\nabla}'  \sigma^{2}C_{\gamma}^{3/2}C_{\rho}^{3\beta/2}}{\mu^{3/2}\psi_{3\alpha/2}^{0}(N_{t}/C_{\rho})^{-1}N_{t}},
\end{align*}
where $\Gamma_{c} =(1/C_{\gamma}C_{\rho}^{\beta}+C_{f})\delta_{0}^{1/2} + C_{f}\sqrt{\pi_{\infty}^{c}A_{\infty}^{c}}/C_{\rho}^{1/2} +\sqrt{\pi_{\infty}^{c}} A_{\infty}^{c}/C_{\gamma} C_{\rho}^{\beta} + C_{\nabla}' \sqrt{\Pi_{\infty}^{c}} A_{\infty}^{c}$.
\end{proof}

\begin{proof}[Proof of \cref{cor:avg:streaming_grad_upper_bound_increasing}]
 The steps of the proof follows the ones of \cref{cor:avg:streaming_grad_upper_bound_constant} with the smart notation of $\phi$ and $\tilde{\rho}$:
The bound for $\bar{\delta}_{t}^{1/2}$ in \cref{eq:avg:thm:nonblock:streaming_grad_upper_bound} is given by
\begin{align} \label{eq:avg:thm:nonblock:streaming_grad_upper_bound_increasing} 
\frac{\Lambda^{1/2}}{N_{t}^{1/2}}
+\frac{1}{N_{t}\mu} \sum_{i=1}^{t-1} \delta_{i}^{1/2} \left\vert \frac{n_{i+1}}{\gamma_{i+1}} - \frac{n_{i}}{\gamma_{i}} \right\vert
+ \frac{n_{t}}{N_{t} \gamma_{t} \mu} \delta_{t}^{1/2} 
+ \frac{n_{1}}{N_{t} \mu} \left( \frac{1}{\gamma_{1}} + C_{f} \right) \delta_{0}^{1/2} 
+ \frac{C_{f}}{N_{t} \mu} \left( \sum_{i=1}^{t-1} n_{i+1} \delta_{i} \right)^{1/2}
+ \frac{C_{\nabla}'}{N_{t} \mu} \sum_{i=0}^{t-1} n_{i+1} \Delta_{i}^{1/2},
\end{align}
where the learning rate and time-varying mini-batches are on the form $\gamma_{t} = C_{\gamma} n_{t}^{\beta} t^{-\alpha}$ and $n_{t} = \lceil C_{\rho} t^{\rho} \rceil$.
The second-order moment $\delta_{t}$ is upper bounded by \cref{eq:ssg:varying:t} from \cref{cor:nonblock:learning_rate_tau_1}.
The fourth-order moment $\Delta_{t}$ from \cref{lem:avg:nonblock:lrate4} can be simplified as follows,
\begin{align*}
\Delta_{t} 
 \leq& \exp \left( - \mu \sum_{i=t/2}^{t} \gamma_{i} \right) \Pi_{\infty}^{v}
+ \frac{32 \sigma^{4}}{\mu^{2}} \max_{t/2 \leq i \leq t} \frac{\gamma_{i}^{2}}{n_{i}^{2}} 
+ \frac{162 \sigma^{4}}{\mu} \max_{t/2 \leq i \leq t} \frac{\gamma_{i}^{3}}{n_{i}^{2}},
\end{align*}
as $n_{t}\geq1$ for any $t\geq1$ and $\beta\leq1$, and
\begin{align*}
\Pi_{\infty}^{v} =& \exp \left( \frac{32(\alpha-\beta\tilde{\rho}) C_{\gamma}^{2} C_{\rho}^{2\beta} ( 2C_{f}^{2}+C_{\nabla}^{2})}{2 (\alpha-\beta\tilde{\rho}) -1} \right)
\exp \left( 192 C_{\gamma}^{4} C_{\rho}^{4\beta} (4C_{f}^{4}+C_{\nabla}^{4}) \right)\left( \Delta_{0} 
+ \frac{2\sigma^{4}}{C_{f}^{4}} + \frac{4\sigma^{4}C_{\gamma}}{ \mu C_{f}^{2} C_{\rho}^{1-\beta}},
\right)
\end{align*}
using that $\sum_{i=1}^{t} i^{-a} \leq 2$ for $a\geq 2$.
Next, for $\rho \geq 0$, we have
\begin{align*}
\Delta_{t} 
\leq& \exp \left( - \frac{\mu C_{\gamma} C_{\rho}^{\beta} t^{1+\beta \rho-\alpha}}{2^{1 + \beta \rho - \alpha}} \right) \Pi_{\infty}^{v}
+ \frac{2^{2\alpha-2\beta\rho+2\rho} 32  \sigma^{4} C_{\gamma}^{2} C_{\rho}^{2\beta}}{\mu^{2} C_{\rho}^{2} t^{2\alpha-2\beta\rho+2\rho}} 
+ \frac{2^{3\alpha-3\beta\rho+2\rho} 162 \sigma^{4} C_{\gamma}^{3} C_{\rho}^{3\beta}}{\mu C_{\rho}^{2} t^{3\alpha-3\beta\rho+2\rho}},
\end{align*}
using that $\alpha - \beta \rho \in (1/2, 1)$.
If $\rho < 0$, one directly have
\begin{align*}
\Delta_{t}
\leq& \exp \left( - \frac{\mu C_{\gamma} C_{\rho}^{\beta} t^{1-\alpha}}{2^{1-\alpha}} \right) \Pi_{\infty}^{v}
+ \frac{2^{2\alpha} 32 \sigma^{4} C_{\gamma}^{2} C_{\rho}^{2\beta}}{ \mu^{2} t^{2\alpha}} 
+ \frac{2^{3\alpha} 162 \sigma^{4} C_{\gamma}^{3} C_{\rho}^{3\beta}}{\mu t^{3\alpha}}.
\end{align*}
With the notion of $\phi$ and $\tilde{\rho}$, we can combine the two $\rho$-cases as follows:
\begin{align*}
\Delta_{t} 
\leq& \exp \left( - \frac{\mu C_{\gamma} C_{\rho}^{\beta\indicator{\rho\geq0}} t^{(1-\phi)(1+\tilde{\rho})}}{2^{(1-\phi)(1+\tilde{\rho})}} \right) \Pi_{\infty}^{v}
+ \frac{2^{2\phi(1+\tilde{\rho})} 32  \sigma^{4} C_{\gamma}^{2} C_{\rho}^{2\beta}}{\mu^{2} C_{\rho}^{2\indicator{\rho\geq0}} t^{2\phi(1+\tilde{\rho})}} 
+ \frac{2^{3\phi(1+\tilde{\rho})-\tilde{\rho}} 162 \sigma^{4} C_{\gamma}^{3} C_{\rho}^{3\beta}}{\mu C_{\rho}^{2\indicator{\rho\geq0} } t^{3\phi(1+\tilde{\rho})-\tilde{\rho}}}.
\end{align*}

We will in the following bound the terms for $t$ but afterwards we will translate it to terms in $N_{t}$.
If $\rho \geq 0$,
the first relation is $t \geq (N_{t}/2 C_{\rho})^{1/(1+\rho)}$, e.g., see the proof of \cref{cor:nonblock:learning_rate_tau_1}.
Similarly, $N_{t} \geq C_{\rho} \sum_{i=1}^{t} i^{\rho} \geq C_{\rho} \int_{0}^{t} x^{\rho} \, dx = C_{\rho} t^{\rho+1}$, thus, $t \leq ( N_{t}/C_{\rho} )^{1/(1+\rho)}$. If $\rho < 0$, one has $t\leq N_{t}$ and $N_{t} \leq C_{\rho}t$, i.e., $t\geq N_{t}/C_{\rho}$.

Bounding $\frac{1}{N_{t}\mu} \sum_{i=1}^{t-1} \delta_{i}^{1/2} \lvert n_{i+1}/\gamma_{i+1} - n_{i}/\gamma_{i} \lvert$, we first note that $n_{t}/\gamma_{t} = C_{\gamma}^{-1} \lceil C_{\rho}t^{\rho}\rceil^{1-\beta}t^{\alpha}$. 
Thus, by the mean value inequality, we obtain for $\rho\geq0$:
\begin{align}
\left\vert \frac{n_{i+1}}{\gamma_{i+1}} - \frac{n_{i}}{\gamma_{i}} \right\vert
\leq \frac{2C_{\rho}^{1-\beta}}{C_{\gamma}} \sup_{\nu \in (i,i+1)}\left\vert \nu^{(1-\beta)\rho + \alpha -1} \right\vert
\leq \frac{2C_{\rho}^{1-\beta}}{C_{\gamma} i^{1-(1-\beta)\rho-\alpha}},
\end{align}
as $\alpha+(1-\beta)\rho\leq1-\rho$ since $\alpha-\beta\rho \in (1/2,1)$.
For $\rho < 0$, the mean value inequality gives us
\begin{align*}
\left\vert \frac{n_{i+1}}{\gamma_{i+1}} - \frac{n_{i}}{\gamma_{i}} \right\vert
\leq \frac{2C_{\rho}^{1-\beta}}{C_{\gamma}} \sup_{\nu \in (i,i+1)}\left\vert \nu^{\alpha -1} \right\vert
\leq \frac{2C_{\rho}^{1-\beta}}{C_{\gamma} i^{1-\alpha}},
\end{align*}
as $(n_{t})_{t\geq1}$ is a decreasing sequence and $\beta\leq1$.
Thus, for any $\rho \in (-1,1)$, we have
\begin{align*}
\left\vert \frac{n_{i+1}}{\gamma_{i+1}} - \frac{n_{i}}{\gamma_{i}} \right\vert
\leq \frac{2C_{\rho}^{1-\beta}}{C_{\gamma} i^{1-\phi(1+\tilde{\rho})}}.
\end{align*}
By using this, we obtain a bound on $\frac{1}{N_{t}\mu} \sum_{i=1}^{t-1} \delta_{i}^{\frac{1}{2}} \left\vert \frac{n_{i+1}}{\gamma_{i+1}} - \frac{n_{i}}{\gamma_{i}} \right\vert$ given as
\begin{align*}
\frac{2C_{\rho}^{1-\beta}}{N_{t}\mu C_{\gamma}} \sum_{i=1}^{t} i^{\phi(1+\tilde{\rho})-1}
\left( \exp \left( - \frac{\mu C_{\gamma} C_{\rho}^{\beta\indicator{\rho\geq0}} i^{(1-\phi)(1+\tilde{\rho})}}{2^{1+(1-\phi)(1+\tilde{\rho})}} \right) \sqrt{\pi_{\infty}^{v}} + \frac{2^{\frac{1+\phi(1+\tilde{\rho})}{2}}\sigma \sqrt{C_{\gamma}}}{\sqrt{\mu} C_{\rho}^{\frac{(1-\beta)}{2}\indicator{\rho\geq0}} i^{\frac{\phi(1+\tilde{\rho})}{2}}} \right).
\end{align*}
Next, let us denote
\begin{align*}
A_{\infty}^{v} =& \sum_{i=0}^{\infty} i^{\tilde{\rho}} \exp \left( - \frac{\mu C_{\gamma} C_{\rho}^{\beta\indicator{\rho\geq0}} i^{(1-\phi)(1+\tilde{\rho})}}{2^{1+(1-\phi)(1+\tilde{\rho})}} \right)
\geq \sum_{i=0}^{\infty} i^{\phi(1+\tilde{\rho})-1} \exp \left( - \frac{\mu C_{\gamma} C_{\rho}^{\beta\indicator{\rho\geq0}} i^{(1-\phi)(1+\tilde{\rho})}}{2^{1+(1-\phi)(1+\tilde{\rho})}} \right),
\end{align*}
since $\phi(1+\tilde{\rho})-1 = \alpha+(1-\beta)\tilde{\rho}-1 \leq \tilde{\rho}$.
Thus,
\begin{align*}
\frac{2C_{\rho}^{1-\beta} \sqrt{\pi_{\infty}^{v}}}{N_{t}\mu C_{\gamma}} \sum_{i=1}^{t} i^{\phi(1+\tilde{\rho})-1}
 \exp \left( - \frac{\mu C_{\gamma} C_{\rho}^{\beta\indicator{\rho\geq0}} i^{(1-\phi)(1+\tilde{\rho})}}{2^{1+(1-\phi)(1+\tilde{\rho})}} \right)
\leq \frac{2C_{\rho}^{1-\beta} \sqrt{\pi_{\infty}^{v}} A_{\infty}^{v}}{N_{t}\mu C_{\gamma}}.
\end{align*}
Furthermore, with the help of an integral test for convergence, we have
\begin{align*}
\frac{2^{\frac{3+\phi(1+\tilde{\rho})}{2}} \sigma C_{\rho}^{\frac{1-\beta}{2}\indicator{\rho\geq0}}}{\mu^{3/2} \sqrt{C_{\gamma}} N_{t}} \sum_{i=1}^{t} i^{\frac{\phi(1+\tilde{\rho})}{2}-1}
\leq& \frac{2^{\frac{3+\phi(1+\tilde{\rho})}{2}} \sigma C_{\rho}^{\frac{1-\beta}{2}\indicator{\rho\geq0}} t^{\frac{\phi(1+\tilde{\rho})}{2}}}{\mu^{3/2} \sqrt{C_{\gamma}} N_{t}}
\leq \frac{2^{\frac{3+\phi(1+\tilde{\rho})}{2}} \sigma C_{\rho}^{\frac{1-\phi-\beta}{2}\indicator{\rho\geq0}}}{\mu^{3/2} \sqrt{C_{\gamma}} N_{t}^{1-\phi/2}}.
\end{align*}
Summarising, we obtain
\begin{align*}
\frac{1}{N_{t}\mu} \sum_{i=1}^{t-1} \delta_{i}^{\frac{1}{2}} \left\vert \frac{n_{i+1}}{\gamma_{i+1}} - \frac{n_{i}}{\gamma_{i}} \right\vert
\leq&  \frac{2C_{\rho}^{1-\beta} \sqrt{\pi_{\infty}^{v}} A_{\infty}^{v}}{N_{t}\mu C_{\gamma}}
+ \frac{2^{\frac{3+\phi(1+\tilde{\rho})}{2}} \sigma C_{\rho}^{\frac{1-\phi-\beta}{2}\indicator{\rho\geq0}}}{\mu^{3/2} \sqrt{C_{\gamma}} N_{t}^{1-\phi/2}}.
\end{align*}
Similarly, for $\frac{n_{t}}{N_{t} \gamma_{t} \mu} \delta_{t}^{1/2}$, one have
\begin{align*}
\frac{n_{t}}{N_{t} \gamma_{t} \mu} \delta_{t}^{\frac{1}{2}}
\leq& \frac{C_{\rho}^{1-\beta} \sqrt{\pi_{\infty}^{v}} t^{\phi(1+\tilde{\rho})}}{N_{t} C_{\gamma} \mu} \exp \left( - \frac{\mu C_{\gamma} C_{\rho}^{\beta\indicator{\rho\geq0}} t^{(1-\phi)(1+\tilde{\rho})}}{2^{1+(1-\phi)(1+\tilde{\rho})}} \right)
+ \frac{2^{\frac{1+\phi(1+\tilde{\rho})}{2}} \sigma C_{\rho}^{\frac{1-\beta}{2}\indicator{\rho\geq0}} t^{\frac{\phi(1+\tilde{\rho})}{2}}}{\mu^{3/2} \sqrt{C_{\gamma}} N_{t}}
\\ \leq& \frac{C_{\rho}^{2-\phi-\beta} \sqrt{\pi_{\infty}^{v}} A_{\infty}^{v}}{\mu C_{\gamma} N_{t}^{2-\phi}} 
+ \frac{2^{\frac{1+\phi(1+\tilde{\rho})}{2}} \sigma C_{\rho}^{\frac{1-\phi-\beta}{2}\indicator{\rho\geq0}}}{\mu^{3/2} \sqrt{C_{\gamma}} N_{t}^{1-\phi/2}}.
\end{align*}
For $\frac{n_{1}}{N_{t}\mu} (\gamma_{1}^{-1}+C_{f}) \delta_{0}^{1/2}$, we insert the definition of our learning functions, giving us
\begin{align*}
\frac{n_{1}}{N_{t} \mu} \left( \frac{1}{\gamma_{1}} + C_{f} \right) \delta_{0}^{1/2} 
= \frac{C_{\rho}}{N_{t} \mu} \left( \frac{1}{C_{\gamma} C_{\rho}^{\beta}} + C_{f} \right) \delta_{0}^{1/2}. 
\end{align*}
Bounding $\frac{C_{f}}{N_{t} \mu} ( \sum_{i=1}^{t-1} n_{i+1} \delta_{i})^{1/2}$, follows the ideas from above, using that $n_{t+1} \leq 2^{\tilde{\rho}} n_{t}$; it can be upper bounded by
\begin{align*}
&\frac{2^{\tilde{\rho}/2} C_{f} }{N_{t} \mu} \left( C_{\rho} \sum_{i=1}^{t} i^{\tilde{\rho}} \left( \exp \left( - \frac{\mu C_{\gamma} C_{\rho}^{\beta\indicator{\rho\geq0}} i^{(1-\phi)(1+\tilde{\rho})}}{2^{(1-\phi)(1+\tilde{\rho})}} \right) \pi_{\infty}^{v} + \frac{2^{1+\phi(1+\tilde{\rho})}\sigma^{2}C_{\gamma}}{\mu C_{\rho}^{(1-\beta)\indicator{\rho\geq0}} i^{\phi(1+\tilde{\rho})}} \right) \right)^{\frac{1}{2}}
\\ =& \frac{2^{\tilde{\rho}/2} C_{f} }{N_{t} \mu} \left( C_{\rho} \pi_{\infty}^{v} \sum_{i=1}^{t} i^{\tilde{\rho}} \exp \left( - \frac{\mu C_{\gamma} C_{\rho}^{\beta\indicator{\rho\geq0}} i^{(1-\phi)(1+\tilde{\rho})}}{2^{(1-\phi)(1+\tilde{\rho})}} \right)
+ \frac{2^{1+\phi(1+\tilde{\rho})} \sigma^{2} C_{\gamma} C_{\rho}^{\beta\indicator{\rho\geq0}}}{\mu} \sum_{i=1}^{t} i^{\beta\tilde{\rho}-\alpha} \right)^{\frac{1}{2}}
\\ \leq& \frac{2^{\tilde{\rho}/2} C_{f} }{N_{t} \mu} \left( C_{\rho} \pi_{\infty}^{v} A_{\infty}^{v}
+ \frac{2^{\phi(1+\tilde{\rho})} \sigma^{2} C_{\gamma} C_{\rho}^{\beta\indicator{\rho\geq0}} t^{(1-\phi)(1+\tilde{\rho})}}{\mu} \right)^{\frac{1}{2}}
\\ \leq& \frac{2^{\tilde{\rho}/2} C_{f} \sqrt{C_{\rho}} \sqrt{\pi_{\infty}^{v}} \sqrt{A_{\infty}^{v}}}{\mu N_{t}} 
+ \frac{2^{\frac{\phi(1+\tilde{\rho})}{2}} C_{f} \sigma \sqrt{C_{\gamma}} C_{\rho}^{\beta/2\indicator{\rho\geq0}} t^{\frac{(1-\phi)(1+\tilde{\rho})}{2}}}{\mu^{3/2} N_{t}}
\\ \leq& \frac{2^{\tilde{\rho}/2} C_{f} \sqrt{C_{\rho}} \sqrt{\pi_{\infty}^{v}} \sqrt{A_{\infty}^{v}}}{\mu N_{t}} 
+ \frac{2^{\frac{\phi(1+\tilde{\rho})}{2}} C_{f} \sigma \sqrt{C_{\gamma}}}{\mu^{3/2} C_{\rho}^{\frac{1-\phi-\beta}{2}\indicator{\rho\geq0}} N_{t}^{\frac{1+\phi}{2}}}.
\end{align*}
Likewise, for $\frac{C_{\nabla}'}{N_{t} \mu} \sum_{i=0}^{t-1} n_{i+1} \Delta_{i}^{1/2}$, we can bound by
\begin{align*}
&\frac{2^{\tilde{\rho}} C_{\nabla}' C_{\rho}}{N_{t} \mu} \sum_{i=1}^{t-1} i^{\tilde{\rho}} \left( \exp \left( - \frac{\mu C_{\gamma} C_{\rho}^{\beta\indicator{\rho\geq0}} i^{(1-\phi)(1+\tilde{\rho})}}{2^{(1-\phi)(1+\tilde{\rho})}} \right) \Pi_{\infty}^{v}
+ \frac{2^{2\phi(1+\tilde{\rho})} 32  \sigma^{4} C_{\gamma}^{2} C_{\rho}^{2\beta}}{\mu^{2} C_{\rho}^{2\indicator{\rho\geq0}} i^{2\phi(1+\tilde{\rho})}} 
+ \frac{2^{3\phi(1+\tilde{\rho})-\tilde{\rho}} 162 \sigma^{4} C_{\gamma}^{3} C_{\rho}^{3\beta}}{\mu C_{\rho}^{2\indicator{\rho\geq0} } i^{3\phi(1+\tilde{\rho})-\tilde{\rho}}} \right)^{\frac{1}{2}}
\\&\leq \frac{2^{\tilde{\rho}} C_{\nabla}' C_{\rho}}{N_{t} \mu} \sum_{i=1}^{t-1} i^{\tilde{\rho}} \left( \exp \left( - \frac{\mu C_{\gamma} C_{\rho}^{\beta\indicator{\rho\geq0}} i^{(1-\phi)(1+\tilde{\rho})}}{2^{1+(1-\phi)(1+\tilde{\rho})}} \right) \sqrt{\Pi_{\infty}^{v}}
+ \frac{2^{\phi(1+\tilde{\rho})} 6  \sigma^{2} C_{\gamma} C_{\rho}^{\beta}}{\mu C_{\rho}^{\indicator{\rho\geq0}} i^{\phi(1+\tilde{\rho})}} 
+ \frac{2^{3\phi(1+\tilde{\rho})/2-\tilde{\rho}/2} 13 \sigma^{2} C_{\gamma}^{3/2} C_{\rho}^{3\beta/2}}{\mu^{1/2} C_{\rho}^{\indicator{\rho\geq0}} i^{3\phi(1+\tilde{\rho})/2}} \right)
\\&\leq \frac{2^{\tilde{\rho}} C_{\nabla}' C_{\rho} \sqrt{\Pi_{\infty}^{v}} A_{\infty}^{v}}{\mu N_{t}}
+ \frac{2^{\phi(1+\tilde{\rho})+\tilde{\rho}} C_{\nabla}' \sigma^{2} C_{\gamma} C_{\rho}^{1+\beta}}{\mu^{2} C_{\rho}^{\indicator{\rho\geq0}} N_{t}} \sum_{i=1}^{t-1} i^{\beta\tilde{\rho}-\alpha} 
+ \frac{2^{3\phi(1+\tilde{\rho})/2+\tilde{\rho}/2} C_{\nabla}' \sigma^{2} C_{\gamma}^{3/2} C_{\rho}^{1+3\beta/2}}{\mu^{3/2} C_{\rho}^{\indicator{\rho\geq0}} N_{t}} \sum_{i=1}^{t-1} i^{3(\beta\tilde{\rho}-\alpha)/2},
\end{align*}
where the second term can be bounded as
\begin{align*}
\frac{2^{(1+\phi)(1+\tilde{\rho})-1} C_{\nabla}' \sigma^{2} C_{\gamma} C_{\rho}^{1+\beta}}{\mu^{2} C_{\rho}^{\indicator{\rho\geq0}} N_{t}} \sum_{i=1}^{t-1} i^{\beta\tilde{\rho}-\alpha}
\leq& \frac{2^{(1+\phi)(1+\tilde{\rho})-1} C_{\nabla}' \sigma^{2} C_{\gamma} C_{\rho}^{1+\beta} t^{1+\beta\tilde{\rho}-\alpha}}{(1+\beta\tilde{\rho}-\alpha)\mu^{2} C_{\rho}^{\indicator{\rho\geq0}} N_{t}}
\leq \frac{2^{(1+\phi)(1+\tilde{\rho})-2} C_{\nabla}' \sigma^{2} C_{\gamma}}{\mu^{2} C_{\rho}^{1-\phi-\beta} N_{t}^{\phi}},
\end{align*}
and the third term by
\begin{align*}
\frac{2^{3(1+\phi)(1+\tilde{\rho})/2} C_{\nabla}' \sigma^{2} C_{\gamma}^{3/2} C_{\rho}^{1+3\beta/2}}{\mu^{3/2} C_{\rho}^{\indicator{\rho\geq0}} N_{t}} \sum_{i=1}^{t-1} i^{3(\beta\tilde{\rho}-\alpha)/2}
\leq& \frac{2^{3(1+\phi)(1+\tilde{\rho})/2} C_{\nabla}' \sigma^{2} C_{\gamma}^{3/2} C_{\rho}^{1+3\beta/2}\psi_{3(\alpha-\beta\tilde{\rho})/2}^{\tilde{\rho}}(N_{t}/C_{\rho})}{\mu^{3/2} C_{\rho}^{\indicator{\rho\geq0}}N_{t}}.
\end{align*}
By collecting these bounds, we get
\begin{align*}
\frac{C_{\nabla}'}{N_{t} \mu} \sum_{i=0}^{t-1} n_{i+1} \Delta_{i}^{\frac{1}{2}}
\leq& \frac{2^{\tilde{\rho}} C_{\nabla}' C_{\rho} \sqrt{\Pi_{\infty}^{v}} A_{\infty}^{v}}{\mu N_{t}}
+\frac{2^{(1+\phi)(1+\tilde{\rho})-2} C_{\nabla}' \sigma^{2} C_{\gamma}}{\mu^{2} C_{\rho}^{1-\phi-\beta} N_{t}^{\phi}}
+ \frac{2^{3(1+\phi)(1+\tilde{\rho})/2} C_{\nabla}' \sigma^{2} C_{\gamma}^{3/2} C_{\rho}^{1+3\beta/2}\psi_{3(\alpha-\beta\tilde{\rho})/2}^{\tilde{\rho}}(N_{t}/C_{\rho})}{\mu^{3/2} C_{\rho}^{\indicator{\rho\geq0}}N_{t}}.
\end{align*}
Combining our findings from above, we have
\begin{align*}
\bar{\delta}_{t}^{1/2}
\leq& \frac{\Lambda^{1/2}}{N_{t}^{1/2}}
+ \frac{2 C_{\rho}^{1-\beta} \sqrt{\pi_{\infty}^{v}} A_{\infty}^{v}}{\mu C_{\gamma} N_{t}}
+ \frac{2^{\frac{3+\phi(1+\tilde{\rho})}{2}} \sigma C_{\rho}^{\frac{1-\phi-\beta}{2}\indicator{\rho\geq0}}}{\mu^{3/2} \sqrt{C_{\gamma}} N_{t}^{1-\phi/2}}
+\frac{C_{\rho}^{2-\phi-\beta} \sqrt{\pi_{\infty}^{v}} A_{\infty}^{v}}{\mu C_{\gamma} N_{t}^{2-\phi}}
+ \frac{2^{\frac{1+\phi(1+\tilde{\rho})}{2}} \sigma C_{\rho}^{\frac{1-\phi-\beta}{2}\indicator{\rho\geq0}}}{\mu^{3/2} \sqrt{C_{\gamma}} N_{t}^{1-\phi/2}}
\\ &+\frac{C_{\rho}}{N_{t} \mu} \left( \frac{1}{C_{\gamma} C_{\rho}^{\beta}} + C_{f} \right) \delta_{0}^{\frac{1}{2}}
+ \frac{2^{\tilde{\rho}/2} C_{f} \sqrt{C_{\rho}} \sqrt{\pi_{\infty}^{v}} \sqrt{A_{\infty}^{v}}}{\mu N_{t}} 
+ \frac{2^{\frac{\phi(1+\tilde{\rho})}{2}} C_{f} \sigma \sqrt{C_{\gamma}}}{\mu^{3/2} C_{\rho}^{\frac{1-\phi-\beta}{2}\indicator{\rho\geq0}} N_{t}^{\frac{1+\phi}{2}}}
+ \frac{2^{\tilde{\rho}} C_{\nabla}' C_{\rho} \sqrt{\Pi_{\infty}^{v}} A_{\infty}^{v}}{\mu N_{t}}
\\ &+\frac{2^{(1+\phi)(1+\tilde{\rho})-2} C_{\nabla}' \sigma^{2} C_{\gamma}}{\mu^{2} C_{\rho}^{1-\phi-\beta} N_{t}^{\phi}}
+ \frac{2^{3(1+\phi)(1+\tilde{\rho})/2} C_{\nabla}' \sigma^{2} C_{\gamma}^{3/2} C_{\rho}^{1+3\beta/2}\psi_{3(\alpha-\beta\tilde{\rho})/2}^{\tilde{\rho}}(N_{t}/C_{\rho})}{\mu^{3/2} C_{\rho}^{\indicator{\rho\geq0}} N_{t}}.
\end{align*}
This can be simplified to the desired using $\Gamma_{v}$ given by $(1/C_{\gamma}C_{\rho}^{\beta}+C_{f})\delta_{0}^{1/2} + 2^{\tilde{\rho}}C_{f}\sqrt{\pi_{\infty}^{v}A_{\infty}^{v}}/C_{\rho}^{1/2}+2\sqrt{\pi_{\infty}^{v}}A_{\infty}^{v}/C_{\gamma}C_{\rho}^{\beta}+2^{\tilde{\rho}}C_{\nabla}' \sqrt{\Pi_{\infty}^{v}}A_{\infty}^{v}$, consisting of the finite constants $\pi_{\infty}^{v}$, $\Pi_{\infty}^{v}$ and $A_{\infty}^{v}$.
\end{proof}

\subsubsection{Proofs for \texorpdfstring{\cref{sec:assg:bounded}}{3.2}}

\begin{theorem}[APSSG] \label{thm:assg:bounded_grads}
Let $\bar{\delta}_{t}=\E[\lVert \bar{\theta}_{t}-\theta^{*}\rVert^{2}]$ with $(\bar{\theta}_{t})$ given by \cref{eq:avg:streaming_grad_est}, where $(\theta_{t})$ follows the recursion in \cref{eq:ssg:proj}.
Suppose \cref{assump:L_strong_convexity,assump:L_lipschitz,assump:measurable,assump:ssg:f_lipschitz,assump:ssg:sigma,avg:assump:L_lipschitz,assump:assg:L_lipschitz} hold with $p=4$.
Then, for any learning rate $(\gamma_{t})$ and time-varying mini-batch $(n_{t})$, we can upper bound $\bar{\delta}_{t}^{1/2}$ by
\begin{align*}
\frac{\Lambda^{1/2}}{N_{t}^{1/2}}
+\frac{1}{N_{t}\mu} \sum_{i=1}^{t-1} \left| \frac{n_{i+1}}{\gamma_{i+1}} - \frac{n_{i}}{\gamma_{i}} \right| \delta_{i}^{1/2}
+ \frac{n_{t}}{N_{t} \gamma_{t} \mu} \delta_{t}^{1/2} 
+ \frac{n_{1}}{N_{t} \mu} \left( \frac{1}{\gamma_{1}} + C_{f} \right) \delta_{0}^{1/2} + \frac{C_{f}}{N_{t} \mu} \left( \sum_{i=1}^{t-1} n_{i+1} \delta_{i} \right)^{1/2}
+ \frac{C_{\nabla}''}{N_{t}\mu}\sum_{i=0}^{t} n_{i+1} \Delta_{i}^{1/2}
\end{align*}
where $\Lambda=\operatorname{Tr}(\nabla_{\theta}^{2}F(\theta^{*})^{-1}\Sigma\nabla_{\theta}^{2}F(\theta^{*})^{-1})$ and $C_{\nabla}''=C_{\nabla}'+2^{2}G_{\Theta}/D_{\theta}^{2}$.
\end{theorem}
\begin{proof}[Proof of \cref{thm:assg:bounded_grads}]
Denote $\E[\lVert \bar{\theta}_{t}-\theta^{*}\rVert^{2}]$ by $\bar{\delta}_{t}$ with $(\bar{\theta}_{t})$ given by \cref{eq:avg:streaming_grad_est} using $(\theta_{t})$ from \cref{eq:ssg:proj}.
As in the proof \cref{thm:avg:nonblock:streaming_grad_upper_bound}, we follow the steps of \citet{polyak1992acceleration}, in which, we can rewrite \cref{eq:ssg:proj} to
\begin{align*}
\frac{1}{\gamma_{t}} \left( \theta_{t-1} - \theta_{t} \right) = \nabla_{\theta} f_{t} \left( \theta_{t-1} \right) - \frac{1}{\gamma_{t}}\Omega_{t},
\end{align*}
where $\nabla_{\theta}f_{t}(\theta_{t-1})=n_{t}^{-1}\sum_{i=1}^{n_{t}}\nabla_{\theta}f_{t,i}(\theta_{t-1})$ and $\Omega_{t} =  \mathcal{P}_{\Theta}(\theta_{t-1}-\gamma_{t}\nabla_{\theta} f_{t}(\theta_{t-1}))-(\theta_{t-1}-\gamma_{t}\nabla_{\theta} f_{t}(\theta_{t-1}))$.
Thus, summing the parts, using the Minkowski's inequality, and bounding each term gives us the same bound as in \cref{thm:avg:nonblock:streaming_grad_upper_bound}, but with an additional term regarding $\Omega_{t}$, namely
\begin{align} \label{eq:omega_term}
\left(\E \left[ \left\lVert\nabla_{\theta}^{2} F \left( \theta^{*} \right)^{-1} \frac{1}{N_{t}} \sum_{i=1}^{t} \frac{n_{i}}{\gamma_{i}} \Omega_{i} \right\rVert^{2} \right] \right)^{\frac{1}{2}}
\leq& \frac{1}{\mu N_{t}}\sum_{i=1}^{t} \frac{n_{i}}{\gamma_{i}} \sqrt{\E\left[\left\lVert\Omega_{i} \right\rVert^{2} \right]}
= \frac{1}{\mu N_{t}}\sum_{i=1}^{t} \frac{n_{i}}{\gamma_{i}} \sqrt{\E\left[\left\lVert\Omega_{i}\right\rVert^{2}\indicator{\theta_{i-1}-\gamma_{i}\nabla_{\theta}f_{i}(\theta_{i-1})\notin\Theta} \right]},
\end{align}
using \citet[Lemma 4.3]{godichon2016estimating}.
Next, we note that $\E[\lVert\Omega_{t}\rVert^{2}\indicator{\theta_{t-1}-\gamma_{t}\nabla_{\theta}f_{t}(\theta_{t-1})\notin\Theta}]=4\gamma_{t}^{2}G_{\Theta}^{2}\mathbb{P}[\theta_{t-1}-\gamma_{t}\nabla_{\theta}f_{t}(\theta_{t-1})\notin\Theta]$, since
\begin{align*}
\left\lVert\Omega_{t}\right\rVert^{2}
=& \left\lVert\mathcal{P}_{\Theta}\left(\theta_{t-1}-\gamma_{t}\nabla_{\theta}f_{t}\left(\theta_{t-1}\right)\right)-\theta_{t-1}+\gamma_{t}\nabla_{\theta}f_{t}\left(\theta_{t-1}\right)\right\rVert^{2}
 \\ \leq& 2\left\lVert\mathcal{P}_{\Theta}\left(\theta_{t-1}-\gamma_{t}\nabla_{\theta}f_{t}\left(\theta_{t-1}\right)\right)-\theta_{t-1}\right\rVert^{2}+2\gamma_{t}^{2}\left\lVert\nabla_{\theta}f_{t}\left(\theta_{t-1}\right)\right\rVert^{2}
\\ =& 2\left\lVert\mathcal{P}_{\Theta}\left(\theta_{t-1}-\gamma_{t}\nabla_{\theta}f_{t}\left(\theta_{t-1}\right)\right)-\mathcal{P}_{\Theta}\left(\theta_{t-1}\right)\right\rVert^{2}+2\gamma_{t}^{2}\left\lVert\nabla_{\theta}f_{t}\left(\theta_{t-1}\right)\right\rVert^{2}
\\ \leq& 2\left\lVert\theta_{t-1}-\gamma_{t}\nabla_{\theta}f_{t}\left(\theta_{t-1}\right)-\theta_{t-1}\right\rVert^{2}+2\gamma_{t}^{2}\left\lVert\nabla_{\theta}f_{t}\left(\theta_{t-1}\right)\right\rVert^{2}
\\ =& 4\gamma_{t}^{2}\left\lVert\nabla_{\theta}f_{t}\left(\theta_{t-1}\right)\right\rVert^{2} \leq 4\gamma_{t}^{2}G_{\Theta}^{2},
\end{align*}
as $\mathcal{P}_{\Theta}$ is Lipschitz and  $\lVert \nabla_{\theta} f_{t,i}(\theta)\rVert^{2} \leq G_{\Theta}^{2}$ for any $\theta\in\Theta$.
Moreover, as in \citet[Theorem 4.2]{godichon2017averaged}, we know that  $\mathbb{P}[\theta_{t-1}-\gamma_{t}\nabla_{\theta}f_{t}(\theta_{t-1})\notin\Theta] \leq \Delta_{t}/D_{\theta}^{4}$, where $D_{\theta} = \inf_{\theta\in\partial\Theta} \lVert\theta-\theta^{*}\rVert$ with $\partial\Theta$ denoting the frontier of $\Theta$.
Thus, \cref{eq:omega_term} can then be bounded by
\begin{align*}
\frac{1}{\mu N_{t}}\sum_{i=1}^{t} \frac{n_{i}}{\gamma_{i}} \sqrt{\E\left[\left\lVert\Omega_{i}\right\rVert^{2}\indicator{\theta_{i-1}-\gamma_{i}\nabla_{\theta}f_{i}(\theta_{i-1})\notin\Theta} \right]} \leq \frac{2G_{\Theta}}{\mu D_{\theta}^{2}N_{t}}\sum_{i=1}^{t}n_{i}\Delta_{i}^{1/2}
\leq& \frac{2^{2}G_{\Theta}}{\mu D_{\theta}^{2}N_{t}}\sum_{i=1}^{t}n_{i+1}\Delta_{i}^{1/2},
\end{align*}
using that the sequence $(n_{t})$ is either constant or time-varying, meaning $n_{t+1}/n_{t}\leq2$.
\end{proof}

\begin{proof}[Proof of \cref{cor:assg:bounded:constant}]
The proof follows directly from \cref{cor:avg:streaming_grad_upper_bound_constant} with use of
\cref{thm:assg:bounded_grads}.
\end{proof}

\begin{proof}[Proof of \cref{cor:assg:bounded:varying}]
The proof follows directly from \cref{cor:avg:streaming_grad_upper_bound_increasing} with use of
\cref{thm:assg:bounded_grads}.
\end{proof}

\section{Technical propositions} \label{sec:appendix}

\Cref{sec:appendix} contains purely technical results used in the proofs presented in \cref{sec:proofs}.
In what follows, we use the convention $\inf \emptyset =0$, $\sum_{t=1}^0 = 0$, and $\prod_{t=1}^0=1$.
\begin{proposition} \label{prop:appendix:upper_bound_sum_prod}
Let $(\gamma_{t})_{t \geq 1}$ be a positive sequence. For any $k \leq t$, and $\omega > 0$, we have
\begin{align} \label{eq:prop:appendix:upper_bound_sum_prod}
\sum_{i=k}^{t} \prod_{j=i+1}^{t} \left[1 + \omega \gamma_{j} \right]  \gamma_{i}
\leq \frac{1}{\omega} \prod_{j=k}^{t} \left[ 1 + \omega \gamma_{j} \right]
\leq \frac{1}{\omega} \exp \left( \omega \sum_{j=k}^{t}  \gamma_{j} \right).
\end{align}
\end{proposition}

\begin{proof}[Proof of \cref{prop:appendix:upper_bound_sum_prod}]
We begin with considering the first inequality in \cref{eq:prop:appendix:upper_bound_sum_prod}, which follows by expanding the sum of product:
\begin{align*}
\sum_{i=k}^{t} \prod_{j=i+1}^{t} \left[1 + \omega \gamma_{j} \right]  \gamma_{i}
=& \frac{1}{\omega} \sum_{i=k}^{t} \prod_{j=i+1}^{t} \left[1 + \omega  \gamma_{j} \right] \omega  \gamma_{i}
= \frac{1}{\omega} \sum_{i=k}^{t} \prod_{j=i+1}^{t} \left[1 + \omega  \gamma_{j}\right] \left[1 + \omega  \gamma_{i} - 1 \right]
\\ =& \frac{1}{\omega} \sum_{i=k}^{t} \left[ \prod_{j=i+1}^{t} \left[1 + \omega  \gamma_{j} \right] \left[1 + \omega  \gamma_{i} \right] -  \prod_{j=i+1}^{t} \left[1 + \omega  \gamma_{j} \right] \right] 
= \frac{1}{\omega} \sum_{i=k}^{t} \left[ \prod_{j=i}^{t} \left[1 + \omega  \gamma_{j} \right] -  \prod_{j=i+1}^{t} \left[1 + \omega  \gamma_{j} \right] \right].
\end{align*}
As the (positive) terms cancel out, we end up with the first inequality in \cref{eq:prop:appendix:upper_bound_sum_prod}:
\begin{align*}
\frac{1}{\omega} \sum_{i=k}^{t} \left[ \prod_{j=i}^{t} \left[1 + \omega  \gamma_{j} \right] -  \prod_{j=i+1}^{t} \left[1 + \omega  \gamma_{j} \right] \right]
=& \frac{1}{\omega} \left[ \prod_{j=k}^{t} \left[1 + \omega  \gamma_{j} \right] -  \prod_{j=k+1}^{t} \left[1 + \omega  \gamma_{j} \right] + \dots -  \prod_{j=t+1}^{t} \left[1 + \omega  \gamma_{j} \right] \right]
\\ =& \frac{1}{\omega} \left[ \prod_{j=k}^{t} \left[1 + \omega  \gamma_{j} \right] -  \prod_{j=t+1}^{t} \left[1 + \omega  \gamma_{j} \right] \right]
\\ =& \frac{1}{\omega} \left[ \prod_{j=k}^{t} \left[1 + \omega  \gamma_{j} \right] -  1 \right]
\leq \frac{1}{\omega} \prod_{j=k}^{t} \left[1 + \omega \gamma_{j} \right],
\end{align*}
as $\prod_{t+1}^{t} = 1$ for all $t \in \N$.
Using the (simple) bound of $1+t \leq \exp (t)$ for all $t \in \R$, we obtain the second inequality of \cref{eq:prop:appendix:upper_bound_sum_prod}:
\begin{align*}
\frac{1}{\omega} \prod_{j=k}^{t} \left[1 + \omega  \gamma_{j} \right]
\leq \frac{1}{\omega} \prod_{j=k}^{t} \exp \left( \omega  \gamma_{j} \right)
= \frac{1}{\omega} \exp \left( \omega \sum_{j=k}^{t}  \gamma_{j} \right).
\end{align*}
\end{proof}
\begin{proposition} \label{prop:appendix:upper_bound_sum_prod_minus}
Let $(\gamma_{t})_{t \geq 1}$ be a positive sequence. Let $\omega > 0$ and $k \leq t$ such that for all $i \geq k$, $\omega \gamma_{i} \leq 1$, then
\begin{align} \label{eq:prop:appendix:upper_bound_sum_prod_minus}
\sum_{i=k}^{t} \prod_{j=i+1}^{t} \left[1 - \omega  \gamma_{j} \right]  \gamma_{i}
\leq \frac{1}{\omega}.
\end{align}
\end{proposition}
\begin{proof}[Proof of \cref{prop:appendix:upper_bound_sum_prod_minus}]
We start with expanding the sums of products term in \cref{eq:prop:appendix:upper_bound_sum_prod_minus}, given us
\begin{align*}
\sum_{i=k}^{t} \prod_{j=i+1}^{t} \left[1 - \omega \gamma_{j} \right]  \gamma_{i}
=& 
- \frac{1}{\omega} \sum_{i=k}^{t} \prod_{j=i+1}^{t} \left[1 - \omega  \gamma_{j} \right] \left[1 - \omega  \gamma_{i} - 1 \right] 
= - \frac{1}{\omega} \sum_{i=k}^{t} \left[ \prod_{j=i+1}^{t} \left[1 - \omega  \gamma_{j} \right] \left[1 - \omega  \gamma_{i} \right] -  \prod_{j=i+1}^{t} \left[1 - \omega  \gamma_{j} \right] \right] 
\\ =& -\frac{1}{\omega} \sum_{i=k}^{t} \left[ \prod_{j=i}^{t} \left[1 - \omega  \gamma_{j} \right] -  \prod_{j=i+1}^{t} \left[1 - \omega  \gamma_{j} \right] \right]
 = \frac{1}{\omega} \sum_{i=k}^{t} \left[ \prod_{j=i+1}^{t} \left[1 - \omega  \gamma_{j} \right] - \prod_{j=i}^{t} \left[1 - \omega  \gamma_{j} \right] \right].
\end{align*}
As we only have positive terms, we can upper bound the term:
\begin{align*}
\frac{1}{\omega} \sum_{i=k}^{t} \left[ \prod_{j=i+1}^{t} \left[1 - \omega  \gamma_{j} \right] - \prod_{j=i}^{t} \left[1 - \omega  \gamma_{j} \right] \right]
\leq \frac{1}{\omega} \left[ 1 - \prod_{j=k}^{t} \left[1 - \omega  \gamma_{j} \right] \right]
\leq \frac{1}{\omega},
\end{align*}
using $\prod_{j=k}^{t} [1 - \omega  \gamma_{j} ] \geq 0$, showing the inequality in \cref{eq:prop:appendix:upper_bound_sum_prod_minus}.
\end{proof}
\begin{proposition} \label{prop:appendix:upper_bound_sum_prod_minus_expand}
Let  $(\gamma_{t})_{t \geq 1}$ and $(\eta_{t})_{t \geq 1}$ be positive sequences.
For any $k \leq t$, we can obtain the (upper) bounds: 
\begin{align} \label{eq:prop:appendix:upper_bound_sum_prod_q}
\sum_{i=k}^{t} \prod_{j=i+1}^{t} \left[1 + \omega \gamma_{j} \right] \eta_{i} \gamma_{i}
\leq \frac{1}{\omega} \max_{k \leq i \leq t} \eta_{i} \exp \left( \omega \sum_{j=k}^{t}  \gamma_{j} \right),
\end{align}
with $\omega > 0$. Furthermore, suppose that for all $i \geq k$, $\omega \gamma_{i} \leq 1$,  then
\begin{align} \label{eq:prop:appendix:upper_bound_sum_prod_minus_q}
\sum_{i=k}^{t} \prod_{j=i+1}^{t} \left[1 - \omega  \gamma_{j} \right] \eta_{i} 
\leq \frac{1}{\omega} \max_{k \leq i \leq t} \eta_{i} .
\end{align}
\end{proposition}
\begin{proof}[Proof of \cref{prop:appendix:upper_bound_sum_prod_minus_expand}]
We obtain the inequality in \cref{eq:prop:appendix:upper_bound_sum_prod_q} directly by \cref{prop:appendix:upper_bound_sum_prod}:
\begin{align*} 
\sum_{i=k}^{t} \prod_{j=i+1}^{t} \left[1 + \omega  \gamma_{j} \right] \eta_{i} \gamma_{i}
\leq \max_{k \leq i \leq t} \eta_{i} \sum_{i=k}^{t} \prod_{j=i+1}^{t} \left[1 + \omega  \gamma_{j} \right]  \gamma_{i}
\leq \frac{1}{\omega} \max_{k \leq i \leq t} \eta_{i} \prod_{j=k}^{t} \left[ 1 + \omega  \gamma_{j} \right]
\leq \frac{1}{\omega} \max_{k \leq i \leq t} \eta_{i} \exp \left( \omega \sum_{j=k}^{t}  \gamma_{j}\right).
\end{align*}
Similarly, for the inequality in \cref{eq:prop:appendix:upper_bound_sum_prod_minus_q}, we have
\begin{align*}
\sum_{i=k}^{t} \prod_{j=i+1}^{t} \left[1 - \omega  \gamma_{j} \right] \eta_{i} \gamma_{i}
\leq \max_{k \leq i \leq t} \eta_{i} \sum_{i=k}^{t} \prod_{j=i+1}^{t} \left[1 - \omega \gamma_{j} \right]  \gamma_{i}
\leq \frac{1}{\omega} \max_{k \leq i \leq t} \eta_{i},
\end{align*}
by \cref{prop:appendix:upper_bound_sum_prod_minus}.
\end{proof}
\begin{proposition} \label{prop:appendix:delta_recursive_upper_bound_nt}
Let $(\delta_{t})_{t \geq 0}$, $(\gamma_{t})_{t \geq 1}$, $(\eta_{t})_{t \geq 1}$, and $(\nu_{t})_{t \geq 1}$ be some positive sequences satisfying the recursive relation:
\begin{align} \label{eq:prop:appendix:delta_recursive_single}
\delta_{t} \leq \left( 1 - 2 \omega \gamma_{t} + \eta_{t} \gamma_{t} \right) \delta_{t-1} + \nu_{t} \gamma_{t},
\end{align}
with $\delta_{0} \geq 0$ and $ \omega > 0$.
Denote $t_{0} = \inf \left\{ t\ge 1\, : \eta_{t} \leq \omega \right\}$, and suppose that for all $t \geq t_{0} +1$, one has $\omega \gamma_{t} \leq 1$.
Then, for $\gamma_{t}$ and $\eta_{t}$ decreasing, we have the upper bound on $(\delta_{t})$:
\begin{align} \label{eq:prop:appendix:delta_recursive_upper_single_bound}
\delta_{t} 
\leq& \exp \left( - \omega \sum_{i=t/2}^{t} \gamma_{i} \right) 
\left[ 
\exp \left( \sum_{i=1}^{t_{0}} \eta_{i} \gamma_{i} \right) \left( \delta_{0} + \max_{1 \leq i \leq t_{0}} \frac{\nu_{i}}{\eta_{i}} \right) + \sum_{i=t_{0}+1}^{t/2-1} \nu_{i} \gamma_{i} \right]
+ \frac{1}{\omega} \max_{t/2 \leq i \leq t} \nu_{i},
\end{align}
for all $t \in \N$ with the convention that $\sum_{t_{0}}^{t/2}=0$ if $t/2 < t_{0}$.
\end{proposition}
\begin{proof}[Proof of \cref{prop:appendix:delta_recursive_upper_bound_nt}]
Applying the recursive relation from \cref{eq:prop:appendix:delta_recursive_single} $t$ times, we derive:
\begin{align*}
\delta_{t} 
\leq& \underbrace{\prod_{i=1}^{t}\left[ 1 - 2 \omega \gamma_{i} + \eta_{i} \gamma_{i} \right]}_{B_{t}} \delta_{0} 
+ \underbrace{\sum_{i=1}^{t} \prod_{j=i+1}^{t} \left[ 1 - 2 \omega \gamma_{j} + \eta_{j} \gamma_{j} \right] \nu_{i} \gamma_{i}}_{A_{t}},
\end{align*}
where $B_{t}$ can be seen as a transient term only depending on the initialisation $\delta_{0}$, and a stationary term $A_{t}$.
The transient term $B_{t}$ can be divided into two products, before and after $t_{0}$,
\begin{align*}
B_{t} = \prod_{i=1}^{t}\left[ 1 - 2 \omega \gamma_{i} + \eta_{i} \gamma_{i} \right]
 = \left( \prod_{i=1}^{t_{0}}\left[ 1 - 2 \omega \gamma_{i} + \eta_{i} \gamma_{i} \right] \right)
\left( \prod_{i=t_{0}+1}^{t}\left[ 1 - 2 \omega \gamma_{i} + \eta_{i} \gamma_{i} \right] \right).
\end{align*}
Using that $t_{0} = \inf \left\{ t\ge 1\, : \eta_{t} \leq \omega \right\}$, and since for all $t \geq t_{0}+1$, we have $2 \omega \gamma_{t}-\eta_{t} \gamma_{t} \geq \omega \gamma_{t}$, it comes 
\begin{align}
B_{t} \leq& \left( \prod_{i=1}^{t_{0}}\left[ 1 - 2 \omega \gamma_{i} + \eta_{i} \gamma_{i} \right] \right) \nonumber
\left( \prod_{i=t_{0}+1}^{t}\left[ 1 - \omega \gamma_{i} \right] \right)
\leq \left( \prod_{i=1}^{t_{0}} \exp \left( - 2 \omega \gamma_{i} + \eta_{i} \gamma_{i} \right) \right) \nonumber
\left( \prod_{i=t_{0}+1}^{t} \exp \left( - \omega \gamma_{i} \right) \right)
\\ =& \exp \left( - 2 \omega \sum_{i=1}^{t_{0}} \gamma_{i}\right) \exp \left( \sum_{i=1}^{t_{0}} \eta_{i} \gamma_{i} \right)
\exp \left( - \omega \sum_{i=t_{0}+1}^{t} \gamma_{i} \right) \nonumber
\leq \exp \left( - \omega \sum_{i=1}^{t} \gamma_{i}\right) 
\exp \left( \sum_{i=1}^{t_{0}} \eta_{i} \gamma_{i} \right) \nonumber
\end{align}
by applying the (simple) bound $1+t \leq \exp(t)$ for all $t \in \R$.
We derive that
\begin{equation}
B_{t} \leq \exp \left( - \omega \sum_{i=t/2}^{t} \gamma_{i}\right) 
\exp \left( \sum_{i=1}^{t_{0}} \eta_{i} \gamma_{i} \right). \label{eq:prop:appendix:delta_recursive_bound_bt}
\end{equation}
Next, the stationary term $A_{t}$ can (similarly) be divided into two sums  (after and before $t_{0}$):
\begin{align*}
A_{t} = \underbrace{\sum_{i=t_{0}+1}^{t} \prod_{j=i+1}^{t} \left[ 1 - 2 \omega \gamma_{j} + \eta_{j} \gamma_{j} \right] \nu_{i} \gamma_{i}}_{A_{t,1}}
+ \underbrace{\sum_{i=1}^{t_{0}} \prod_{j=i+1}^{t} \left[ 1 - 2 \omega \gamma_{j} + \eta_{j} \gamma_{j} \right] \nu_{i} \gamma_{i}}_{A_{t,2}}.
\end{align*}
The first stationary term $A_{t,1}$ (with $t > t_{0})$ can be bounded as follows: if $t/2 \leq t_{0} +1 $, we have
\begin{align*}
A_{t,1} 
\leq \max_{t_{0}+1 \leq i \leq t }\nu_{i} \sum_{i=t_{0}+1}^{t} \prod_{j=i+1}^{t} \left[ 1- \omega \gamma_{j} \right] \gamma_{i} 
= \frac{1}{\omega } \max_{t_{0}+1 \leq i \leq t } \nu_{i} 
\leq \frac{1}{\omega } \max_{t /2 \leq i \leq t } \nu_{i},
\end{align*}
by \cref{prop:appendix:upper_bound_sum_prod_minus_expand}.
Furthermore, if $t/2 > t_{0}+1$, we get
\begin{align*}
A_{t,1} \leq& \sum_{i=t_{0}+1}^{t} \prod_{j=i+1}^{t} \left[ 1 - \omega \gamma_{j} \right] \nu_{i} \gamma_{i}
= \sum_{i=t_{0}+1}^{t/2-1} \prod_{j=i+1}^{t} \left[ 1 - \omega \gamma_{j} \right] \nu_{i} \gamma_{i} + \sum_{i=t/2}^{t} \prod_{j=i+1}^{t} \left[ 1 - \omega \gamma_{j} \right] \nu_{i} \gamma_{i} 
\\ \leq& \sum_{i=t_{0}+1}^{t/2-1} \prod_{j=t/2}^{t} \left[ 1 - \omega \gamma_{j} \right] \nu_{i} \gamma_{i} + \max_{t/2 \leq i \leq t} \nu_{i} \sum_{i=t/2}^{t} \prod_{j=i+1}^{t} \left[ 1 - \omega \gamma_{j} \right] \gamma_{i} 
= \prod_{j=t/2}^{t} \left[ 1 - \omega \gamma_{j} \right] \sum_{i=t_{0}+1}^{t/2-1} \nu_{i} \gamma_{i} + \frac{1}{\omega} \max_{t/2 \leq i \leq t} \nu_{i},
\end{align*}
where $\prod_{j=t/2}^{t}[1-\omega\gamma_{j}]\leq\exp(-\omega\sum_{j=t/2}^{t}\gamma_{j})$ as $1+t \leq \exp(t)$ for all $t \in \R$.
Thus, for all $t \in \R$,
\begin{align}
A_{t,1} \leq& \exp \left(- \omega \sum_{j=t/2}^{t} \gamma_{j} \right) \sum_{i=t_{0}+1}^{t/2-1} \nu_{i} \gamma_{i} + \frac{1}{\omega} \max_{t/2 \leq i \leq t} \nu_{i},
\label{eq:prop:appendix:delta_recursive_bound_at1}
\end{align}
where $\sum_{t_{0}}^{t/2}=0$ if $t/2 < t_{0}$.
The second stationary term $A_{t,2}$ can be bounded, thanks to \cref{prop:appendix:upper_bound_sum_prod}, as follows:
\begin{align*}
A_{t,2} =& \sum_{i=1}^{t_{0}} \prod_{j=i+1}^{t} \left[ 1 - 2 \omega \gamma_{j} + \eta_{j} \gamma_{j} \right] \nu_{i} \gamma_{i}
= \left( \prod_{j=t_{0}+1}^{t} \left[ 1 - 2 \omega \gamma_{j} + \eta_{j} \gamma_{j} \right]\right) \sum_{i=1}^{t_{0}} \prod_{j=i+1}^{t_{0}} \left[ 1 - 2 \omega \gamma_{j} + \eta_{j} \gamma_{j} \right] \nu_{i} \gamma_{i}
\\ \leq& \left( \prod_{j=t_{0}+1}^{t} \left[ 1 - \omega \gamma_{j} \right]\right) \sum_{i=1}^{t_{0}} \prod_{j=i+1}^{t_{0}} \left[ 1 + \eta_{j} \gamma_{j} \right] \nu_{i} \gamma_{i}
\leq \exp \left( - \omega \sum_{j=t_{0}+1}^{t} \gamma_{j} \right) \max_{1 \leq i \leq t_{0}} \frac{\nu_{i}}{\eta_{i}}\sum_{i=1}^{t_{0}} \prod_{j=i+1}^{t_{0}} \left[ 1 + \eta_{j} \gamma_{j} \right] \eta_{i} \gamma_{i}
\\ \leq& \exp \left( - \omega \sum_{j=t_{0}+1}^{t} \gamma_{j} \right) \max_{1 \leq i \leq t_{0}} \frac{\nu_{i}}{\eta_{i}} \exp \left( \sum_{i=1}^{t_{0}} \eta_{i} \gamma_{i} \right)
\leq \exp \left( - \omega \sum_{j=1}^{t} \gamma_{j} \right) \max_{1 \leq i \leq t_{0}} \frac{\nu_{i}}{\eta_{i}} \exp \left( 2\sum_{i=1}^{t_{0}} \eta_{i} \gamma_{i} \right),
\end{align*}
by the definition of $t_{0}$, thus
\begin{align}
A_{t,2} \leq \exp \left( - \omega \sum_{j=1}^{t} \gamma_{j} \right) \max_{1 \leq i \leq t_{0}} \frac{\nu_{i}}{\eta_{i}} \exp \left( 2 \sum_{i=1}^{t_{0}} \eta_{i} \gamma_{i} \right) 
\leq& \exp \left( - \omega \sum_{j=t/2}^{t} \gamma_{j} \right)
\max_{1 \leq i \leq t_{0}} \frac{\nu_{i}}{\eta_{i}}
\exp \left( 2 \sum_{i=1}^{t_{0}} \eta_{j} \gamma_{j} \right).
\label{eq:prop:appendix:delta_recursive_bound_at2}
\end{align}
Then, using the bound for $A_{t,1}$ in \cref{eq:prop:appendix:delta_recursive_bound_at1} and $A_{t,2}$ in \cref{eq:prop:appendix:delta_recursive_bound_at2}, we can bound $A_{t}$ by
\begin{align}
A_{t} \leq& \exp \left( - \omega \sum_{j=t/2}^{t} \gamma_{j} \right)
\left[ \exp \left( 2 \sum_{i=1}^{t_{0}} \eta_{j} \gamma_{j} \right)
 \max_{1 \leq i \leq t_{0}} \frac{\nu_{i}}{\eta_{i}}
 + \sum_{i=t_{0}+1}^{t/2-1} \nu_{i} \gamma_{i} 
\right]
+ \frac{1}{\omega} \max_{t/2 \leq i \leq t} \nu_{i}.
\label{eq:prop:appendix:delta_recursive_bound_at}
\end{align}
Finally, combining the bound for $B_{t}$ in \cref{eq:prop:appendix:delta_recursive_bound_bt} and $A_{t}$ in \cref{eq:prop:appendix:delta_recursive_bound_at}, we achieve the bound for $\delta_{t} \leq B_{t} \delta_{0} + A_{t}$, namely the upper bound in \cref{eq:prop:appendix:delta_recursive_upper_single_bound}.
\end{proof}
The following proposition is a more simplistic but rougher version of the bound in \cref{prop:appendix:delta_recursive_upper_bound_nt}.
\begin{proposition} \label{prop:appendix:delta_recursive_upper_bound_nt_simpler}
Let $(\delta_{t})_{t \geq 0}$, $(\gamma_{t})_{t \geq 1}$, $(\eta_{t})_{t \geq 1}$, and $(\nu_{t})_{t \geq 1}$ be some positive sequences satisfying the recursive relation in \cref{eq:prop:appendix:delta_recursive_single}.
Denote $t_{0} = \inf \left\{ t\ge 1\, : \eta_{t} \leq \omega \right\}$, and suppose that for all $t \geq t_{0} +1$, one has $\omega \gamma_{t} \leq 1$.
Then, for $\gamma_{t}$ and $\eta_{t}$ decreasing, we have for all $t \in \N$,
\begin{align} \label{eq:prop:appendix:delta_recursive_upper_single_bound_simpler}
\delta_{t} 
\leq& \exp \left( - \omega \sum_{i=t/2}^{t} \gamma_{i} \right) 
\exp \left( 2 \sum_{i=1}^{t} \eta_{i} \gamma_{i} \right) 
\left( \delta_{0} + 2 \max_{1 \leq i \leq t} \frac{\nu_{i}}{\eta_{i}} \right)
+ \frac{1}{\omega} \max_{t/2 \leq i \leq t} \nu_{i}.
\end{align}
\end{proposition}
\begin{proof}[Proof of \cref{prop:appendix:delta_recursive_upper_bound_nt_simpler}]
The resulting (upper) bound in \cref{eq:prop:appendix:delta_recursive_upper_single_bound_simpler} follows directly from \cref{eq:prop:appendix:delta_recursive_upper_single_bound} by noting that $t_{0} \leq t$, giving us $\sum_{i=t_{0}+1}^{t/2-1}\nu_{i}\gamma_{i} \leq \sum_{i=1}^{t}\nu_{i}\gamma_{i} \leq \max_{1 \leq i \leq t}(\nu_{i}/\eta_{i})\sum_{i=1}^{t}\eta_{i}\gamma_{i} \leq \max_{1 \leq i \leq t}(\nu_{i}/\eta_{i}) \exp( 2 \sum_{i=1}^{t} \eta_{i}\gamma_{i})$, as $(\nu_{t})$ and $(\gamma_{t})$ are positive sequences.
\end{proof}

\end{document}